\documentclass[preprint,12pt]{elsarticle}

\usepackage[margin=1in]{geometry}
\usepackage[x11names]{xcolor}
\usepackage{array}
\usepackage{tabularx}
\usepackage{booktabs}
\usepackage{tcolorbox}
\usepackage{hyperref}
\usepackage{amsfonts} 
\usepackage{breqn}
\usepackage{multirow}
\usepackage{pdflscape}
\usepackage{float}
\usepackage{caption}
\usepackage[dvipsnames]{xcolor}
\usepackage{bm} 
\definecolor{darkred}{rgb}{0.9,0.1,0.1}
\usepackage{subcaption}
\usepackage{amsthm}

\theoremstyle{definition}

\usepackage{mathtools}
\usepackage{algorithmic}

\newcommand{\rev}[1]{#1}

\usepackage[ruled,vlined]{algorithm2e}
\hypersetup{
 colorlinks,
 citecolor=black,
 filecolor=black,
 linkcolor=black,
 urlcolor=black
}
\usepackage{amssymb}

\journal{arXiv}

\begin{document}

\begin{frontmatter}

\title{ATHENA: Agentic Team for Hierarchical Evolutionary Numerical Algorithms}

\author[inst1]{Juan Diego Toscano}
\author[inst1]{Daniel T.~Chen}
\author[inst1,label2]{George Em Karniadakis}
\affiliation[inst1]{organization={Division of Applied Mathematics, Brown University},
   city={Providence},
   postcode={02912}, 
   state={RI},
   country={USA}}

\fntext[label2]{Corresponding author: george\_karniadakis@brown.edu}
\begin{abstract}
Progress in computational science depends on complex numerical workflows that must faithfully encode physical laws, yet translating conceptual insight into reliable code remains a major bottleneck. Although large language models can generate isolated code fragments, they lack the structured reasoning required to design, verify, and iteratively refine complete scientific pipelines. Here we introduce ATHENA, an agentic framework explicitly designed to emulate scientific research modeled as a knowledge-driven contextual bandit process. Its core loop separates conceptual policy from numerical realization through expert-derived conceptual scaffolding, enabling principled diagnosis, reformulation, and repair of computational strategies. Across scientific computing and scientific machine learning tasks, ATHENA \rev{autonomously derives and correctly applies} exact analytical solutions, constructs stable numerical solvers, diagnoses ill-posed formulations, and orchestrates hybrid symbolic–numeric workflows. \rev{Quantitatively, ATHENA matches and frequently surpasses the accuracy of expert-authored reference solutions reported in the literature on canonical benchmarks.} By reframing computation as an object of agentic reasoning, our framework enables autonomous orchestration of heterogeneous algorithms across scientific domains.
\end{abstract}


\begin{keyword}
Scientific Computing, Scientific Machine Learning, Contextual Bandit, Agentic Systems, Autonomous Research Frameworks
\end{keyword}

\end{frontmatter}


\section{Introduction}

The rapid progress of large-scale foundation models (FMs) has expanded their role in scientific research, enabling code generation, data synthesis, and hypothesis exploration \cite{wiggins2022opportunities,schneider2024foundation,lu2024ai}. In Applied Mathematics and Scientific Computing, these capabilities suggest a possible path toward automating parts of numerical analysis, where high-fidelity simulation relies on carefully constructed methods for the partial differential equations (PDEs) that govern physical systems \cite{choi2025defining,subramanian2023towards,herde2024poseidon}. Developing such methods requires satisfying the stability, consistency, and convergence constraints \cite{karniadakis2005spectral,raissi2017physicsI}. Although FMs can generate fragments of these workflows, the gap between probabilistic text generation and the deterministic rigor required for reliable simulation remains significant \cite{jiang2025deepseek}. In practice, implementation and debugging form a substantial barrier: small errors in discretization, constraint enforcement, or solver configuration can prevent an idea from being evaluated.

Several limitations of single-model workflows become apparent in this setting. As documentation length and prompt size increase, models often overlook assumptions embedded deep within a prompt—the “Lost in the Middle’’ failure mode \cite{liu2024lost}. Their autoregressive generation process also introduces cumulative deviations from the intended scientific design, known as conceptual drift \cite{gama2014survey,ross2011reduction}. Furthermore, current models cannot visually inspect loss curves, diagnostic plots, or solution fields to determine whether a computation satisfies expected physical or numerical properties \cite{wu2023gpt}. Together, these issues limit the reliability of long-horizon reasoning and motivate a shift toward systems that distribute reasoning and execution across multiple components.

These challenges have motivated the development of agentic frameworks for scientific work. Some systems focus on high-level conceptual reasoning, generating hypotheses or navigating structured scientific knowledge \cite{ghafarollahi2025sciagents,samadi2025flexible}. Others automate specific numerical tasks such as hyperparameter optimization for PINNs \cite{wuwu2025pinnsagent}, natural-language-to-code translation \cite{he2025lang}, ODE-inference pipelines \cite{bhatnagar2025agentic,ghafarollahi2025rapid}, or surrogate-assisted design \cite{lupoiu2025multi,ghafarollahi2025rapid}. A complementary line of research uses evolutionary strategies and swarm intelligence to refine candidate solutions \cite{georgiev2025mathematical,aygun2025ai,jiang2025agenticsciml,wang2025swarms}. While these systems reduce implementation burden or search for improved variants, they typically address isolated parts of the workflow. Their agent interactions are often heuristic, making it difficult to ensure coherent refinement across the full lifecycle of scientific computation. \rev{Most directly related are recent LLM-driven systems for automated PDE solving, such as CodePDE, which generates PDE solvers through LLM reasoning, debugging, and self-refinement~\cite{li2025codepde}, and PDE-Agent, which couples a tool-chain with multi-agent planning and validation~\cite{liu2025pde}. A complementary line of program-search systems evolves code against automated evaluators, including FunSearch~\cite{romera2024mathematical} and AlphaEvolve~\cite{georgiev2025mathematical} for mathematical and algorithmic discovery, and AutoNumerics-Zero for the discovery of efficient numerical functions~\cite{real2023autonumerics}. A further line of agentic systems targets general machine-learning engineering and its evaluation, such as AIDE~\cite{jiang2025aide}, MLAgentBench~\cite{huang2023mlagentbench}, and MLE-bench~\cite{chan2025mle}. ATHENA shares the closed-loop propose--execute--evaluate pattern of these systems, but rather than optimizing code against a fixed scalar objective, it reasons about \emph{which} computational strategy to adopt using compact, theory-grounded conceptual scaffolds and closes the loop with multimodal diagnosis of solution fields and convergence behavior across the full scientific-computing lifecycle.}

To introduce a principled mechanism for coordinating agents, we draw on ideas from {\it Online Learning}. Scientific computation proceeds through a sequence of decisions: selecting modeling assumptions, constructing or modifying algorithms, implementing them, evaluating the results, and refining the approach. We model this process as a Contextual Bandit problem \cite[Chapter 18]{lattimore2020bandit}. At each step $n$, the system chooses an action $A_n$ representing a structural modification to a computational pipeline. This action is translated into an executable code state $S_n$, producing observations $O_n$ from which a scientific reward $R_n$ is computed. This reward reflects numerical correctness, physical plausibility, diagnostic behavior, and implementation fidelity. The bandit formulation provides a natural structure for organizing agent responsibilities and information flow.

We operationalize this framework through the HENA (Hierarchical Evolutionary Numerical Algorithms) loop. HENA assigns specific roles to different agentic teams: Conceptualization Teams analyze prior outcomes and propose the next action $A_n$, while Implementation Teams translate these decisions into code, execute the resulting programs, and collect observations. This division mirrors how scientific work is typically performed, separating high-level reasoning from detailed implementation and verification.

Because the space of numerical strategies is large, we constrain the policy using an organizing principle we call \emph{Conceptual Scaffolding}. Rather than relying on large context windows or raw literature—which tend to amplify context-related failures—we restrict actions to a compact set of expert-derived blueprints. These blueprints encode fundamental principles from approximation theory \cite{cybenko1989approximation,hornik1989multilayer,Kolmogorov1957,song2025explicit}, optimization \cite{toscano2025variational}, and numerical analysis. This structure reduces the search space and provides stable guidance for diagnosing failures and proposing coherent modifications.

Within the HENA loop, Advisor agents interpret results produced by the Implementation Teams, identify potential failure modes, and reason about appropriate adjustments within the scaffolding. This mechanism allows the system to detect numerical inconsistencies, interpret diagnostic plots, and revise modeling choices based on observed behavior rather than textual cues alone. As a result, the system can derive exact solutions, impose physical constraints, or restructure computational pipelines to address specific challenges in Scientific Computing (SciC) and Scientific Machine Learning (SciML).

Building on this architecture, we introduce ATHENA (Agentic Team for Hierarchical Evolutionary Numerical Algorithms), a general framework for constructing and refining scientific workflows. \rev{ATHENA coordinates a team of specialized agents that propose a computational strategy, retrieve and adapt reference code, implement and debug it, execute it in an external compute environment, and critically evaluate the result, iterating until the solution is verified.} ATHENA applies HENA across several task domains—including SciC, PINNs, and Operator Learning—and can assemble hybrid numerical pipelines that combine data-driven models with classical solvers for multiphysics problems. \rev{Crucially, the reward trajectories we observe show sublinear, submartingale-like improvement, indicating that the conceptual scaffolding guides the search efficiently rather than relying on unguided trial and error. This design is \emph{inspired by} the contextual-bandit formulation of the research lifecycle, which provides the conceptual model for how ATHENA balances the exploration of new strategies against the exploitation of those already working.}  While ATHENA can operate autonomously on standard benchmarks, it is also designed for expert-in-the-loop collaboration. This allows researchers to guide high-level decisions while offloading implementation and diagnostic tasks to the agentic system, shifting the human role toward conceptual reasoning and reducing the implementation burden that often slows progress in both SciC and SciML.

\section{Results}\label{sec2}

\subsection{ATHENA: an agentic framework inspired by Contextual Bandits}
\label{Agentic_Research}

\begin{figure}[H]
    \centering
    \includegraphics[width=1\linewidth]{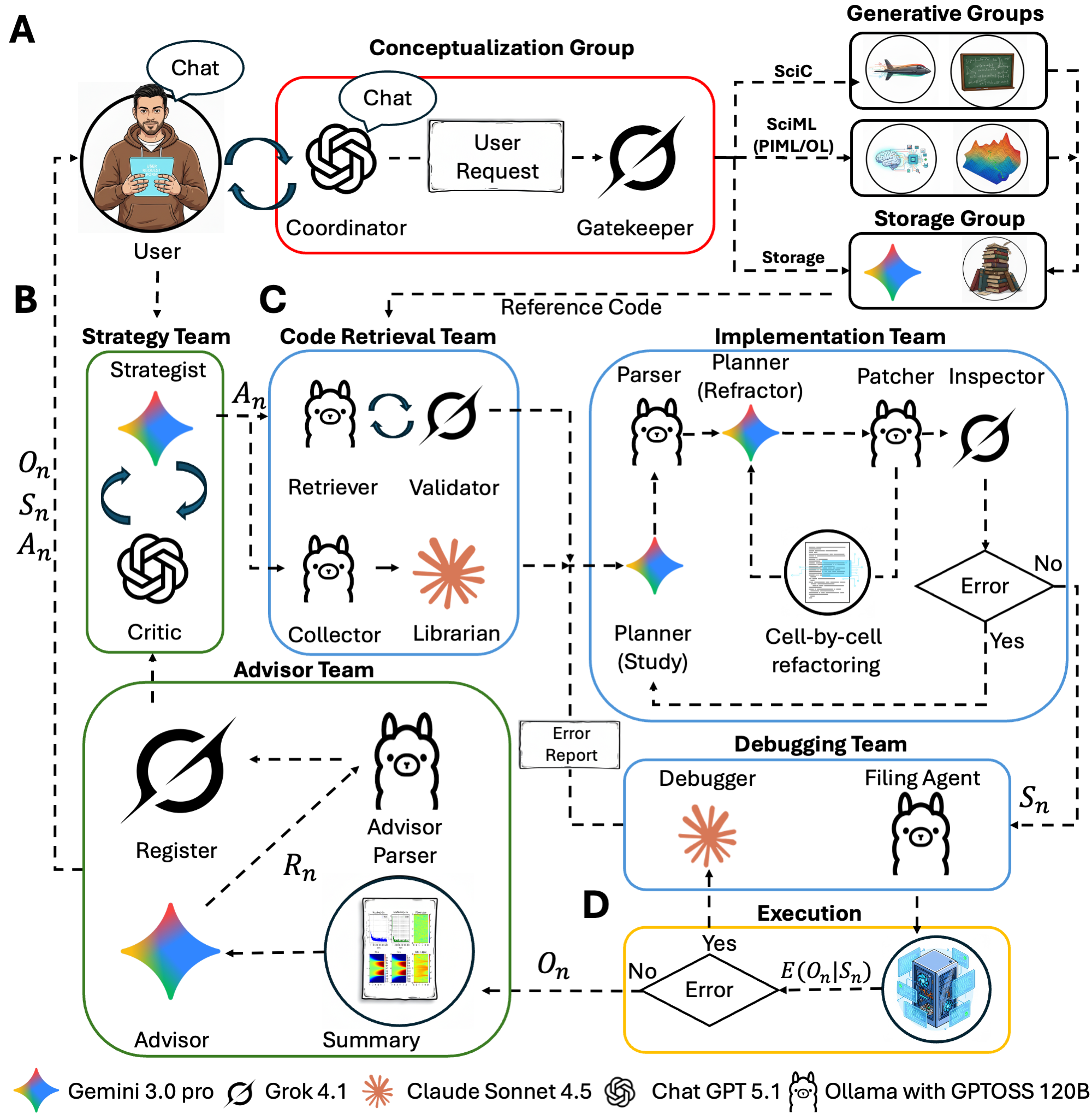}
\caption{
\textbf{ATHENA framework}. Unlike planner-only agentic systems, ATHENA executes, validates, and iteratively repairs scientific code. The framework is organized into four logical groups, with specific icons indicating the heterogeneous allocation of LLMs to specialized roles (see legend). (A) The Conceptualization Group (Red): The user-facing triage system. The User interacts with a Coordinator to define a User Request, which the Gatekeeper routes to the appropriate team. (B) The `Policy' Operator $\pi$ (Green): The ``brain" of the HENA loop, composed of the Strategy Team and Advisor Team. This group analyzes the context $C_n$ to formulate the structural action $A_n$ and assigns the scientific reward $R_n$ based on the resulting observations. (C) The `Implementation' Operator $\mathcal{I}$ (Blue): The ``executor" teams (Code Retrieval, Implementation, Debugging) that build, patch (``cell-by-cell refactoring"), and debug the code to produce the executable state $S_n$. Crucially, this workflow includes an Inspector Agent, which strictly verifies that the implementation faithfully follows the desired plan $A_n$ before execution. (D) The Execution Block (Yellow): This component runs the observation function $E(O_n | S_n)$ on the state $S_n$ to generate the multi-modal observation $O_n$ (plots, logs). This observation $O_n$ is sent back to the Advisor Team (B), completing the evolutionary cycle. The user also has access to $A_n$, $S_n$, and $O_n$ at each iteration, enabling a transparent expert-in-the-loop workflow.} \label{Framework} 
\end{figure}

In some scientific laboratories, research proceeds through a structured, iterative cycle of study, strategy design, implementation, and error analysis (Supplementary Fig.~\ref{fig:research_cycle}). 
We model this natural workflow as a Contextual Bandit problem \cite[Chapter 18]{lattimore2020bandit}, formalized into four distinct stages: the proposal of a structured hypothesis or action $A_n$ (Strategy), the translation of this plan into a concrete executable code state $S_n$ (Implementation), the execution of the code to produce observations $O_n$ (Execution), and the rigorous evaluation of these results to assign a scalar scientific reward $R_n$ (Error Analysis). 
The critical phase of refinement is governed by a policy $\pi$, which determines the distribution of the next action based on the history of prior outcomes. 
Much like a human expert learning from failure, the policy dictates \emph{how} the strategy should evolve after analyzing the reward and history.

To operationalize this viewpoint, we introduce ATHENA. 
The framework employs a \emph{Conceptualization Team} to interact with the user and formulate the formal problem scope (See Figure~\ref{Framework}(A)). 
This definition triggers the HENA (Hierarchical Evolutionary Numerical Algorithms) loop, which is explicitly designed to follow the Contextual Bandit formulation to refine the solution. 
HENA consists of a \emph{Strategy Group} (Strategy), an \emph{Implementation Group} (Implementation), and an \emph{Error Refinement Group} (Advisor) that analyzes the results of execution. 
By structuring the agents in this manner, our goal is to design a policy that is \emph{sample-efficient}; formally, we aim for the expected reward to behave as a submartingale, i.e., $\mathbb{E}[R_{n+1} \mid \text{History}_n] \ge R_n$, ensuring that the iterative process improves on average rather than relying on unguided trial and error.

At each iteration, the Strategy group proposes the next action $A_n$ (See Figure~\ref{Framework}(B)); the Implementation Teams translate this decision into an executable code state; and the Advisor team analyzes the resulting observables to assign the reward $R_n$ (See Figure~\ref{Framework}(C)-(D)).
The action space $\mathcal{A}$ is discrete and combinatorial, obtained through a modular decomposition of the task. 
For instance, in Physics-Informed Machine Learning, we factor the choices into representation, constraint enforcement, and optimization. 
This construction ensures that the search space is expressive while satisfying the finiteness required for the Contextual Bandit formulation.

To guide the policy toward meaningful structural updates, we constrain decisions using a compact set of expert-derived \emph{Conceptual Scaffolds} drawn from approximation theory \cite{cybenko1989approximation,hornik1989multilayer,Kolmogorov1957,song2025explicit}, optimization \cite{toscano2025variational}, and numerical analysis. 
These scaffolds are concretely encoded into the system prompts and base code templates used by the agents. 
They anchor the decision process to fundamental principles, allowing the system to diagnose failure modes without relying on large context windows.

Finally, to ensure that the maximal reward is attainable under realistic numerical constraints, we design a composite reward function that evaluates code integrity, quantitative accuracy, diagnostic behavior, and computational efficiency. 
By basing accuracy thresholds on physically meaningful tolerances rather than exact minima, the resulting scoring remains compatible with finite-precision computation. 
Together, these components yield a structured agentic loop that enables ATHENA to interpret scientific research as an online decision-making problem, supporting iterative refinement and principled navigation of complex computational workflows.

\rev{Unless otherwise stated, every result reported below corresponds to the configuration accepted by the Advisor agent; for trained models such as PINNs and neural operators, the reported error is taken at the best checkpoint of that accepted run.}

\subsection{ATHENA identifies hidden symmetries to \rev{derive} exact analytical solutions}

\begin{figure}[H]
    \centering
    \includegraphics[width=1\linewidth]{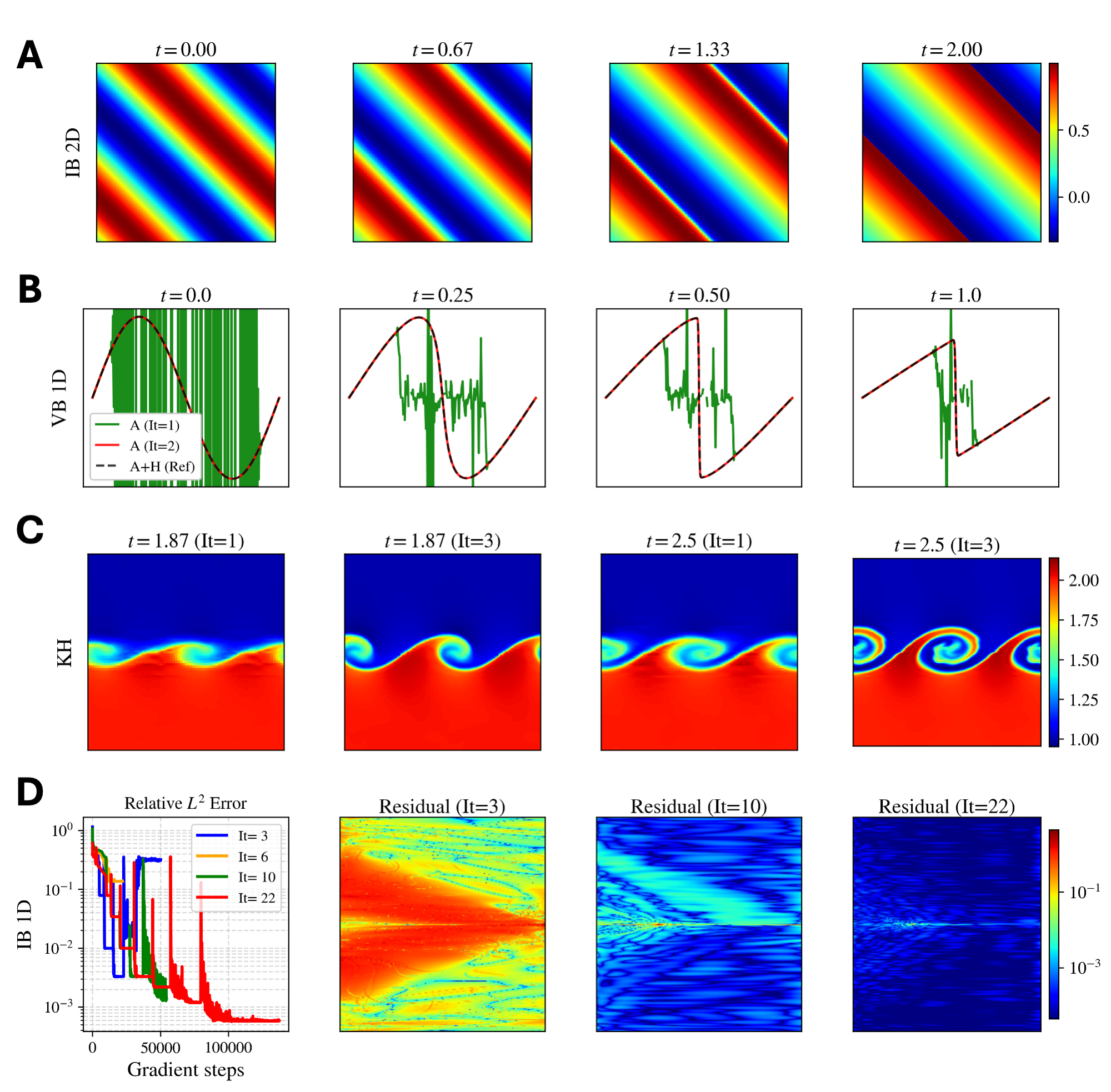}
\caption{\textbf{ATHENA utilizes conceptual scaffolding for evolutionary refinement across analytical, numerical, and machine learning problems.} (A) For the 2D Inviscid Burgers equation, the agent identifies a symmetry reducing the domain dimensionality, enabling the application of the method of characteristics to obtain exact solutions within 2 coding iterations and a single strategy. (B) In the 1D Viscous Burgers regime, the system initially attempts a Cole-Hopf transformation but detects numerical instability with standard quadrature. It autonomously pivots to a Fourier spectral method in the second iteration (solid lines). The dashed black line demonstrates that expert human intervention (suggesting Gauss-Hermite quadrature) allows for an exact solution in a single iteration (A+H). (C) Evolution of the Kelvin-Helmholtz instability using DGSEM. The first iteration (left) exhibits excessive numerical dissipation. The agents identify this deficiency and implement Hennemann--Gassner shock capturing with blended fluxes (DG + FV) and density-gradient adaptive mesh refinement (AMR) to recover sharp features by the third iteration (right). (D) Performance on an ill-posed inverse problem. The leftmost plot tracks the relative $L^2$ error across four representative iterations. The subsequent heatmaps visualize the residual landscape for the first successful run (It=3) compared to the final iteration (It=22), which incorporates the KAN architecture. As a result of these enhancements, the residual error reduces by four orders of magnitude and becomes significantly more uniform in the final solution.}\label{refine_athena} 
\end{figure}

Guided by Conceptual Scaffolding that prioritizes optimality over approximation, ATHENA autonomously determines that for specific regimes, the ``optimal" numerical strategy is to bypass discretization entirely. To evaluate this symbolic reasoning capacity, we tasked the autonomous HENA loop with solving a suite of five canonical PDEs: Black-Scholes, Viscous Burgers, Anisotropic Diffusion, 2D Inviscid Burgers, and the 3D Wave Equation (See Supplementary Figure~\ref{Exact_solutions} (A)-(E)). In all cases, the Strategist agent diagnostically mapped the complex operators to solvable canonical primitives, autonomously deriving the analytical solution and  generating and implementing the  code to compute it.

In the parabolic regime, the agent discovered that nonlinear or variable-coefficient operators could be linearized or diagonalized into the standard Heat Equation. For the 1D Viscous Burgers' equation, it identified the Cole-Hopf transformation ($u = -2\nu \phi_x/\phi$), converting the nonlinear conservation law into a linear diffusion problem solvable via heat kernel convolution. Similarly, for the Black-Scholes equation, it applied a log-price transformation ($y = \ln(S/K)$) akin to the Feynman-Kac formula to remove the drift terms, recovering the exact European call option pricing formula. In the case of 2D Anisotropic Diffusion, the agent diagonalized the cross-derivative operator via a rotational change of variables ($\phi = x+y, \psi = x-y$), decoupling the modes to enable an exact Fourier spectral reconstruction.

In the hyperbolic regime, where numerical methods struggle with dispersion, the agent exploited geometric invariances. For the 3D Wave equation, it utilized spherical symmetry ($v = r u$) to reduce the system to a 1D wave equation solvable via eigenfunction expansion. Most notably, for the 2D Inviscid Burgers' equation, the agent derived the Method of Characteristics, solving the implicit transport equation along characteristic curves. Upon detecting shock formation, it autonomously enforced entropy conditions (branch selection) to recover the exact discontinuous solution without the smearing typical of finite difference schemes. 

Throughout this process, ATHENA generates a comprehensive analysis document at every iteration of the HENA loop. These reports capture the specific user request, the rigorous mathematical strategy (including full symbolic derivations), the diagnostic analysis from the Advisor agent, and the cumulative history of prior refinement attempts. The final strategies corresponding to these successful exact solutions are provided in the Supplementary Information (SI). Furthermore, the complete evolutionary history, along with the executable code generated by the agents, is provided in the associated code repository.

\newpage

\subsection{Autonomous orchestration to obtain high-order numerical fidelity}

\begin{figure}[H]
    \centering
    \includegraphics[width=\linewidth]{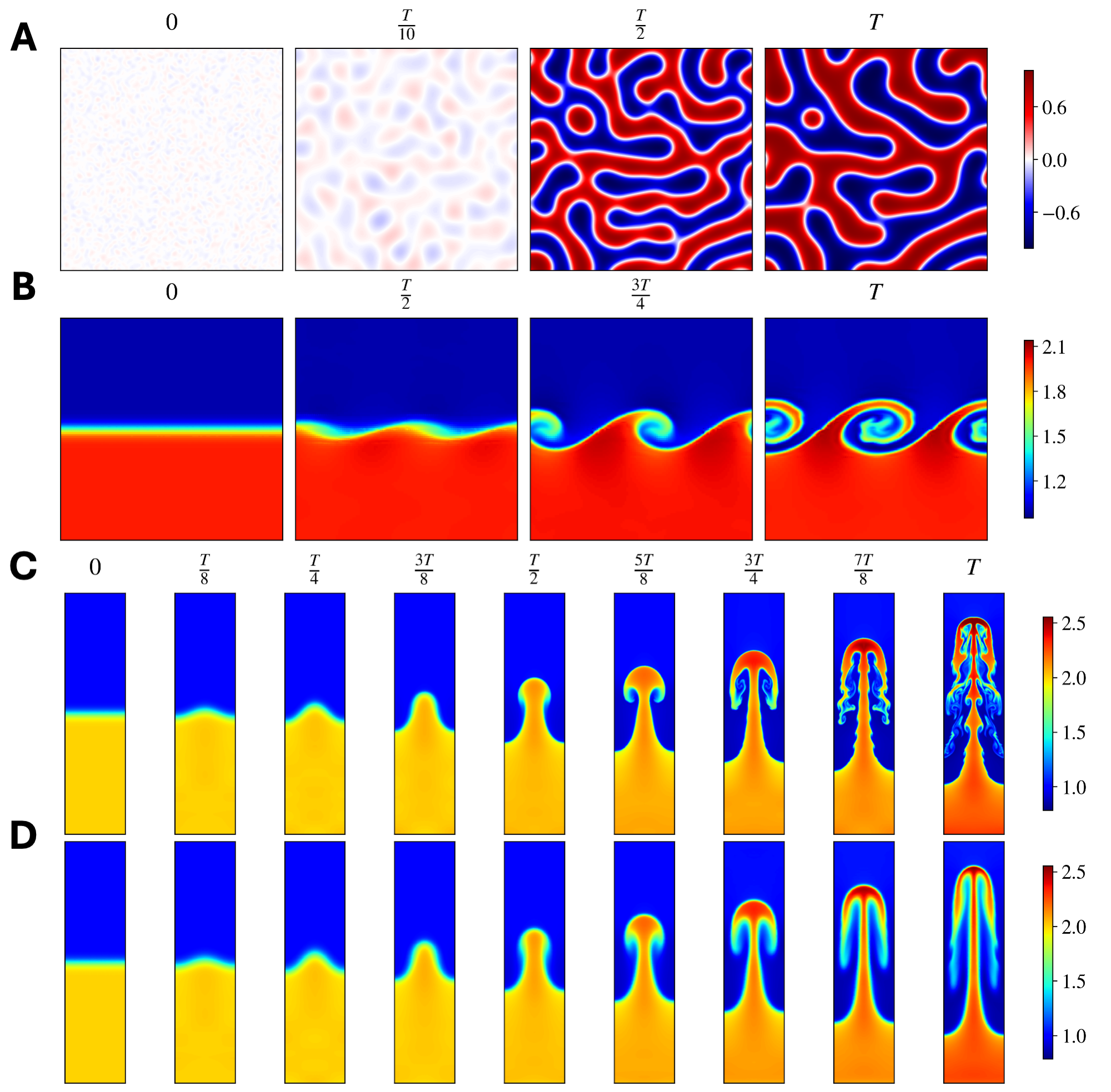}
\caption{ \textbf{Autonomous orchestration of high-order numerical solvers for stiff and complex fluid regimes.} (A) 2D Cahn-Hilliard (Stiff Parabolic). Evolution of spinodal decomposition. The agent diagnosed the stiffness of the biharmonic operator ($\Delta^2 u$) and selected a Fourier Spectral method coupled with Exponential Time Differencing (ETDRK4) to treat the linear term exactly, utilizing the 1/2-rule to prevent aliasing in the cubic nonlinearity while conserving mass to machine precision. (B) Kelvin-Helmholtz Instability (Compressible Euler). The agent deployed a high-order Discontinuous Galerkin Spectral Element Method (DGSEM) to minimize dispersion. Crucially, it autonomously switched the Riemann solver from HLL to HLLC to prevent the smearing of contact discontinuities and attached a density-gradient AMR indicator to resolve the vortex roll-ups. (C) Rayleigh-Taylor Euler (Hyperbolic Balance). Capturing the mixing interface by enforcing hydrostatic balance and utilizing Hennemann-Gassner shock capturing to sharpen the density gradient. (D) Rayleigh-Taylor Navier-Stokes (Viscous). Upon detecting the shift to viscous physics, the agent adjusted the polynomial degree to resolve the viscous shear layers.}
 \label{Numerical_Examples} 
\end{figure}

ATHENA acts as an expert strategist, iteratively refining its approach to prioritize solution fidelity and optimal numerical stability. This autonomous ``self-correction" is best exemplified by the resolution of the 1D Viscous Burgers' equation; while the agent initially derived the exact Cole-Hopf transformation, it diagnosed that the reconstruction step was numerically ill-conditioned for thin shock layers. To prevent truncation error amplification, it autonomously pivoted to a Fourier Pseudospectral method (See Figure~\ref{refine_athena}). Similarly, for the Kelvin-Helmholtz instability, the system identified numerical dissipation and deployed Hennemann–Gassner shock capturing with blended fluxes (DG + FV) and density-gradient AMR to recover sharp features (See Figure~\ref{refine_athena}). This insight extended to the Rayleigh-Taylor instability, where the agent explicitly selected contact-preserving fluxes (HLLC) to resolve sharp interfacial transitions (\ref{Numerical_Examples}). \rev{For these fluid cases, ATHENA autonomously selects and correctly configures established high-order components (e.g., DGSEM, HLLC/HLL fluxes, AMR, and Hennemann–Gassner shock capturing); the underlying discretizations are standard and available in existing libraries such as Trixi.jl.} Complementary pressure and velocity fields for these Rayleigh-Taylor and Kelvin-Helmholtz cases are provided in the Supplementary Information (Figs.~\ref{RT_aux} and~\ref{KH_aux}).

Beyond refinement, the system enforces optimality for complex, stiff operators. For the 2D Cahn-Hilliard equation, the agent identified the  stiffness of the biharmonic operator ($\Delta^2 u$) and the necessity of mass conservation. It selected a Fourier spectral discretization paired with Exponential Time Differencing (ETDRK4) to handle the $k^4$-stiff linear term exactly, while enforcing the $1/2$-rule for dealiasing to prevent spectral blocking (See Figure~\ref{Numerical_Examples}(A)). This strategies enable the system's capacity to solve the full compressible Navier-Stokes equations, demonstrating robust performance in viscous regimes without manual tuning (See Figure~\ref{Numerical_Examples}(D)).

\subsection{ATHENA achieves near machine precision on PIML benchmarks}

\begin{figure}[H]
    \centering
    \includegraphics[width=1\linewidth]{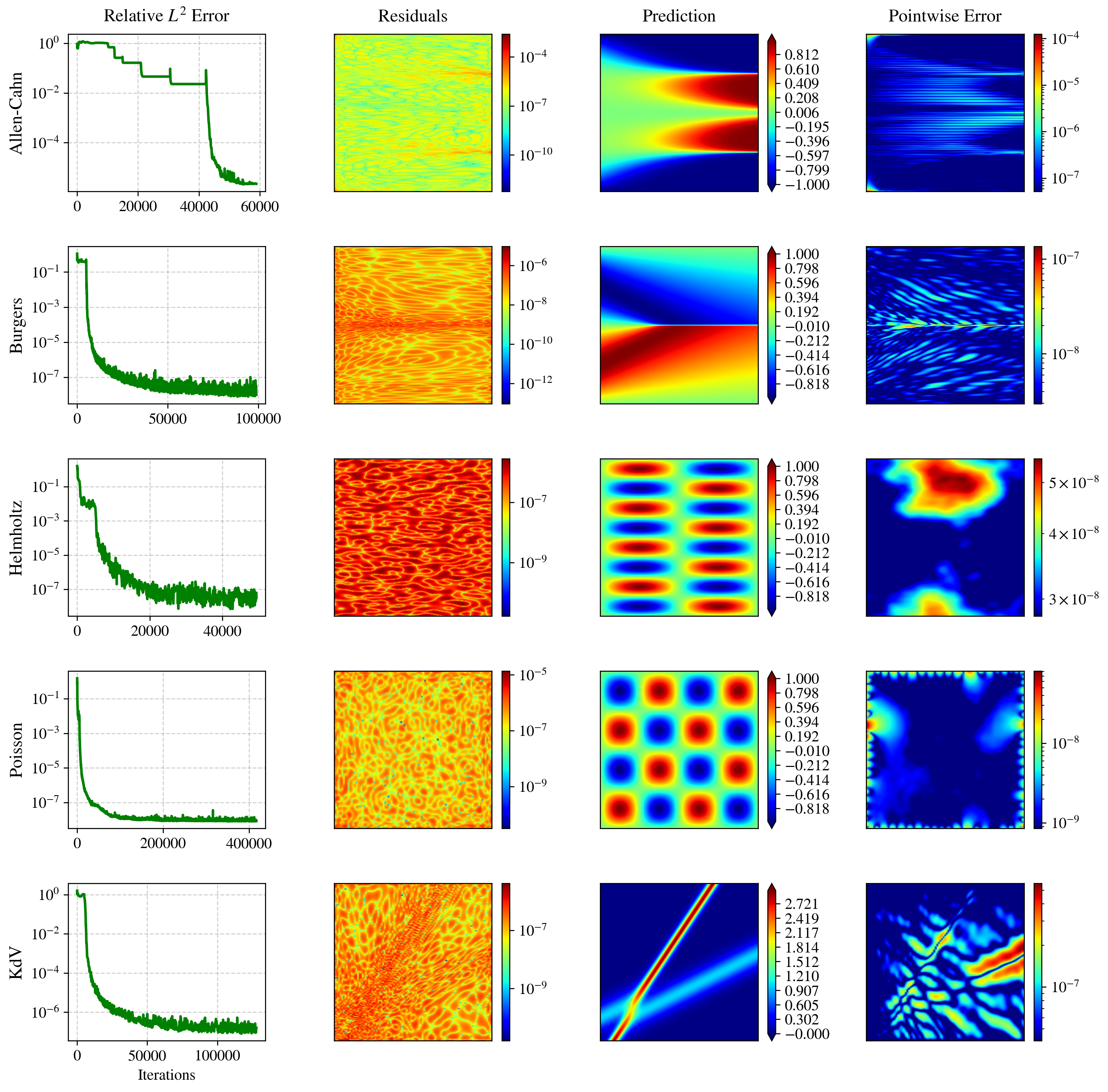}

\caption{
\textbf{ATHENA achieves high-accuracy results on SciML benchmarks.}
The rows correspond to (A) Allen–Cahn, (B) Viscous Burgers, (C) Helmholtz, (D) Poisson, and (E) KdV. The columns display (1) Evolution of the relative $L^2$ error, (2) Final PDE residuals, (3) Model predictions, and (4) Pointwise absolute error distributions.
For Allen–Cahn (A), the stepped convergence profile explicitly visualizes the automated discrete viscosity continuation (homotopy).
For Viscous Burgers (B), residuals reach a stable plateau approaching machine precision ($\sim 10^{-13}$).
In the Helmholtz (C) and Poisson (D) cases, periodic embeddings and Random Fourier Features efficiently capture high-frequency oscillations.
For KdV (E), the second-order optimization reconstructs dispersive interactions without diffusive smoothing. Across all cases, the residuals distribution exhibits a distinct spatial uniformity induced by the vRBA weighting scheme.
Note that the color scales for the pointwise error maps were adjusted to the maximum and median values for better visualization. 
}
\label{fig:pinn_benchmarks}
\end{figure}

To maximize accuracy, the system autonomously differentiates its strategy based on the spectral properties of the governing operator. A primary driver of high precision is the enforcement of boundary conditions through architectural constraints rather than soft penalties. For periodic systems like the Viscous Burgers and Helmholtz equations (See Figure~\ref{fig:pinn_benchmarks}), the agent explicitly selects periodic feature embeddings \citep{dong2021method}. This ensures that boundary conditions are satisfied by construction, eliminating boundary loss terms and significantly reducing the optimization search space.

Simultaneously, the agent adapts the training process to handle stiffness and spectral bias. For the Allen–Cahn equation (See Figure~\ref{fig:pinn_benchmarks} (A)), it identifies sharp-interface dynamics and implements discrete-continuation viscosity—effectively a homotopy method—to stabilize the optimization trajectory. Conversely, for high-frequency oscillatory problems like Poisson and KdV, the system selects Random Fourier Features (RFF). This choice aligns the neural network's frequency, alleviating the spectral bias typical of standard MLPs \citep{Wang2020_Fourier_nets}.

\begin{table}[H]
\small
\centering
\caption{
\textbf{Quantitative comparison against state-of-the-art baselines.}
We evaluate ATHENA across canonical benchmarks, contrasting performance with recent expert-authored publications (Human) and emerging agentic frameworks.
For PIML benchmarks, errors are reported as the relative $L^2$ error $RL_2$ defined in Eq.~\eqref{eq:RL2} and computed on a fixed dense evaluation grid spanning the full solution domain.
ATHENA consistently achieves lower error rates, and for the Viscous Burgers equation, reaches a residual MSE of $2.64 \times 10^{-14}$.
\rev{The benchmark problem is fixed and identical across all rows; each reports a method at its best published configuration against the same reference solution.}
}

\label{tab:sota_comparison}
\begin{tabular}{lllll}
\hline
\textbf{Problem} & \textbf{Reference} & \textbf{Method Origin} & \textbf{Rel.\ $L^2$ Error} & \textbf{MSE} \\
\hline
Allen Cahn
  & \citep{raissi2019physics} & Human & $4.98 \times 10^{-1}$ & -- \\
  & \citep{mcclenny2023self} & Human & $2.10 \times 10^{-2}$ & -- \\
  & \citep{gao2023failure} & Human & $6.50 \times 10^{-3}$ & -- \\
  & \citep{zhang2023dasa} & Human & $8.57 \times 10^{-5}$ & -- \\
  & \citep{toscano2025kkans} & Human & $2.56 \times 10^{-5}$ & -- \\
  & \citep{wang2024piratenets} & Human & $2.24 \times 10^{-5}$ & -- \\
  & \citep{chen2025self} & Human & $1.45 \times 10^{-5}$ & -- \\
  & \citep{wu2025propinn} & Human & $5.62 \times 10^{-1}$ & -- \\
  & \citep{zhao2024pinnsformer} & Human & $5.10 \times 10^{-2}$ & -- \\
  & \citep{xu2025fp64} & Human & $5.45 \times 10^{-2}$ & -- \\
  & \citep{wu2025propinn} & Human & $3.60 \times 10^{-2}$ & -- \\
  & \citep{wang2025gradient} & Human & $3.48 \times 10^{-6}$ & -- \\
  & \citep{urban2024unveiling} & Human & $2.20 \times 10^{-6}$ & -- \\
  & \cite{toscano2025variational} & Human & $2.14 \times 10^{-6}$ & -- \\
  & (Ours) & Agents & $\mathbf{2.07 \times 10^{-6}}$ & $8.11 \times 10^{-11}$ \\
\hline
Burgers ($\nu=\frac{1}{100\pi}$)
  & \citep{zhongkai2024pinnacle} & Human & $1.45 \times 10^{-2}$ & -- \\
  & \citep{gao2023failure} & Human & $5.59 \times 10^{-3}$ & -- \\
  & \citep{raissi2019physics} & Human & $6.7 \times 10^{-4}$ & -- \\
  & \citep{mcclenny2023self} & Human & $4.80 \times 10^{-4}$ & -- \\
  & \citep{chen2025self} & Human & $1.38 \times 10^{-4}$ & -- \\
  & \citep{wang2024piratenets} & Human & $8.20 \times 10^{-5}$ & -- \\
  & \citep{wang2025gradient} & Human & $4.03 \times 10^{-5}$ & -- \\
  & \citep{urban2024unveiling} & Human & $2.90 \times 10^{-8}$ & -- \\
  & \citep{kiyani2025optimizing} & Human & $1.62 \times 10^{-8}$ & -- \\
  & \cite{toscano2025variational} & Human & $1.51 \times 10^{-8}$ & -- \\
  & \citep{wuwu2025pinnsagent} & Agents & -- & $6.51 \times 10^{-5}$ \\
  & \citep{he2025lang} & Agents & -- & $6.48 \times 10^{-5}$ \\
  & (Ours) & Agents & $\mathbf{7.87 \times 10^{-9}}$ & $2.64 \times 10^{-14}$ \\
\hline
Helmholtz
  & \citep{chen2025self} & Human & $4.86 \times 10^{-5}$ & -- \\
  & \citep{urban2024unveiling} & Human & $3.60 \times 10^{-7}$ & -- \\
  & (Ours) & Agents & $\mathbf{7.25 \times 10^{-9}}$ & $7.25 \times 10^{-13}$ \\
\hline
Poisson& \citep{wang2025gradient} & Human & $2.99 \times 10^{-7}$ & -- \\
    & \citep{muller2023achieving} & Human & $1.0 \times 10^{-7}$ & -- \\

  & (Ours) & Agents & $\mathbf{8.13 \times 10^{-9}}$ & $3.95 \times 10^{-13}$ \\

\hline
KdV
  & \citep{wang2024piratenets} & Human & $4.27 \times 10^{-4}$ & -- \\
  & \citep{wang2025gradient} & Human & $3.40 \times 10^{-4}$ & -- \\
  & \citep{urban2024unveiling} & Human & $6.00 \times 10^{-6}$ & -- \\
  & (Ours) & Agents & $\mathbf{7.92 \times 10^{-8}}$ & $1.22 \times 10^{-13}$ \\
\hline
Inviscid Burgers
  & \citep{takamoto2022pdebench} & Human & $1.20 \times 10^{-1}$ & -- \\
  & \citep{kim2021dpm} & Human & $8.30 \times 10^{-2}$ & -- \\
  & (Ours) & Agents & $5.65 \times 10^{-4}$ & $ 6.95\times 10^{-7}$ \\
  & (Ours) & Agents + Human & $\mathbf{3.05 \times 10^{-4}}$ & $1.02 \times 10^{-5}$ \\
\hline
\end{tabular}
\end{table}

As shown in Table \ref{tab:sota_comparison}, these autonomous choices result in substantial accuracy gains. \rev{ATHENA matches and in several cases improves upon the reported accuracy of recent expert-authored benchmarks from top-tier venues (e.g., NeurIPS, ICLR) and emerging agentic frameworks, in some cases by orders of magnitude (Table~\ref{tab:sota_comparison}).} Notably, the system minimizes residuals to near machine precision (MSE of $2.64 \times 10^{-14}$) for the Viscous Burgers equation, demonstrating that dynamic architectural adaptation exceeds the precision limits of static, manually tuned baselines (See Table~\ref{tab:sota_comparison}).

\subsection{Physical diagnosis autonomously detects and repairs ill-posed formulations}

While previous examples demonstrated performance on well-posed problems, we challenged ATHENA with the shock-forming Inviscid Burgers' equation—a problem where standard PINN formulations are mathematically ill-posed due to the lack of a unique entropy solution. The system did not merely optimize a static architecture; it engaged in a 22-iteration evolutionary process (comprising 13 distinct strategic reformulations) to discover a stable solver.

The process began with a fundamental rejection: the Critic Agent identified the direct inviscid formulation as scientifically unsound, forcing the Strategist to implement a Viscous Continuation strategy derived from the scaffolding's 'Reynolds Principle.' As illustrated in Figure~\ref{refine_athena}, which details the residual evolution for three representative iterations and the error history for four, the initial models exhibited significant instability. To address this, the Advisor Agent intervened with a hierarchical series of structural cures. First, it mandated the addition of Entropy Constraints to enforce the physical selection principle, stabilizing the optimization.

However, recognizing that the error remained high due to 'Spectral Deficiency,' the Advisor initially attempted to tune the Fourier embedding degree. When oscillations persisted, it autonomously escalated the intervention, prescribing a fundamental architectural shift from standard MLPs to a Chebyshev-based KAN (Kolmogorov-Arnold Network) to better capture the non-periodic discontinuity. Finally, to bridge the stability gap during the critical vanishing viscosity phase, the system inserted additional annealing substeps, polishing the solution to ensure monotonicity. Figure~\ref{refine_athena} captures how this iterative refinement progressively reduces and stabilizes the error, ultimately decreasing the residual error by more than four orders of magnitude. This autonomous diagnosis and repair process yielded a final error of $5.6 \times 10^{-4}$—two orders of magnitude lower than standard human benchmarks (See Table~\ref{tab:sota_comparison}).

\subsection{Symbiotic orchestration: Human-in-the-loop refinement and hybrid symbolic-numeric workflows}

\begin{figure}[H]
    \centering
    \includegraphics[width=0.9\linewidth]{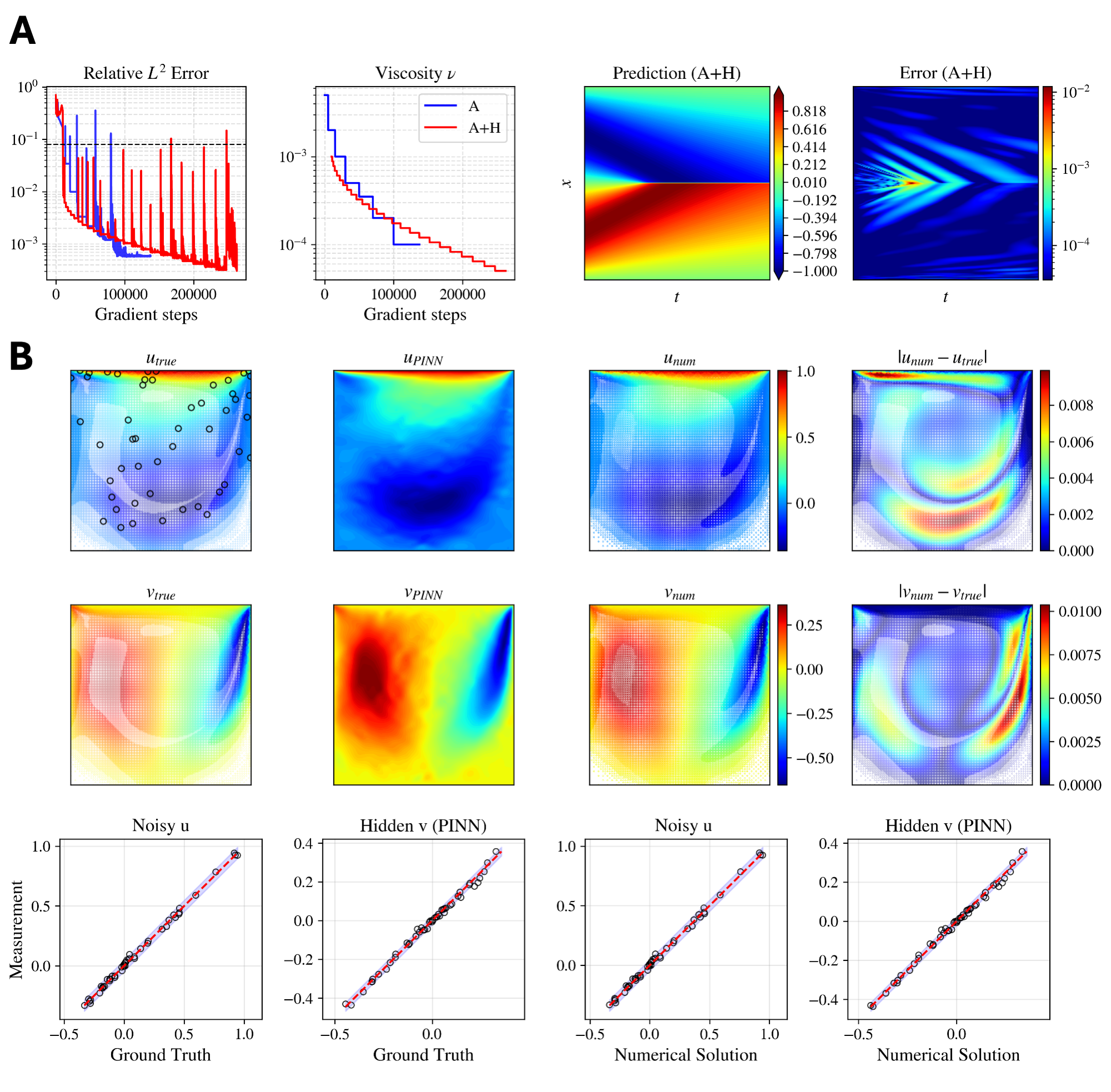}
\caption{\textbf{Symbiotic orchestration facilitates superior convergence in stiffness-dominated and inverse regimes.} (A) Inviscid Burgers equation. The autonomous agent (blue, A) employs a standard viscous continuation schedule but stalls at $\nu = 10^{-4}$ due to spectral deficiencies. A human intervention (red, A+H) suggesting periodic wavelet embeddings and a refined dwelling schedule enables the system to extend shock-capturing capabilities to $\nu = 5 \times 10^{-5}$, reducing the relative $L^2$ error by an order of magnitude. (B) Closed-loop AI velocimetry for lid-driven cavity flow at $Re=500$. The system receives sparse observations of the $u$-velocity component corrupted by $5\%$ Gaussian noise (black markers). Phase 1: A PINN infers the hidden Reynolds number ($Re$), vertical velocity $v$, and pressure, yielding noisy reconstructions ($u_{PINN}$, $v_{PINN}$) with $\approx 6\%$ relative error and an inferred $Re=525$. Phase 2: The agents autonomously instantiate a rigorous FEM solver using the inferred parameter to verify the physics. This step filters the aleatoric noise, recovering smooth fields ($u_{num}$, $v_{num}$) and reducing the error to $\approx 1\%$ (rightmost column). The bottom scatter plots demonstrate that the discrepancies fall within the $5\%$ noise floor, making the exact recovery of $Re=500$ statistically impossible; the inferred $Re=525$ thus represents the optimal solution within the observational uncertainty.}
\label{symbiosis}
\end{figure}

While ATHENA is designed for fully autonomous \rev{operation}, its architecture \rev{also} supports a collaborative modality where expert intuition works in symbiosis with the agentic workflow. Crucially, this interaction is streamlined: the user provides a single sentence in the run request, which the agents autonomously interpret and integrate into the computational graph. For instance, in the Viscous Burgers regime, the agents successfully identified the Cole-Hopf transformation to obtain exact solutions, yet it failed due to standard quadrature limitations. A human intervention suggesting a superior quadrature scheme (Gauss-Hermite) enabled the system to resolve the integral to machine precision (See Figure~\ref{refine_athena} (B)). Similarly, for the Inviscid Burgers equation, the agents autonomously implemented viscous continuation and entropy conditions but failed to converge below a viscosity of $\nu = 10^{-4}$ due to spectral deficiency, as the periodic Fourier embeddings induced high-frequency oscillations. A human suggestion to utilize periodic wavelet embeddings combined with a more refined viscosity schedule allowed the agents to mitigate these oscillations, reducing the error and extending shock-capturing capabilities to $\nu = 5 \times 10^{-5}$ (See Figure~\ref{symbiosis}(A)). 

Beyond individual solver refinement, the conceptualization team orchestrates hybrid workflows that bridge data-driven inference and rigorous simulation. In a closed-loop AI Velocimetry experiment, this process was executed in two distinct phases. First, the agents ingested noisy measurements and deployed a SciML model (PINN) to infer the Reynolds number and reconstruct the remaining flow fields (See Figure~\ref{symbiosis} (B) ). 

Second, utilizing these inferred parameters, the team autonomously instantiated a SciC solver (FEM) to verify the physics. This symbiotic loop reduced the relative error from approximately $6\%$ (inference phase) to $1\%$ (verification phase). Furthermore, users can explicitly direct this orchestration for validation tasks; for instance, the ground truth data used to validate the Inviscid Burgers equation was generated by explicitly tasking the SciC agent group to compute the exact solution prior to the learning phase .

\subsection{\rev{Generative reasoning identifies diverse strategies and competitive neural operators}}

\begin{figure}[H]
    \centering
    \includegraphics[width=1\linewidth]{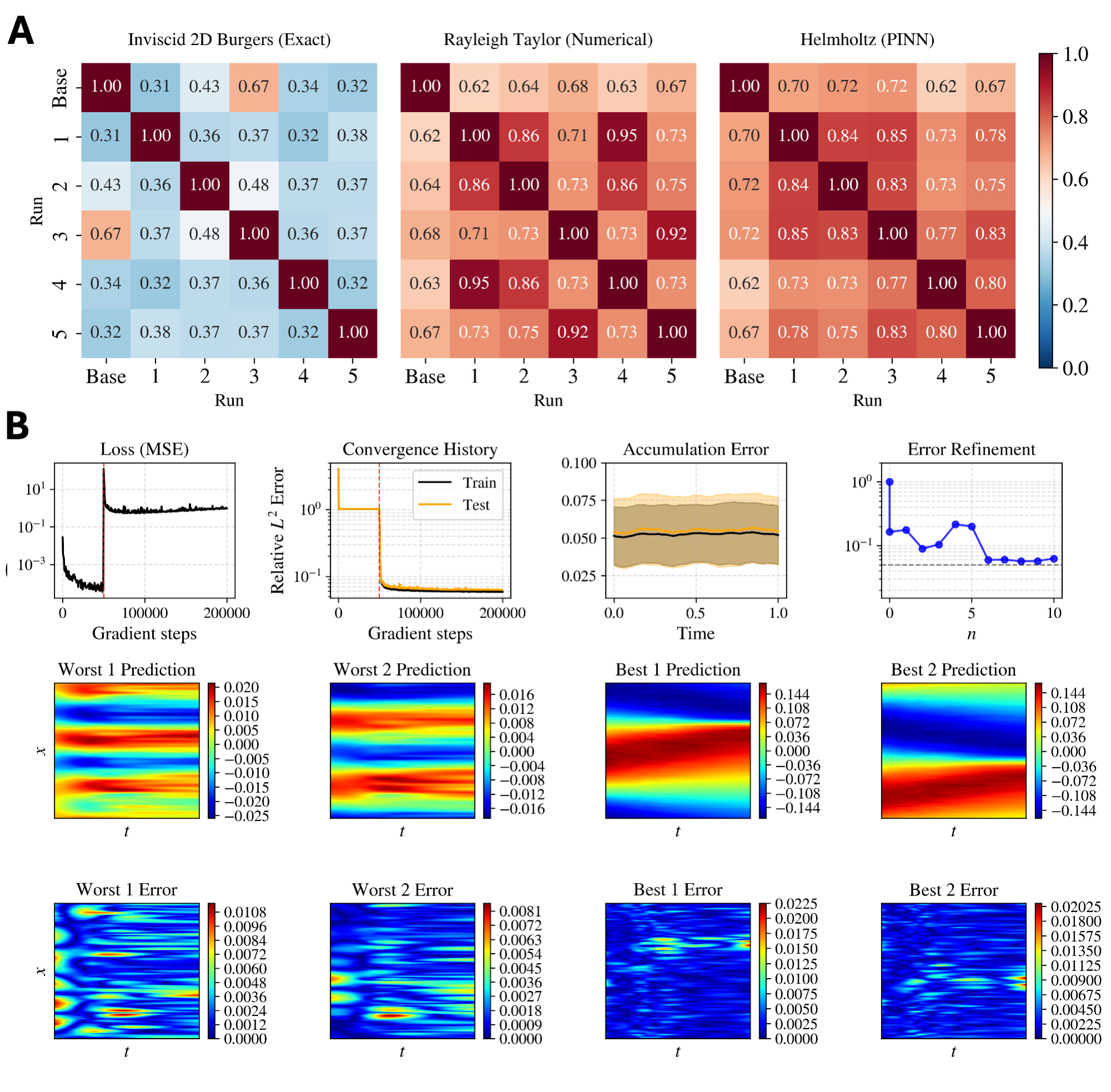}
\caption{\textbf{\rev{Generative reasoning enables structural adaptability and competitive neural operator selection.}} (A) Code similarity matrices quantify the deviation between a generic reference template (Base) and five independent solutions generated by ATHENA. For the Inviscid Burgers equation (left), the agent demonstrates significant architectural adaptation ($\approx 30\%$ similarity), substantially modifying the template to implement the specialized method of characteristics. In contrast, for Rayleigh-Taylor (center) and Helmholtz (right) tasks, the agent retains more of the scaffold ($\approx 60\%$ similarity). \rev{Crucially, the low inter-run similarity shows that ATHENA produces diverse implementation paths that are structurally distinct yet functionally equivalent.} (B) Performance of the \rev{best-performing architecture among the configurations searched} (QR-DeepONet with KKAN trunk and RBF basis). The top row illustrates the training dynamics, where the vertical red dashed line demarcates the two-stage training process. The ``Error Refinement'' panel tracks the systematic reduction of the relative $L^2$ error across the 11 evolutionary iterations, while the ``Accumulation Error'' plot confirms that the final architecture exhibits negligible temporal error propagation. The middle and bottom rows display the prediction, reference, and pointwise error for the worst-case (Sample 432) and best-case (Sample 888) inputs. Even in the worst-case scenario, the model captures the dominant shock dynamics, confirming that the remaining error ($5.67\%$) is structural rather than a failure of generalization.}
\label{Variability_analysis}
\end{figure}

ATHENA’s conceptual scaffolding functions as a navigational guide rather than a restrictive template; it directs the attention of Large Language Models toward a vast combinatorial search space without limiting their reliance on extensive training knowledge. To quantify this behavior, we conducted an experiment on three representative benchmarks—2D Inviscid Burgers, Rayleigh-Taylor instability, and a Helmholtz PINN—executing the code generation pipeline five times for each case using identical prompts. While the system achieved a perfect evaluation score across all runs, the generated solutions were not duplicates. We quantified this diversity through code difference analysis, measuring the percentage of altered lines relative to a generic base template. As illustrated in Figure \ref{Variability_analysis} (A), the agents exhibited architectural adaptation, particularly in the Inviscid Burgers regime, where the code similarity dropped to approximately $30\%$, reflecting a substantial departure from standard boilerplate to implement specialized solvers.

\begin{table}[H]
\small
\centering
\caption{
\textbf{Generative diversity in autonomous solution strategies for the Inviscid Burgers' benchmark.}
While all five ATHENA runs achieved a perfect total score by correctly identifying the Method of Characteristics (MoC) as the optimal solver, the mathematical formulation for enforcing the \textit{Entropy Condition} (shock handling) varied significantly. The agent autonomously discovered three distinct approaches: Variational Calculus (Run 1), Geometric Shock Fitting (Runs 2 \& 3), and Domain-Specific Bracketing (Runs 4 \& 5).
}
\label{tab:burgers_diversity}
\begin{tabular}{lllll}
\hline
\textbf{Run} & \textbf{Reduction} & \textbf{Entropy Strategy} & \textbf{Implementation Logic} & \textbf{Grid} \\
\hline
1 & $x+y$ & Lax-Oleinik Principle & Variational Minimization of $J(s)$ & $200 \times 200$ \\
2 & $x+y$ & Geometric Shock Fitting & Explicit Tracking \& Equal-Area Rule & $128 \times 128$ \\
3 & $(x+y)/2$ & Domain Decomposition & Explicit Splitting at $\xi_{shock}(t)$ & $256 \times 256$ \\
4 & $(x+y)/2$ & Interval Bracketing & Constrained Bisection on $[0,\pi]$ & $128 \times 128$ \\
5 & $x+y$ & Local Bracketing & Implicit Root Finding in $(s-5, s+5)$ & $128 \times 128$ \\
\hline
\end{tabular}
\end{table}

Beyond syntactic variation, the system proposed diverse mathematical strategies. In the Inviscid Burgers benchmark, the core challenge lies in resolving the discontinuity after shock formation. Across the five independent runs, the agents autonomously derived three distinct yet mathematically equivalent entropy strategies to handle this singularity: Variational Calculus (Lax-Oleinik), Geometric Shock Fitting (Equal-Area Rule), and Domain-Specific Bracketing (Table \ref{tab:burgers_diversity}). This suggests that the agents are not merely retrieving a single ``correct" answer from their training data but are actively reasoning through the problem to construct valid mathematical alternatives. Detailed comparison tables analyzing the strategic variations for the Rayleigh-Taylor and Helmholtz benchmarks are provided in the Supplementary Information.

\begin{table}[H]
\small
\centering
\caption{
\textbf{Architectural exploration history for the operator-learning benchmark.}
This table reports the trial log reviewed by the Advisor Agent during the meta-reasoning phase. Across 11 configurations—spanning Standard DeepONets, QR-DeepONets, and an SVD-based variant—the agent observed a performance plateau and selected Run~9 as the structural optimum for final refinement. The autonomously identified optimum ($5.67\%$) is competitive with the $5.70\%$ reported in \cite{toscano2025kkans} and approaches the $3.4\%$ achieved by specialized physics-informed solvers~\cite{chen2024self}.
}
\label{tab:operator_history}
\begin{tabular}{lll}
\hline
\textbf{Run} & \textbf{Architecture Family} & \textbf{Rel.\ $L^2$ Error} \\
\hline
1  & Standard DeepONet (MLP + Periodic Fourier)              & 16.37\% \\
2  & Standard DeepONet (KAN Trunk -- Sin Basis)              & 17.60\% \\
3  & Standard DeepONet (KAN Trunk -- Wavelet Basis)          & 8.98\%  \\
4  & Standard DeepONet (KKAN Trunk -- Default)               & 10.40\% \\
5  & Standard DeepONet (KKAN Trunk -- Fourier Basis)         & 21.50\% \\
6  & Standard DeepONet (KAN Trunk -- Wavelet Degree 5)       & 20.02\% \\
7  & QR-DeepONet (MLP Trunk)                                 & 5.95\%  \\
8  & QR-DeepONet (MLP + Periodic Fourier)                    & 6.02\%  \\
\textbf{9}  & \textbf{QR-DeepONet (KKAN Trunk -- RBF Basis)}     & \textbf{5.67\%} \\
10 & QR-DeepONet (KKAN Trunk -- Fourier Basis)               & 5.72\%  \\
11 & SVD-DeepONet (KKAN Trunk -- RBF Basis)                  & 6.27\%  \\
\hline
\end{tabular}
\end{table}

This generative capability extends to the design of neural operators for high-Reynolds number flows ($\nu = 10^{-3}$). Rather than relying on static parameter tuning, the system performed an evolutionary architectural search, analyzing the trade-offs between 11 distinct configurations (Table \ref{tab:operator_history}). The agents explored a wide spectrum of inductive biases, ranging from periodic Fourier embeddings and wavelet-based KANs to rigorous matrix decompositions (QR and SVD). While the search highlighted the strong baseline performance of wavelet-enhanced architectures, the system ultimately converged on a QR-decomposed DeepONet with a Kurkova-KAN (KKAN) trunk using Radial Basis Functions (Run 9). This autonomously identified architecture achieved a relative error of $5.67\%$, exhibiting minimal error accumulation and consistent performance across both training and testing regimes (See Figure~\ref{Variability_analysis}(B)).

\subsection{\rev{Conceptual scaffolding induces efficient, sub-linear-like reward improvement across diverse scientific tasks}}

\begin{figure}[H]
    \centering
    \includegraphics[width=1\linewidth]{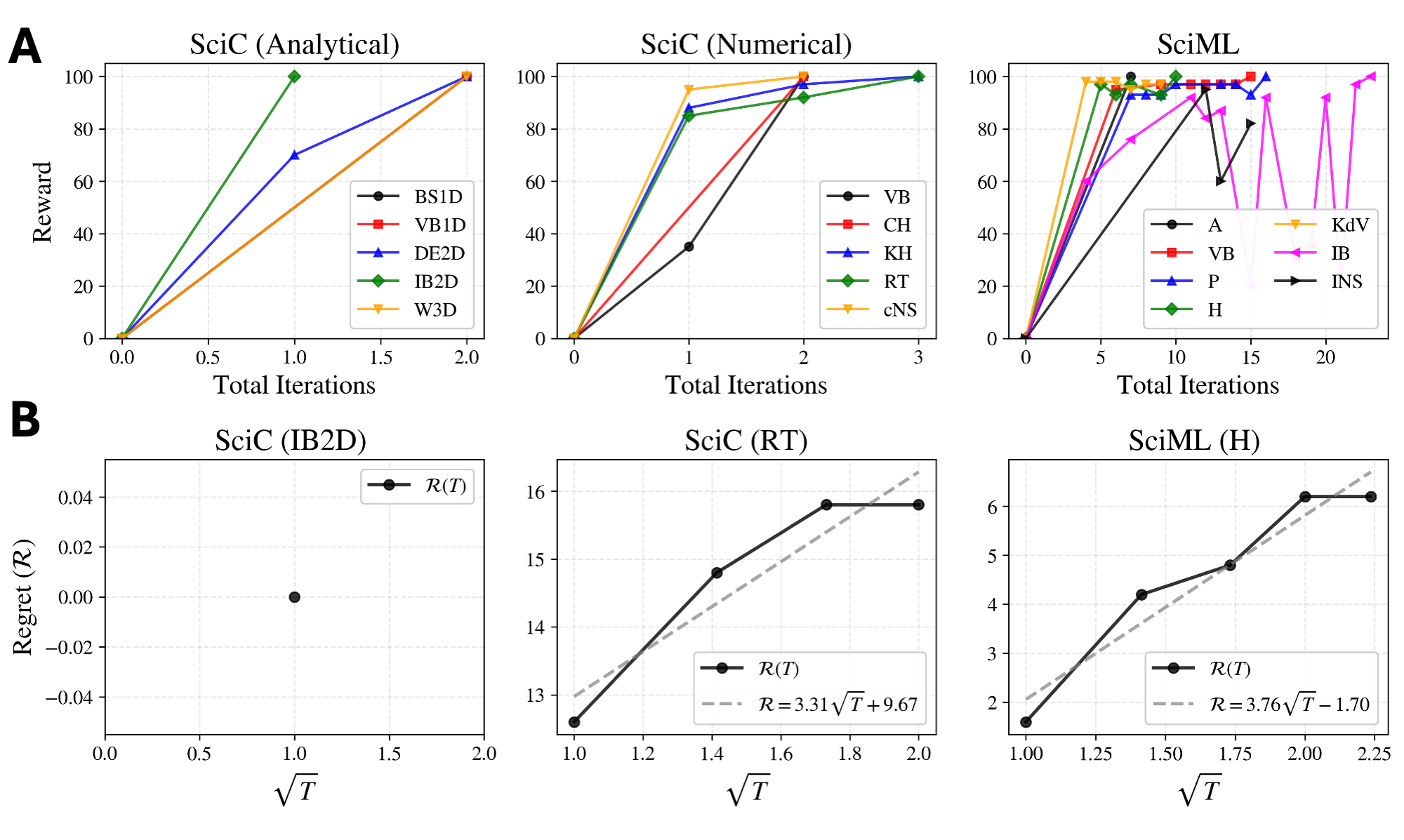}
\caption{\textbf{Reward dynamics and regret scaling demonstrate efficient autonomous optimization.}
\textbf{Top Row:} Evolution of the evaluation score (Reward) as a function of Total Iterations. Note that the majority of these steps are \textit{debugging iterations} required to stabilize a specific approach, with only a subset representing distinct \textit{strategy iterations}. \rev{The total-iteration axis therefore also reports the per-task retry/debugging count (the number of execution-repair cycles before a successful run), providing a direct measure of implementation cost.}
\textit{SciC (Analytical):} Convergence is immediate; once the agent identifies the existence of an exact solution, the implementation is syntactically straightforward, requiring minimal debugging.
\textit{SciC (Numerical):} Convergence scales with physical complexity. Tasks aligned with the base template (e.g., Incompressible Navier-Stokes, INS) are solved instantly. Conversely, multiscale problems (e.g., Rayleigh-Taylor, RT; Kelvin-Helmholtz, KH) require extended debugging sequences to resolve shock transitions and fine features.
\textit{SciML:} Optimization trajectories are longer (10--23 iterations) due to the stochastic nature of training. The transient drops in reward (e.g., Inviscid Burgers, IB) reflect aggressive exploration of new strategies that, while initially unstable, are rapidly corrected by historical memory, resulting in a trajectory that heuristically approximates a submartingale.
\textbf{Bottom Row:} Regret analysis ($R(T)$) plotted against the square root of \textit{Strategy Iterations} ($\sqrt{T}$), which represent high-level formulation changes.
\textit{Left:} Analytical cases exhibit zero regret as the optimal strategy is identified at initialization.
\textit{Center \& Right:} For complex Numerical and SciML tasks, the regret grows linearly with $\sqrt{T}$, indicating that the cumulative regret scales as $R(T) \propto \sqrt{T}$. \rev{This sublinear, efficient improvement of the reward over strategy iterations reflects the contextual-bandit formulation that inspires ATHENA's design, in which the conceptual scaffolding balances exploration and exploitation.}}
\label{reward_analysis}
\end{figure}

ATHENA's agentic architecture is formally formulated as a conceptual bandit, where the optimization policy is implicitly encoded via conceptual scaffolding. This scaffolding—instantiated through rigid code templates and system prompts containing distilled expert knowledge—serves to constrain the combinatorial search space effectively, inducing a trajectory that heuristically approximates a submartingale on the reward process (Figure~\ref{reward_analysis}, Top Row). For tasks closely aligned with the encoded priors, such as the Incompressible Navier-Stokes equation, this guidance enables immediate convergence. (The detailed descriptions of these templates are provided in the Supplementary Information, while the codebase is available in the associated repository. Conversely, for complex numerical regimes like Rayleigh-Taylor (RT) and Kelvin-Helmholtz (KH) instabilities, the agents require extended iterative cycles to refine shock transitions and capture fine-scale features. In stochastic SciML tasks such as the Inviscid Burgers equation, the reward trajectory exhibits transient dips. These fluctuations do not indicate failure but rather aggressive exploration, where ATHENA actively tests diverse hypothesis classes before stabilizing on an optimal strategy using its historical memory.

To quantify the efficiency of these learning dynamics, we analyzed the cumulative regret across the full experimental suite (Figure~\ref{reward_analysis}, Bottom Row). By plotting the regret $R(T)$ against the square root of strategy iterations $\sqrt{T}$, we observe a distinct linear relationship for the complex numerical and SciML tasks. \rev{This shows sublinear improvement of the reward over strategy iterations---empirical evidence of efficient, guided search. This dynamic is \emph{inspired by} the contextual-bandit formulation of the research lifecycle, which models the search as a knowledge-driven policy that balances the exploration of new strategies against the exploitation of those already working.} While analytical cases show zero regret due to immediate solution identification, the sub-linear growth observed in multiscale problems verifies that the system successfully balances the exploitation of known mathematical motifs with the necessary exploration of novel solver structures.

\section{Discussion}

Inspired by the hypothesis that the scientific research lifecycle functions as a Contextual Bandit process, we designed ATHENA to explicitly operationalize this framework. In this view, discovery proceeds through discrete, high-stakes decisions where an agent selects a structural hypothesis based on diagnostic evidence to maximize scientific fidelity. We prioritize this formulation over full Reinforcement Learning (RL) to address the prohibitive cost of scientific evaluation; unlike environments where trial-and-error is inexpensive, each scientific iteration requires code generation, computationally intensive execution, and rigorous validation. Crucially, this formulation enables us to treat the action space as discrete and finite, rendering the combinatorial search tractable. \rev{Furthermore, the Inspector's plan-compliance check keeps each realized code state $S_n$ aligned with the requested action $A_n$ regardless of the specific prior implementation $S_{n-1}$; this decouples the structural strategy from code generation and lets the framework treat implementation variations as stochastic fluctuations. Empirically, the reward trajectories show sublinear, submartingale-like improvement (Figure~\ref{reward_analysis})---the efficient, guided search that the contextual-bandit formulation inspiring ATHENA's design predicts.} The architectural organization of ATHENA directly mirrors this theoretical structure; the formal mathematical definition of these operators is detailed in the Methods section.

A central design choice in ATHENA is the implementation of the optimization policy $\pi$ as a fixed, knowledge-driven structure rather than a learned parameter set. Unlike Reinforcement Learning, where the agent must discover transition dynamics from scratch, our policy is instantiated through 'Conceptual Scaffolding'—a compact set of expert-derived principles encoded into the code templates and system prompts of the Strategy and Advisor teams. This scaffolding acts as a navigational guide, constraining the combinatorial search space to scientifically valid regions and enabling the discovery of optimal solutions that unconstrained models fail to find. The necessity of this approach is demonstrated in our ablation analysis (Supplementary Figures~\ref{ablation_SiC} and~\ref{ablation_sciML}). For analytical tasks like the Inviscid Burgers equation, the scaffolding guided the agents to identify the problem's symmetry and retrieve the exact Method of Characteristics, whereas baselines produced oscillatory pseudo-spectral solutions or diffusive finite-volume approximations characterized by artifactual ``yellow lines" at shock interfaces. Similarly, in complex regimes like the Rayleigh-Taylor instability, the scaffold enabled the deployment of a 5th-order Discontinuous Galerkin Spectral Element Method (DGSEM) with adaptive mesh refinement. In contrast, the baseline agents, forced to generate extensive monolithic code, succumbed to the ``Lost in the Middle" failure mode; the overwhelming context length led them to overlook a critical sign in the perturbation, resulting in a physically inverted bubble alongside the smearing typical of low-order schemes. This advantage is most pronounced in Scientific Machine Learning tasks involving ill-posed problems, such as the Inviscid Burgers benchmark. While unconstrained agents attempted to implement advanced strategies found in the literature—such as causality and adaptive activation functions—these implementations were often hallucinated or inconsistent with established mathematical definitions \citep{lu2021deepxde,jagtap2020adaptive}, resulting in bloated models with nearly 300,000 parameters and a failure-level error of 19\%. ATHENA, conversely, leveraged the scaffolding to diagnose the ill-posedness inherent to hyperbolic conservation laws. Crucially, the agent explicitly added an entropy constraint to the loss formulation to enforce uniqueness, and simultaneously employed a viscous continuation strategy to stabilize the shock capture. This principled combination prioritized mathematical admissibility over raw residual minimization, reducing the relative $L^2$ error to $5.6 \times 10^{-4}$ with a parsimonious architecture of only 3,416 parameters, demonstrating that constrained reasoning effectively prevents both the context-induced errors and the hallucinated complexity typical of unguided foundation models. \rev{Our ablations isolate the most consequential factor---the conceptual scaffolding---by comparing the full system against both a direct-prompting baseline and a no-scaffolding multi-agent baseline (Supplementary Figures~\ref{ablation_SiC},~\ref{ablation_sciML}). This is the meaningful contrast: the remaining components are functionally integral to the loop---the advisor provides the feedback signal that drives refinement, and the visual diagnostics are what detect spectral bias and oscillations---so the scaffolding comparison directly measures the effect of the framework's central design choice.}

Quantitatively, the effectiveness of this guided evolution is indicated by the reward dynamics presented in Figure~\ref{reward_analysis}. ATHENA’s trajectory heuristically approximates a submartingale ($\mathbb{E}[R_{n+1} \mid \mathcal{H}_n] \ge R_n$), suggesting that the system systematically converges toward higher-fidelity solutions. This behavior diverges from the ``No Scaffolding" baseline (Supplementary Figure~\ref{ablation_sciML}(C)), which, particularly in complex regimes like the SciML Inviscid Burgers task, exhibits random wandering behavior where the reward fails to establish an increasing trend. While one might assume that such guidance reduces the framework to a rigid template-matching engine, our variability analysis implies the opposite: the scaffolding appears to function as a navigational guide rather than a restrictive mold. As quantified by the code similarity matrices in Figure~\ref{Variability_analysis}(A), the agents exhibit significant architectural adaptation, retaining only $\approx 30\%$ of the base template for the Inviscid Burgers equation as they restructure the code to implement specialized solvers. Furthermore, rather than limiting the model, this constraint appears to channel the vast training knowledge of the Large Language Model into diverse mathematical realizations. As detailed in Table~\ref{tab:burgers_diversity}, across independent runs, the agents autonomously derived three mathematically distinct but equivalent entropy strategies—Variational Calculus (Lax-Oleinik), Geometric Shock Fitting, and Domain-Specific Bracketing. This suggests that the system is not merely retrieving rigid templates but is effectively leveraging its internal knowledge base to construct diverse, valid solutions. \rev{The conceptual scaffolding is also what enables the system to recover and correctly implement these methods: when the same base models are run \emph{without} it, they fail to identify or apply the correct method (Supplementary Figures~\ref{ablation_SiC},~\ref{ablation_sciML}). The scaffolding-guided reasoning and execution-feedback loop, not pretrained recall alone, is what drives the results.}

We emphasize, however, that the conceptual scaffolding employed here is not intended as an exhaustive taxonomy of scientific methods, nor as a prescriptive enumeration of uniquely optimal techniques. Rather, these choices serve to anchor agentic reasoning in compact, theory-grounded principles that are widely applicable across problem classes. This perspective positions ATHENA not merely as a solver for specific equations, but as a methodological layer above individual models—an ``agentic computational laboratory" that targets the structure of computational reasoning itself. Although demonstrated here on SciC and SciML tasks, the HENA loop’s separation of conceptual policy from numerical realization generalizes naturally to broader scientific challenges, including automated consistency checks for discretizations, experimental design, and the adaptive coupling of heterogeneous simulators in multiphysics pipelines. Crucially, ATHENA achieves this generalization without reliance on large task-specific datasets or exhaustive evolutionary search, distinguishing it from AutoML systems or AlphaEvolve-style approaches~\cite{georgiev2025mathematical} that operate within fixed optimization templates. Instead of merely asking \textit{how} to optimize a predefined structure, ATHENA reasons about \textit{what} computation should be performed, treating the discovery process itself as an object of agentic diagnosis, validation, and reformulation.

Viewed more broadly, ATHENA represents a step toward agentic computational methods as a methodological layer above individual solvers, models, and scientific domains. By elevating computation itself to an object of agentic reasoning—subject to diagnosis, validation, and reformulation—the framework points toward \rev{a workflow in which agentic systems reduce the implementation burden that often limits discovery}, enabling human experts to focus on conceptual insight, methodological innovation, and scientific interpretation.

\section{Methods}\label{sec11}

\subsection{Online Learning and HENA}

We model the scientific workflow using a Contextual Bandit formulation \cite[Chapter~18]{lattimore2020bandit}. At iteration $n$, the system has access to the history $\mathcal{H}_n = (A_1, O_1, R_1, \dots, A_n, O_n, R_n)$ and selects the subsequent action through the stochastic policy
\[
A_{n+1} \sim \pi(A_{n+1} \mid \mathcal{H}_n).
\]
This action is translated into an executable code state via the implementation operator
\[
S_n \sim \mathcal{I}(S_{n-1}, A_n),
\]
which is executed to produce observations
\[
O_n \sim E(O_n \mid S_n).
\]
The Advisor agent processes these results to assign a scalar reward $R_n = R(A_n, O_n)$. The performance of the policy $\pi$ is evaluated through the regret,
\[
\mathrm{Regret}_T(\pi) = \mathbb{E}_\pi \left[ \sum_{t=1}^T (R^* - R_t) \right],
\]
where $R^*$ denotes the expected reward of the best fixed action in retrospect. Our goal is not to learn the policy but to design one that is sample-efficient. Accordingly, we require that the expected rewards form a submartingale,
\[
\mathbb{E}[R_{n+1} \mid \mathcal{H}_n] \ge R_n,
\]
ensuring that the sequence improves on average. In this construction we do not treat the code state as a latent variable; variations in implementation are regarded as stochastic fluctuations rather than a conceptual component of the bandit process.

To satisfy the finiteness requirement of the bandit model while maintaining expressive capacity, the action space is constructed as a discrete combinatorial set. In the context of Physics-Informed Machine Learning, for example, we decompose the choices into representation, constraint formulation, and optimization strategy, yielding
\[
\mathcal{A} = \mathcal{A}_{\mathrm{rep}} \times \mathcal{A}_{\mathrm{constraint}} \times \mathcal{A}_{\mathrm{opt}}.
\]
This modular structure allows the system to explore meaningful alternatives while keeping the decision space tractable.

The Conceptualization Teams implement the policy $\pi$. Their prompting structure is designed to balance the exploitation of previously successful strategies with the exploration of alternative options when stagnation is detected. This prevents premature convergence to suboptimal solutions and preserves the stochastic behavior required in a bandit formulation.

The operator $\mathcal{I}$ is realized through a coordinated set of coding agents that produce and refine the implementation corresponding to the selected action. \rev{The Inspector agent enforces \emph{plan compliance}: it verifies that the realized code state $S_n$ faithfully implements the requested action $A_n$, regardless of the specific prior implementation $S_{n-1}$. This plan-fidelity check keeps each realized state aligned with the requested structural change, decoupling the strategy from the particulars of code generation.} If inconsistencies arise, the implementation cycle is repeated until a consistent state is reached.

Execution of the code state yields observations consisting of numerical fields, residual curves, stability metrics, and other diagnostic quantities relevant to the evaluation of scientific correctness. These observations are summarized by the Advisor and used to determine the reward. To guarantee that the optimal reward is attainable and that the regret is well-defined, we construct a bounded composite reward
\[
R_n = R_{\mathrm{integrity}} + R_{\mathrm{accuracy}} + R_{\mathrm{details}} + R_{\mathrm{optimality}}.
\]
The integrity term $R_{\mathrm{integrity}}\in[0,50]$ evaluates the executability and structural alignment of the implementation with the requested action. The accuracy term $R_{\mathrm{accuracy}}\in[0,20]$ is decomposed into a precision component, which rewards solutions below a tolerance $\varepsilon$, and a consistency component, which evaluates adherence to physical constraints. The details term $R_{\mathrm{details}}\in[0,15]$ penalizes localized artifacts such as oscillations or unstable modes. The optimality term $R_{\mathrm{optimality}}\in[0,15]$ measures computational efficiency. By imposing these upper bounds on each contribution, the per-step reward is bounded ($R_n\in[0,100]$), and the maximal reward is achievable under realistic numerical conditions, enabling vanishing regret.

\subsection{Conceptual Scaffolding}

To constrain the policy to scientifically meaningful reasoning, we introduce a small set of expert-derived conceptual scaffolds drawn from approximation theory \cite{cybenko1989approximation,hornik1989multilayer,Kolmogorov1957,song2025explicit}, variational formulations \cite{toscano2025variational}, and standard numerical analysis. These scaffolds restrict the decision-making process to a principled subset of actions, reducing reliance on long prompts and extensive technical documentation, and mitigating the degeneracy typically observed in unconstrained LLM-driven exploration.

\subsubsection{Universal Approximation Blueprint}

A primary principle in this scaffold is the classical universal approximation viewpoint, which states that a continuous target function $f$ on a compact domain can be approximated by a finite linear combination of basis functions,
\[
f(x) \approx \sum_{i=1}^m c_i\,\phi_i(x).
\]
Density in this context refers to the ability of the span of $\{\phi_i\}_{i=1}^m$ to approximate $f$ arbitrarily well as $m$ increases. This representation captures the essence of many architectures used in scientific machine learning: the modeling task consists of selecting an appropriate basis $\{\phi_i\}$ and determining coefficients $\{c_i\}$ that reconstruct the target function. This viewpoint applies directly to single-layer networks, and extends naturally to multilayer constructions. In the case of MLPs, one may interpret the functions $\phi_i$ as the outputs of the last hidden layer and the coefficients $c_i$ as the weights of the final linear projection, aligning with the adaptive basis interpretation of neural networks.

This approximation principle also unifies the design of alternative backbone models. Architectures such as Kolmogorov–Arnold Networks (KANs) attempt to follow the structure suggested by the Kolmogorov–Arnold representation theorem, which asserts that any multivariate continuous function can be represented as a finite superposition of univariate functions and addition. Likewise, KKANs are motivated by the Kůrková–Kolmogorov approximation theorem, which provides a refined, ridge-function–based representation for continuous functions and suggests structural forms that may be advantageous for high-dimensional approximation. Within this blueprint, these networks are viewed not as ad hoc architectural variants but as different choices of basis families $\{\phi_i\}$ derived from distinct theoretical representation results.

The same approximation viewpoint extends to operator learning architectures. Models such as DeepONets explicitly follow the factorization above: the branch network learns the coefficients $c_i$ and the trunk network learns the basis functions $\phi_i$. From this perspective, improving an operator network amounts to modifying either the basis or the coefficient map. Replacing a standard trunk with an orthogonalized or QR-based construction yields a more stable and better conditioned basis, while SVD-based variants provide an alternative factorization that mitigates ill-conditioning.

Framing the scaffold within this approximation structure provides a clear and mathematically coherent path for innovation. Agents may explore new classes of universal approximators such as KANs, KKANs, or radial basis models, or propose structural variants of existing networks by altering their internal basis functions, for example, by replacing spline activations with Chebyshev functions. They may also introduce input transformations that improve the basis quality of an MLP and reduce spectral bias, or incorporate architectural elements such as residual connections and normalization layers to stabilize the coefficient computation. Within this blueprint, such modifications arise naturally as consequences of the underlying approximation theory rather than as heuristic architectural experiments.

\subsubsection{Optimization Blueprint}

The optimization scaffold is motivated by the fact that, in PIML and operator learning, performance is largely determined by optimization. Conceptually, the optimization process contains two coupled design elements: the objective functional (loss) $\mathcal{L}(\theta)$ and the update rule used to minimize it. The blueprint therefore instructs agents to treat both the loss construction and the optimizer selection as explicit and modifiable components of the workflow, providing clear structural points of intervention when training stagnates.

On the loss side, the blueprint begins from the observation that many scientific learning tasks can be expressed as residual minimization over a continuous domain. The performance of such a model is quantified by a residual function $r(x;\theta)$ measuring local error at each point $x\in\Omega$. Training is then interpreted as minimizing an aggregate residual energy over $\Omega$, implemented through Monte Carlo discretization under a sampling distribution $q$,
\[
\mathcal{L}(\theta)=\mathbb{E}_{x\sim q}\big[r(x;\theta)^2\big].
\]
 
In this scaffold, $q$ is treated as a design variable that determines which regions of $\Omega$ dominate the discrete objective. When $q$ is uniform (or more generally $q=p$ for a prescribed base measure $p$), this recovers the standard mean-squared formulation and its discrete $L^2$ interpretation. However, uniform sampling is not always aligned with the scientific objective; stiff problems or problems with localized structure often require concentrating evaluations where the residual is large or controlling variance where the residual is noisy. The blueprint therefore treats the choice of $q$ and the induced functional being optimized as part of the loss design.

This perspective is formalized through the variational dual framework of \cite{toscano2025variational}. For a base measure $p$ on $\Omega$ and $\varepsilon>0$, one considers objectives induced by $\phi$-divergence regularization through the dual problem
\[
\sup_{q\in\mathcal{P}(\Omega)}
\left\{
\mathbb{E}_{x\sim q}\big[r(x;\theta)^2\big]
-\varepsilon\, D_\phi(q\Vert p)
\right\}.
\]
The optimizer $q$ admits a closed-form structure determined by the chosen potential,
\[
q_\phi(dx)\propto \phi'\!\left(\frac{r(x;\theta)^2}{\varepsilon}\right)p(dx),
\]
which shows that adaptive sampling and weighting arise from a single principle: the variational formulation selects a distribution that increases mass on regions of high residual while remaining controlled by the divergence penalty. Different choices of $\phi$ induce different effective objectives. When $\phi(z)=e^{z}$, the resulting exponential tilting concentrates mass near maximizers of $r(x;\theta)^2$ as $\varepsilon\to 0$, yielding an $L^\infty$-type behavior that targets uniform error control. When $\phi(z)=\tfrac12(z^2+1)$, the resulting chi-square tilting induces variance-reduction behavior and recovers a discretized form of RAD. Intermediate potentials interpolate between these behaviors and define a family of norm-like objectives. In this way, selecting $\phi$ provides a direct recipe for generating new norms and corresponding sampling or weighting rules, yielding mathematically coherent alternatives to fixed $L^2$ objectives.

On the optimizer side, once an objective $\mathcal{L}(\theta^k)$ is defined and evaluated, parameters are updated by an iterative scheme. A broad class of methods can be written as a line-search update
\begin{align}
    \label{eq:line-search}
    \theta^{k+1} &= \theta^{k}+\alpha^{k} p^{k},
\end{align}
where $\alpha^{k}$ is a step size and $p^{k}=-H_k\nabla_{\theta}\mathcal{L}(\theta^{k})$ is a descent direction determined by the gradient and a symmetric matrix $H_k$ \citep{urban2024unveiling}. When $H_k=I$, this reduces to standard gradient descent. When $H_k$ is the Hessian or an approximation of it, one obtains Newton or quasi-Newton methods. The blueprint therefore instructs agents to treat the choice of $H_k$, the step-size strategy, and the overall update rule as modifiable design components. In particular, for stiff or ill-conditioned regimes common in PIML, the scaffold permits the use of quasi-Newton updates such as SSBroyden, as well as first-order methods such as Adam, together with appropriate line-search or step-size control. This decomposition into loss design and optimizer design provides the agent with explicit points of action for diagnosing failures and proposing targeted modifications rather than relying on ad hoc adjustments.

\subsubsection{Physics-Informed Machine Learning Blueprint}

As discussed in \cite{toscano2025pinns}, we view a Physics-Informed Machine Learning (PIML) problem as a solver, namely a representation model that approximates the solution of a PDE through an optimization problem. Under this viewpoint, improving a PIML solver decomposes naturally into three coupled components: the representation model, the optimization procedure, and the PDE itself. The approximation viewpoint blueprint governs structural improvements of the representation, while the optimization blueprint governs the construction of effective objectives, sampling rules, and optimizers. The remaining component is therefore the PDE-level design space, and this blueprint instructs agents to treat the governing equations and constraints as explicit objects of modification rather than fixed inputs.

A critical first step in this blueprint is that agents are forced to assess ill-posedness before proposing modifications. This diagnostic check determines whether the available information is sufficient to identify the desired solution or parameters, and it guides whether the appropriate response is to reformulate the inverse problem, modify the constraint structure, introduce additional regularization, or change the learning objective. This step is essential for distinguishing optimization failures from fundamental identifiability limitations.

At the PDE level, the scaffold emphasizes targeted interventions that alter the analytical structure presented to the learner. This includes the way derivatives are computed and represented, the specific differential operator enforced, and the manner in which constraints are imposed. In particular, PDE or ODE reformulations are treated as a principled mechanism for simplifying the constraint structure and improving conditioning. A standard example is rewriting incompressible flow formulations in terms of a stream function or vector potential, which reduces the burden of explicitly enforcing conservation of mass by encoding it structurally through the reformulated variables. More generally, reformulations that reduce the number of variables or eliminate redundant constraints are permitted whenever they preserve the underlying physics while simplifying the optimization landscape.

The blueprint also instructs agents to encode constraints through architecture when appropriate, which naturally interacts with the approximation blueprint. Input embeddings can be used to encode periodicity, output transformations or ansatz constructions can enforce Dirichlet boundary conditions, and potential-based formulations can encode divergence-free structure through stream functions or vector potentials. In addition, the scaffold permits continuation or homotopy strategies in regimes where direct training is unstable, such as high Reynolds number flows, where gradually increasing the difficulty of the PDE or interpolating between related operators can improve convergence. Finally, the blueprint allows the agent to inherit stable design choices from classical numerical analysis, including stabilization and regularization mechanisms that are traditionally introduced at the PDE level to control oscillations, stiffness, or ill-conditioning. A descriptive decomposition of a broad set of PINN studies into these categories is provided in \cite{toscano2025pinns}.

\subsubsection{Numerical Methods Blueprint}

A primary principle in this blueprint is that the numerical treatment is governed first by the mathematical character of the PDE and the regularity of its solutions. A large class of models in scientific computing can be written in conservation form
\[
\partial_t u + \nabla\cdot f(u)=s(u),
\]
for which hyperbolic structure is associated with real characteristic speeds, while elliptic structure is associated with coercive steady operators. This distinction is not cosmetic: it determines whether information propagates along characteristics and whether discontinuities and steep gradients are structurally admissible. In hyperbolic regimes, shocks and contact discontinuities are generic, and the numerical objective is to approximate weak solutions while controlling nonphysical oscillations. In elliptic regimes, solutions are typically smoother and constrained globally by the operator, and the numerical objective is to construct stable approximations in a coercive energy norm.

A second organizing concept is the characteristic viewpoint. For transport-dominated equations, the method of characteristics formalizes the fact that the PDE can reduce to an ODE along curves $x(t)$ whose speed is determined by the flux, yielding an exact or semi-exact representation when such a reduction is available. This provides a principled analytic route in low-dimensional hyperbolic settings and clarifies which solution features are intrinsic to the PDE rather than artifacts of discretization.

A third principle is the role of basis representations in the presence of smoothness and symmetry. On periodic domains with sufficiently smooth solutions, Fourier expansions
\[
u(x,t)\approx \sum_{k=-K}^{K}\hat u_k(t)e^{ikx}
\]
provide a natural global representation and yield rapid convergence when the solution is regular. In contrast, when the solution contains localized nonsmooth structure, global bases become inefficient and can exhibit oscillatory artifacts, reflecting the mismatch between smooth basis functions and discontinuous targets.

A fourth principle is the variational viewpoint underlying Galerkin discretizations. For steady elliptic problems, the weak formulation
\[
a(u,v)=\ell(v)\qquad \forall v\in V
\]
suggests conforming approximations in subspaces $V_h\subset V$, which motivates continuous Galerkin methods when coercivity and regularity dominate. For hyperbolic conservation laws, the same Galerkin logic applied on broken spaces leads naturally to discontinuous Galerkin formulations, where inter-element coupling is defined through numerical fluxes and stabilization mechanisms that control oscillations while preserving conservation. High-order discontinuous Galerkin spectral element methods fit within this principle by combining element-local polynomial representations with interface fluxes, allowing the discretization to represent both smooth regions and sharp transitions within a single framework.

A further unifying concept is the method-of-lines viewpoint for time-dependent problems. After selecting a spatial representation, whether spectral or Galerkin on conforming or broken spaces, one obtains a finite-dimensional semi-discrete system of the form
\[
\frac{d}{dt}U(t)=F(U(t)),
\]
where $U(t)$ collects the coefficients of the chosen basis expansion and $F$ encodes the discrete differential operator (including interface coupling in discontinuous formulations). This separation clarifies that many differences between numerical solvers arise from the spatial discretization and its stability mechanisms, while the remaining component is the choice of a time-integration map compatible with the stiffness and invariants induced by $F$.

Finally, nonlinear steady regimes introduce an additional structural idea: continuation. When the operator depends on a parameter (for instance a Reynolds number), one may consider a family of problems $\mathcal{P}_\lambda$ and track solutions as $\lambda$ varies, using the solution at one parameter value as an initializer for the next. This perspective treats convergence difficulty as a property of the parameterized PDE family and provides a principled mechanism for stabilizing strongly nonlinear solves without altering the underlying model.

With these principles in place, numerical modeling reduces to matching the expected solution structure to an appropriate representation paradigm. The classification into hyperbolic versus elliptic behavior, the presence or absence of characteristic structure, and the role of smoothness and symmetry determine whether the natural formulation is characteristic-based, spectral, or Galerkin in conforming or broken spaces. This yields a compact but theory-driven decision space in which discretization choices are aligned with the PDE character rather than introduced by ad hoc trial-and-error.

\subsection{ATHENA (Agentic Team for Hierarchical Evolutionary Numerical Algorithms)}

Our complete agentic framework, ATHENA (Agentic Team for Hierarchical Evolutionary Numerical Algorithms), is illustrated in Figure~\ref{Framework}. The architecture is organized into four distinct logical groups (A-D). The process begins with the Conceptualization Group (Figure~\ref{Framework} A, Red), which acts as the user-facing triage system. The User interacts with a Coordinator to formalize a problem, and the Gatekeeper routes this User Request to the appropriate generative group or the Storage Group, effectively establishing the formal problem definition.

This initialization triggers the core HENA (Hierarchical Evolutionary Numerical Algorithms) loop—comprising Groups B, C, and D—which operationalizes the Contextual Bandit formulation described in the previous section. This loop is driven by the `policy' operator ($\pi$) teams (Figure~\ref{Framework} B, Green), which represent the ``brain" of the system. This group, containing the Strategy and Advisor Teams, receives the full history from previous cycles and formulates the new high-level structural action, $A_n$. To address the known limitation of LLMs regarding the lack of intrinsic verification and awareness of consequences, this group employs a proposer-critic architecture where a Critic Agent validates the Strategist's plan. This plan $A_n$ is then passed to the `implementation' operator ($\mathcal{I}$) teams (Figure~\ref{Framework} C, Blue).

The execution phase begins with the Code Retrieval Team. To prevent the downstream Planner agent from succumbing to the ``Lost in the Middle" phenomenon caused by excessive context, this team functions as an intelligent filter. It employs an Agentic RAG system to retrieve the most suitable base templates and, crucially, performs Agentic RAG Generation to synthesize missing modules (e.g., custom layers) that are designed to be compatible with the retrieved context.
 This ensures that the Planner receives a focused cohesive codebase, significantly simplifying its reasoning task.

Once the context is prepared, the Implementation Team handles the assembly. This group is designed to mitigate the \textit{autoregressive error accumulation}—formally known as {conceptual drift}—inherent in long-context generation. Since the probability of a correct sequence decays exponentially with length ($P(\text{correct}) \approx p^N \to 0$), monolithic generation is mathematically prone to divergence. To mitigate this, the group employs a strict `cell-by-cell refactoring' protocol driven by a Planner Agent. This agent operates in two focusing stages: first, it studies the codebase to identify only the specific cells requiring modification; second, it targets the refactoring exclusively to those cells. This containment strategy enables the framework to robustly generate and maintain extensive scientific codebases scaling to thousands of lines of code. To further address the lack of consequence awareness at the code level, the workflow includes an Inspector Agent. This agent acts as a verifier, strictly auditing the Planner's modifications against the desired plan $A_n$ before the code produces the final executable state, $S_n$. If runtime errors occur, the Debugging Team is activated to iteratively patch the code until execution is successful.

Finally, the state $S_n$ is processed by the Execution block (Figure~\ref{Framework} D, Yellow), which runs the observation function $E(O_n | S_n)$ on the state $S_n$ to generate the multi-modal observation, $O_n$ (e.g., plots and logs). This observation $O_n$ is sent back to the Advisor Team (Green), which analyzes it to generate the scientific reward $R_n$ and the diagnostic report, completing the cycle. Importantly, the key components of each cycle ($A_n$, $S_n$, $O_n$) are available to both the agent teams and the user, creating a transparent, fully closed-loop system. Furthermore, the final validated codes from successful experiments are fed back to the Storage Group, allowing the framework's knowledge base to grow over time. A detailed description of each agent's role is provided in the SI.

\subsection{Error Computation}
For all PIML benchmarks, we report the relative $L^2$ error between the predicted field and the corresponding reference solution evaluated on a dense grid spanning the full solution domain:
\begin{equation}
RL_2
=\frac{\|\hat{u}(x)-u(x)\|_2}{\|\hat{u}(x)\|_2}
=\frac{\sqrt{ \sum_{i=1}^n\left(\hat{u}(\mathbf{x}_i) - u(\mathbf{x}_i, \theta)\right)^2}}{\sqrt{\sum_{i=1}^n\hat{u}(\mathbf{x}_i)^2}}.
\label{eq:RL2}
\end{equation}
Here, $\hat{u}(\mathbf{x}_i)$ denotes the reference solution value at point $\mathbf{x}_i$. For Allen--Cahn and viscous Burgers, we use the publicly available reference data released with the original studies~\citep{mcclenny2023self}. For Helmholtz, Poisson, KdV, and inviscid Burgers, we use analytical solutions to compute $\hat{u}$ on the evaluation grid. The model prediction is denoted by $u(\mathbf{x}_i,\theta)$. Unless otherwise stated, the evaluation set is a fixed dense grid of $n\approx 2\times 10^{5}$ points covering the full spatial domain (and, for time-dependent problems, the full space--time domain), and is kept identical across methods to enable direct comparison. This definition is standard in the PINN/PIML literature for reporting solution-field accuracy~\citep{urban2024unveiling,kiyani2025optimizing,wang2025gradient}. \rev{For each benchmark, this error is taken at the best checkpoint of the Advisor-approved run, namely the training iteration of lowest relative $L^2$ error.}


\section*{Declarations}

\paragraph*{Funding}
This work was supported by the NIH grant R01AT012312, the MURI/AFOSR FA9550-20-1-0358 project, the DOE-MMICS SEA-CROGS DE-SC0023191 award, and the ONR Vannevar Bush Faculty Fellowship N00014-22-1-2795 (J.D.T. and G.E.K.), and by the DARPA grant HR0011-25-3-0223 (D.T.C.).

\paragraph*{Competing interests}
The authors declare no competing interests.

\paragraph*{Author contribution}
\begin{enumerate}
    \item \textbf{Conceptualization:} G.E.K, J.D.T,  D.T.C.
    \item \textbf{Methodology:} J.D.T., D.T.C., G.E.K.
    \item \textbf{Software:} J.D.T.
    \item \textbf{Formal analysis:} J.D.T., D.T.C.
    \item \textbf{Investigation:} J.D.T., D.T.C..
    \item \textbf{Resources:} G.E.K.
    \item \textbf{Writing – original draft:} J.D.T., D.T.C., G.E.K.
    \item \textbf{Writing – review \& editing:} J.D.T., D.T.C., G.E.K.
    \item \textbf{Visualization:} J.D.T.
    \item \textbf{Supervision:} G.E.K.
    \item \textbf{Project administration:} G.E.K.
    \item \textbf{Funding acquisition:} G.E.K.
\end{enumerate}
\paragraph*{Data and code availability:}

To support reproducibility, all data and code for this study will be made publicly available upon publication. 



\bibliographystyle{elsarticle-num}
\bibliography{cas-refs}

@article{ghafarollahi2025sciagents,
  title={{SciAgents}: automating scientific discovery through bioinspired multi-agent intelligent graph reasoning},
  author={Ghafarollahi, Alireza and Buehler, Markus J},
  journal={Advanced Materials},
  volume={37},
  number={22},
  pages={2413523},
  year={2025},
  publisher={Wiley Online Library}
}

@article{takamoto2022pdebench,
  title={{PDEBench}: An extensive benchmark for scientific machine learning},
  author={Takamoto, Makoto and Praditia, Timothy and Leiteritz, Raphael and MacKinlay, Daniel and Alesiani, Francesco and Pfl{\"u}ger, Dirk and Niepert, Mathias},
  journal={Advances in Neural Information Processing Systems},
  volume={35},
  pages={1596--1611},
  year={2022}
}

@inproceedings{kim2021dpm,
  title={{DPM}: A novel training method for physics-informed neural networks in extrapolation},
  author={Kim, Jungeun and Lee, Kookjin and Lee, Dongeun and Jhin, Sheo Yon and Park, Noseong},
  booktitle={Proceedings of the AAAI conference on artificial intelligence},
  volume={35},
  number={9},
  pages={8146--8154},
  year={2021}
}

@article{toscano2025variational,
  title={A variational framework for residual-based adaptivity in neural PDE solvers and operator learning},
  author={Toscano, Juan Diego and Chen, Daniel T and Ooomen, Vivek and Darbon, J{\'e}r{\^o}me and Karniadakis, George Em},
  journal={npj Artificial Intelligence},
  volume={2},
  number={1},
  pages={32},
  year={2026},
  publisher={Nature Publishing Group UK London}
}

@article{georgiev2025mathematical,
  title={Mathematical exploration and discovery at scale},
  author={Georgiev, Bogdan and G{\'o}mez-Serrano, Javier and Tao, Terence and Wagner, Adam Zsolt},
  journal={arXiv preprint arXiv:2511.02864},
  year={2025}
}

@article{he2025lang,
  title={{Lang-PINN}: From Language to Physics-Informed Neural Networks via a Multi-Agent Framework},
  author={He, Xin and You, Liangliang and Tian, Hongduan and Han, Bo and Tsang, Ivor and Ong, Yew-Soon},
  journal={arXiv preprint arXiv:2510.05158},
  year={2025}
}

@article{jiang2025agenticsciml,
  title={Agenticsciml: Collaborative multi-agent systems for emergent discovery in scientific machine learning},
  author={Jiang, Qile and Karniadakis, George},
  journal={npj Artificial Intelligence},
  year={2026},
  publisher={Nature Publishing Group UK London}
}

@article{aygun2025ai,
  title={An {AI} system to help scientists write expert-level empirical software},
  author={Ayg{\"u}n, Eser and Belyaeva, Anastasiya and Comanici, Gheorghe and Coram, Marc and Cui, Hao and Garrison, Jake and Kast, Renee Johnston Anton and McLean, Cory Y and Norgaard, Peter and Shamsi, Zahra and others},
  journal={arXiv preprint arXiv:2509.06503},
  year={2025}
}

@article{bhatnagar2025agentic,
  title={An Agentic {AI} Workflow to Simplify Parameter Estimation of Complex Differential Equation Systems},
  author={Bhatnagar, Saakaar},
  journal={arXiv preprint arXiv:2509.07283},
  year={2025}
}

@article{lupoiu2025multi,
  title={A multi-agentic framework for real-time, autonomous freeform metasurface design},
  author={Lupoiu, Robert and Shao, Yixuan and Dai, Tianxiang and Mao, Chenkai and Ed{\'e}e, Kofi and Fan, Jonathan A},
  journal={Science Advances},
  volume={11},
  number={44},
  pages={eadx8006},
  year={2025},
  publisher={American Association for the Advancement of Science}
}

@article{wuwu2025pinnsagent,
  title={{PINNsAgent}: Automated {PDE} Surrogation with Large Language Models},
  author={Wuwu, Qingpo and Gao, Chonghan and Chen, Tianyu and Huang, Yihang and Zhang, Yuekai and Wang, Jianing and Li, Jianxin and Zhou, Haoyi and Zhang, Shanghang},
  journal={arXiv preprint arXiv:2501.12053},
  year={2025}
}

@article{li2025codepde,
  title={{CodePDE}: An inference framework for {LLM}-driven {PDE} solver generation},
  author={Li, Shanda and Marwah, Tanya and Shen, Junhong and Sun, Weiwei and Risteski, Andrej and Yang, Yiming and Talwalkar, Ameet},
  journal={arXiv preprint arXiv:2505.08783},
  year={2025}
}

@article{liu2025pde,
  title={{PDE-Agent}: A toolchain-augmented multi-agent framework for {PDE} solving},
  author={Liu, Jianming and Zhu, Ren and Xu, Jian and Ding, Kun and Zhang, Xu-Yao and Meng, Gaofeng and Liu, Cheng-Lin},
  journal={arXiv preprint arXiv:2512.16214},
  year={2025}
}

@article{real2023autonumerics,
  title={{AutoNumerics-Zero}: Automated discovery of state-of-the-art mathematical functions},
  author={Real, Esteban and Chen, Yao and Rossini, Mirko and de Souza, Connal and Garg, Manav and Verghese, Akhil and Firsching, Moritz and Le, Quoc V and Cubuk, Ekin Dogus and Park, David H},
  journal={arXiv preprint arXiv:2312.08472},
  year={2023}
}

@article{jiang2025aide,
  title={{AIDE}: {AI}-driven exploration in the space of code},
  author={Jiang, Zhengyao and Schmidt, Dominik and Srikanth, Dhruv and Xu, Dixing and Kaplan, Ian and Jacenko, Deniss and Wu, Yuxiang},
  journal={arXiv preprint arXiv:2502.13138},
  year={2025}
}

@article{huang2023mlagentbench,
  title={{MLAgentBench}: Evaluating language agents on machine learning experimentation},
  author={Huang, Qian and Vora, Jian and Liang, Percy and Leskovec, Jure},
  journal={arXiv preprint arXiv:2310.03302},
  year={2023}
}

@inproceedings{chan2025mle,
  title={{MLE-bench}: Evaluating machine learning agents on machine learning engineering},
  author={Chan, Jun Shern and Chowdhury, Neil and Jaffe, Oliver and Aung, James and Sherburn, Dane and Mays, Evan and Starace, Giulio and Liu, Kevin and Maksin, Leon and Patwardhan, Tejal and others},
  booktitle={International Conference on Learning Representations},
  volume={2025},
  pages={50466--50494},
  year={2025}
}

@article{romera2024mathematical,
  title={Mathematical discoveries from program search with large language models},
  author={Romera-Paredes, Bernardino and Barekatain, Mohammadamin and Novikov, Alexander and Balog, Matej and Kumar, M Pawan and Dupont, Emilien and Ruiz, Francisco JR and Ellenberg, Jordan S and Wang, Pengming and Fawzi, Omar and others},
  journal={Nature},
  volume={625},
  number={7995},
  pages={468--475},
  year={2024},
  publisher={Nature Publishing Group UK London}
}

@article{samadi2025flexible,
  title={Flexible Swarm Learning May Outpace Foundation Models in Essential Tasks},
  author={Samadi, Moein E and Schuppert, Andreas},
  journal={arXiv preprint arXiv:2510.06349},
  year={2025}
}

@article{kiyani2025optimizing,
  title={Optimizing the optimizer for physics-informed neural networks and {Kolmogorov-Arnold} networks},
  author={Kiyani, Elham and Shukla, Khemraj and Urb{\'a}n, Jorge F and Darbon, J{\'e}r{\^o}me and Karniadakis, George Em},
  journal={Computer Methods in Applied Mechanics and Engineering},
  volume={446},
  pages={118308},
  year={2025},
  publisher={Elsevier}
}

@inproceedings{
wang2025gradient,
title={Gradient Alignment in Physics-informed Neural Networks: A Second-Order Optimization Perspective},
author={Sifan Wang and Ananyae Kumar bhartari and Bowen Li and Paris Perdikaris},
booktitle={The Thirty-ninth Annual Conference on Neural Information Processing Systems},
year={2025},
url={https://openreview.net/forum?id=iweeVl1RHU}
}

@article{wu2025propinn,
  title={{ProPINN}: Demystifying propagation failures in physics-informed neural networks},
  author={Wu, Haixu and Ma, Yuezhou and Zhou, Hang and Weng, Huikun and Wang, Jianmin and Long, Mingsheng},
  journal={arXiv preprint arXiv:2502.00803},
  year={2025}
}

@inproceedings{zhao2024pinnsformer,
  title={{PINNsFormer}: A Transformer-Based Framework For Physics-Informed Neural Networks},
  author={Zhao, Zhiyuan and Ding, Xueying and Prakash, B. Aditya},
  booktitle={The Twelfth International Conference on Learning Representations},
  year={2024},
  url={https://openreview.net/forum?id=a6f2763089c0bd8f56006c42f09ee24c}
}

@article{dong2021method,
  title={A method for representing periodic functions and enforcing exactly periodic boundary conditions with deep neural networks},
  author={Dong, Suchuan and Ni, Naxian},
  journal={Journal of Computational Physics},
  volume={435},
  pages={110242},
  year={2021},
  publisher={Elsevier}
}

@article{zhongkai2024pinnacle,
  title={{PINNacle}: A comprehensive benchmark of physics-informed neural networks for solving {PDEs}},
  author={Zhongkai, Hao and Yao, Jiachen and Su, Chang and Su, Hang and Wang, Ziao and Lu, Fanzhi and Xia, Zeyu and Zhang, Yichi and Liu, Songming and Lu, Lu and others},
  journal={Advances in Neural Information Processing Systems},
  volume={37},
  pages={76721--76774},
  year={2024}
}

@inproceedings{xu2025fp64,
  title={{FP64} is All You Need: Rethinking Failure Modes in Physics-Informed Neural Networks},
  author={Xu, Chenhui and Liu, Dancheng and Nassereldine, Amir and Xiong, Jinjun},
  booktitle={Advances in Neural Information Processing Systems},
  year={2025}
}

@article{toscano2025kkans,
  title={{KKANs: Kurkova-Kolmogorov-Arnold networks and their learning dynamics}},
  author={Toscano, Juan Diego and Wang, Li-Lian and Karniadakis, George Em},
  journal={Neural Networks},
  pages={107831},
  year={2025},
  publisher={Elsevier}
}

@article{chen2025self,
  title={Self-adaptive weights based on balanced residual decay rate for physics-informed neural networks and deep operator networks},
  author={Chen, Wenqian and Howard, Amanda A and Stinis, Panos},
  journal={Journal of Computational Physics},
  pages={114226},
  year={2025},
  publisher={Elsevier}
}

@article{raissi2019physics,
  title = {{Physics-informed neural networks: A deep learning framework for solving forward and inverse problems involving nonlinear partial differential equations}},
  author={Raissi, Maziar and Perdikaris, Paris and Karniadakis, George E},
  journal={Journal of Computational Physics},
  volume={378},
  pages={686--707},
  year={2019},
  publisher={Elsevier}
}

@article{zhang2023dasa,
  title = {{{DASA-PINNs: Differentiable Adversarial Self-Adaptive Pointwise Weighting Scheme for Physics-Informed Neural Networks}}},
  journal={SSRN},
  author={Zhang, Guangtao and Yang, Huiyu and Zhu, Fang and Chen, Yang and others},
  year={2023}
}

@article{jagtap2020adaptive,
  title = {{{Adaptive activation functions accelerate convergence in deep and physics-informed neural networks}}},
  author={Jagtap, Ameya D and Kawaguchi, Kenji and Karniadakis, George Em},
  journal={Journal of Computational Physics},
  volume={404},
  pages={109136},
  year={2020},
  publisher={Elsevier}
}

@article{ghafarollahi2025rapid,
  title={Rapid and automated alloy design with graph neural network-powered large language model-driven multi-agent {AI}},
  author={Ghafarollahi, Alireza and Buehler, Markus J},
  journal={MRS Bulletin},
  volume={50},
  number={11},
  pages={1309--1324},
  year={2025},
  publisher={Springer}
}

@article{wang2025swarms,
  title={Swarms of Large Language Model Agents for Protein Sequence Design with Experimental Validation},
  author={Wang, Fiona Y and Lee, Di Sheng and Kaplan, David L and Buehler, Markus J},
  journal={arXiv preprint arXiv:2511.22311},
  year={2025}
}

@article{wang2024piratenets,
  title = {{PirateNets: Physics-informed Deep Learning with Residual Adaptive Networks}},
  author={Wang, Sifan and Li, Bowen and Chen, Yuhan and Perdikaris, Paris},
  journal={arXiv preprint arXiv:2402.00326},
  year={2024}
}

@article{Wang2020_Fourier_nets,
  author       = {Sifan Wang and
                  Hanwen Wang and
                  Paris Perdikaris},
  title = {{On the eigenvector bias of Fourier feature networks: From regression
                  to solving multi-scale {PDEs} with physics-informed neural networks}},
  journal      = {arXiv preprint arXiv:2012.10047},
  year         = {2020},
}

@article{mcclenny2023self,
  title = {{Self-adaptive physics-informed neural networks}},
  author={McClenny, Levi D and Braga-Neto, Ulisses M},
  journal={Journal of Computational Physics},
  volume={474},
  pages={111722},
  year={2023},
  publisher={Elsevier}
}

@inproceedings{ross2011reduction,
  title={A reduction of imitation learning and structured prediction to no-regret online learning},
  author={Ross, St{\'e}phane and Gordon, Geoffrey and Bagnell, Drew},
  booktitle={Proceedings of the fourteenth international conference on artificial intelligence and statistics},
  pages={627--635},
  year={2011},
  organization={JMLR Workshop and Conference Proceedings}
}

@article{wiggins2022opportunities,
  title={On the opportunities and risks of foundation models for natural language processing in radiology},
  author={Wiggins, Walter F and Tejani, Ali S},
  journal={Radiology: Artificial Intelligence},
  volume={4},
  number={4},
  pages={e220119},
  year={2022},
  publisher={Radiological Society of North America}
}

@article{lu2024ai,
  title={The {AI} scientist: Towards fully automated open-ended scientific discovery},
  author={Lu, Chris and Lu, Cong and Lange, Robert Tjarko and Foerster, Jakob and Clune, Jeff and Ha, David},
  journal={arXiv preprint arXiv:2408.06292},
  year={2024}
}

@article{choi2025defining,
  title={Defining Foundation Models for Computational Science: A Call for Clarity and Rigor},
  author={Choi, Youngsoo and Cheung, Siu Wun and Kim, Youngkyu and Tsai, Ping-Hsuan and Diaz, Alejandro N and Zanardi, Ivan and Chung, Seung Whan and Copeland, Dylan Matthew and Kendrick, Coleman and Anderson, William and others},
  journal={arXiv preprint arXiv:2505.22904},
  year={2025}
}

@article{subramanian2023towards,
  title={Towards foundation models for scientific machine learning: Characterizing scaling and transfer behavior},
  author={Subramanian, Shashank and Harrington, Peter and Keutzer, Kurt and Bhimji, Wahid and Morozov, Dmitriy and Mahoney, Michael W and Gholami, Amir},
  journal={Advances in Neural Information Processing Systems},
  volume={36},
  pages={71242--71262},
  year={2023}
}

@article{openai2025gpt51,
  title={{GPT-5.1}: A Smarter, More Conversational {ChatGPT}},
  author={OpenAI},
  journal={OpenAI Blog},
  year={2025},
  month={November},
  url={https://openai.com/index/gpt-5-1/},
  note={Released November 12, 2025}
}

@techreport{google2025gemini3,
  title={Gemini 3 Technical Report},
  author={Google DeepMind},
  institution={Google},
  year={2025},
  month={November},
  note={Released November 18, 2025},
  url={https://blog.google/technology/ai/}
}

@techreport{anthropic2025claude45,
  title={Claude {Sonnet} 4.5 System Card},
  author={Anthropic},
  year={2025},
  month={November},
  institution={Anthropic},
  url={https://www.anthropic.com/claude-sonnet-4-5-system-card},
  note={Released November 24, 2025}
}

@techreport{xai2025grok41,
  title={Grok 4.1 Model Card},
  author={xAI},
  year={2025},
  month={November},
  institution={xAI},
  url={https://data.x.ai/2025-11-17-grok-4-1-model-card.pdf},
  note={Released November 17, 2025}
}

@article{herde2024poseidon,
  title={Poseidon: Efficient foundation models for {PDEs}},
  author={Herde, Maximilian and Raonic, Bogdan and Rohner, Tobias and K{\"a}ppeli, Roger and Molinaro, Roberto and de B{\'e}zenac, Emmanuel and Mishra, Siddhartha},
  journal={Advances in Neural Information Processing Systems},
  volume={37},
  pages={72525--72624},
  year={2024}
}

@article{schneider2024foundation,
  title={Foundation models: A new paradigm for artificial intelligence},
  author={Schneider, Johannes and Meske, Christian and Kuss, Pauline},
  journal={Business \& Information Systems Engineering},
  volume={66},
  number={2},
  pages={221--231},
  year={2024},
  publisher={Springer}
}

@article{gama2014survey,
  title={A survey on concept drift adaptation},
  author={Gama, Jo{\~a}o and {\v{Z}}liobait{\.e}, Indr{\.e} and Bifet, Albert and Pechenizkiy, Mykola and Bouchachia, Abdelhamid},
  journal={ACM computing surveys (CSUR)},
  volume={46},
  number={4},
  pages={1--37},
  year={2014},
  publisher={ACM New York, NY, USA}
}

@article{liu2024lost,
  title={Lost in the middle: How language models use long contexts},
  author={Liu, Nelson F and Lin, Kevin and Hewitt, John and Paranjape, Ashwin and Bevilacqua, Michele and Petroni, Fabio and Liang, Percy},
  journal={Transactions of the Association for Computational Linguistics},
  volume={12},
  pages={157--173},
  year={2024}
}

@article{cybenko1989approximation,
  title = {{Approximation by superpositions of a sigmoidal function}},
  author={Cybenko, George},
  journal={Mathematics of Control, Signals and Systems},
  volume={2},
  number={4},
  pages={303--314},
  year={1989},
  publisher={Springer}
}

@article{hornik1989multilayer,
  title = {{Multilayer feedforward networks are universal approximators}},
  author={Hornik, Kurt and Stinchcombe, Maxwell and White, Halbert},
  journal={Neural Networks},
  volume={2},
  number={5},
  pages={359--366},
  year={1989},
  publisher={Elsevier}
}

@book{karniadakis2005spectral,
  title = {{Spectral/hp element methods for computational fluid dynamics}},
  author={Karniadakis, George and Sherwin, Spencer J},
  year={2005},
  publisher={Oxford University Press, USA}
}

@article{lu2021deepxde,
  title = {{DeepXDE: A deep learning library for solving differential equations}},
  author={Lu, Lu and Meng, Xuhui and Mao, Zhiping and Karniadakis, George Em},
  journal={SIAM Review},
  volume={63},
  number={1},
  pages={208--228},
  year={2021},
  publisher={SIAM}
}

@article{chen2024self,
  title={Self-adaptive weights based on balanced residual decay rate for physics-informed neural networks and deep operator networks},
  author={Chen, Wenqian and Howard, Amanda A and Stinis, Panos},
  journal={Journal of Computational Physics},
  pages={114226},
  year={2025},
  publisher={Elsevier}
}

@article{urban2024unveiling,
  title = {{Unveiling the optimization process of Physics Informed Neural Networks: How accurate and competitive can {PINNs}} be?},
  author={Urb{\'a}n, Jorge F and Stefanou, Petros and Pons, Jos{\'e} A},
  journal={arXiv preprint arXiv:2405.04230},
  year={2024}
}

@article{song2025explicit,
  title={Explicit Construction of Approximate {Kolmogorov-Arnold} Superpositions with {C2}-Smoothness},
  author={Song, Lunji and Toscano, Juan Diego and Wang, Li-Lian},
  journal={arXiv preprint arXiv:2508.04392},
  year={2025}
}

@article{gao2023failure,
  title={Failure-informed adaptive sampling for {PINNs}},
  author={Gao, Zhiwei and Yan, Liang and Zhou, Tao},
  journal={SIAM Journal on Scientific Computing},
  volume={45},
  number={4},
  pages={A1971--A1994},
  year={2023},
  publisher={SIAM}
}

@article{jiang2025deepseek,
  title={{DeepSeek} vs. {ChatGPT} vs. {Claude}: A comparative study for scientific computing and scientific machine learning tasks},
  author={Jiang, Qile and Gao, Zhiwei and Karniadakis, George Em},
  journal={Theoretical and Applied Mechanics Letters},
  volume={15},
  number={3},
  pages={100583},
  year={2025},
  publisher={Elsevier}
}

@article{wu2023gpt,
  title={{GPT} vs human for scientific reviews: A dual source review on applications of {ChatGPT} in science},
  author={Wu, Chenxi and Varghese, Alan John and Oommen, Vivek and Karniadakis, George Em},
  journal={arXiv preprint arXiv:2312.03769},
  year={2023}
}

@article{raissi2017physicsI,
  title = {{Physics Informed Deep Learning (Part I): Data-driven Solutions of Nonlinear Partial Differential Equations}},
  author={Raissi, Maziar and Perdikaris, Paris and Karniadakis, George Em},
  journal={arXiv preprint arXiv:1711.10561},
  year={2017}
}

@article{toscano2025pinns,
  title={From {PINNs} to {PIKANs}: Recent advances in physics-informed machine learning},
  author={Toscano, Juan Diego and Oommen, Vivek and Varghese, Alan John and Zou, Zongren and Ahmadi Daryakenari, Nazanin and Wu, Chenxi and Karniadakis, George Em},
  journal={Machine Learning for Computational Science and Engineering},
  volume={1},
  number={1},
  pages={1--43},
  year={2025},
  publisher={Springer}
}

@article{Kolmogorov1957,
  author       = {Andrey Kolmogorov},
  title        = {On the representation of continuous functions of several variables as superpositions of continuous functions of one variable and addition},
  year         = {1957},
  note         = {English translation: Amer. Math. Soc. Transl., 28: Sixteen Papers on Analysis (1963)}
}

@inproceedings{muller2023achieving,
  title={Achieving high accuracy with {PINNs} via energy natural gradient descent},
  author={M{\"u}ller, Johannes and Zeinhofer, Marius},
  booktitle={International Conference on Machine Learning},
  pages={25471--25485},
  year={2023},
  organization={PMLR}
}

@book{lattimore2020bandit,
  title={Bandit algorithms},
  author={Lattimore, Tor and Szepesv{\'a}ri, Csaba},
  year={2020},
  publisher={Cambridge University Press}
}

\newpage
\begin{center}
{\Large\bfseries Supplementary Information}
\end{center}
\bigskip
\setcounter{figure}{0}
\setcounter{table}{0}
\setcounter{equation}{0}
\setcounter{page}{1} %

\appendix

\tableofcontents
\section{Agentic Teams}
\label{agentic_teams}
\setcounter{figure}{0}

\begin{figure}[H]
    \centering
    \includegraphics[width=0.6\linewidth]{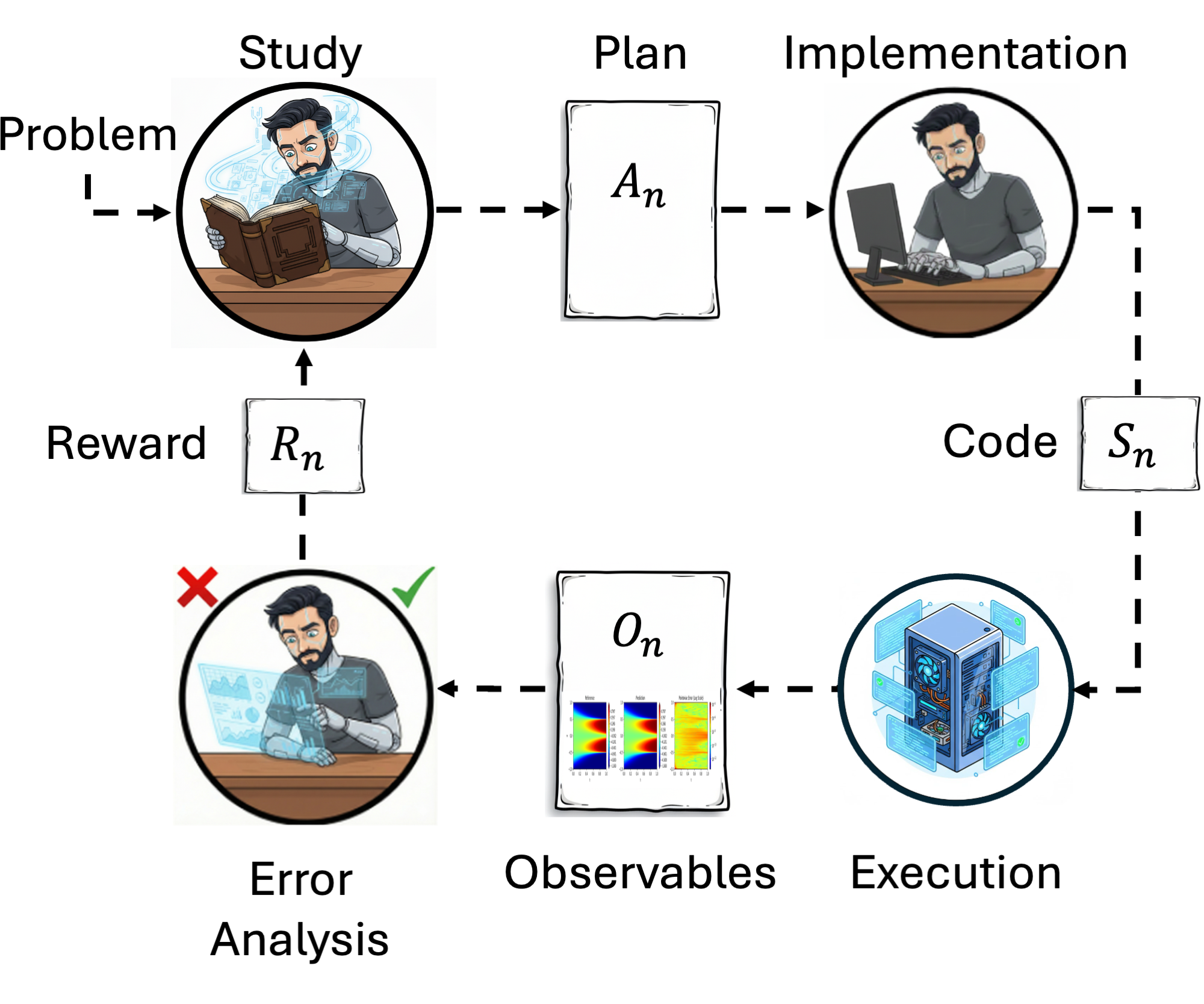}
    \caption{\textbf{Online Learning as a Model for Agentic Research.}
    The research lifecycle interpreted as a Contextual Bandit process.
    Each iteration consists of selecting an action $A_n$, implementing it to produce a code state $S_n$, generating observations $O_n$, and assigning a reward $R_n$.
    This loop enables iterative refinement based on structured feedback.}
    \label{fig:research_cycle}
\end{figure}

The conceptual scaffolding is instantiated within HENA (Hierarchical Evolutionary Numerical Algorithms), the practical agentic framework designed to manage the end-to-end research lifecycle. To capture the hierarchical nature of this process, we formally distinguish between two levels of reasoning that map directly to the Contextual Bandit formulation defined in the Methods section.

\begin{enumerate}
    \item The `first-order operator' ($\mathcal{I}$), which acts as the "executor." Its role is to translate a specific scientific strategy into a concrete solution. In our formulation, this corresponds to the Implementation Operator $\mathcal{I}: A_n \to S_n$, performed by the Implementation Teams (Code Retrieval, Implementation, Debugging).
    \item The `second-order' or `meta-reasoning' operator ($\pi$), which acts as the "adapter." Its role is not to solve the equation directly, but to evaluate the performance of the first-order operator and prescribe structural improvements. In our formulation, this corresponds to the Policy $\pi: \text{History} \to A_{n+1}$, performed by the Conceptualization Teams (Strategy and Advisor).
\end{enumerate}

The entire framework is organized around a key interaction flow managed by four primary logical groups, as illustrated in Figure~\ref{Framework}.

\subsection{The Conceptualization Group}
The HENA process is initiated by the user through the Conceptualization Group. This group includes a Coordinator, who functions as a high-level developer and domain expert. The Coordinator interacts with the user to help define the problem, performing an initial analysis to determine if the problem is well-posed, such as checking for sufficient boundary conditions and classifying the problem as forward or inverse. For advanced users, the Coordinator can also engage in collaborative development, translating high-level methodological suggestions into new, actionable strategies for the lab. The final output of this interaction is a formalized, detailed User Request. This formalized request is then passed to the Gatekeeper, which analyzes the problem formulation and decides the optimal execution path, such as storing data via the storage group or generating new code by routing the task to the SciC, SciML, or both groups. The Gatekeeper is also responsible for managing dependencies, such as first dispatching a task to the SciC group to generate validation data before routing the main PINN task to the SciML group. The Gatekeeper's final output is a "modified user request" that serves as the kick-off for the Generative Groups.

In practice, this hand-off defines a fully specified execution pipeline rather than a single task. Once the user approves the formalized request, the Gatekeeper decomposes it into an ordered sequence of stages and assigns each stage to the appropriate team, for example a Scientific Computing stage that generates a high-fidelity reference solution followed by a SciML stage that trains and validates a model against it. The user supplies the relevant data locations as part of the request, such as the path to an external dataset or to a reference solution produced by an earlier stage; given these inputs, the framework runs each stage as an independent HENA cycle, with downstream teams consuming the declared upstream artifacts. The user therefore defines the problem and its data sources, while the staging, routing, implementation, and per-stage execution of the solvers are carried out by the agents.

\subsection{The Storage Group}
The research framework's knowledge is managed by a specialized Storage Team, which functions as the central curator for all code templates and modules. Its primary role is to pre-process and structure this information to enable a powerful, guided Agentic RAG system. This team consists of two agents: the Analyzer and the Splitter. The Analyzer agent acts as a computational researcher, analyzing all code snippets. For modular, "cell-based" codes, it labels them and provides a clear description so they can be easily found later. For full PDE solvers, it extracts key metadata, such as the equation and method, into a structured JSON. The Splitter agent then analyzes long code scripts and identifies the most logical line numbers to break them apart for semantic chunking, such as at the end of functions or logical blocks. This knowledge base is dynamic; the user can deposit information directly, and final, validated codes from successful experiments are also fed back to the Storage Team. These agents then analyze and integrate the new solutions, allowing the framework's knowledge base to grow sequentially over time.

\subsection{The Generative Groups}

The generative teams are specialized by domain, including the Scientific Computing (SciC) Group, the SciML PIML Group, and the SciML Operator Learning Group. A key design principle is that all Generative Groups share an essentially identical agentic structure, which allows the framework to apply a consistent, robust process across all problem domains. This shared structure is a complete end-to-end scientific workflow.

\subsubsection{Strategic Team} The evolutionary cycle begins with the Strategy Team, which functions as the "scientific brain" of the framework. In its autonomous mode, it takes three primary inputs: the modified user request (defining the core problem), the trial history log (detailing all previous experimental runs), and the advisor report. The framework is also designed for expert-in-the-loop collaboration. An advanced user can interact directly with the Strategy Team. In this mode, the user can enhance, guide, or even directly alter the scientific plan, effectively overriding or supplementing the advisor's report with their own expertise.

In either operational mode, the advisor report (or the user's enhanced plan) is the critical driver of the evolutionary loop. It provides the specific scientific diagnosis and the explicit, corrective information (e.g., "replace the MLP trunk with a KAN trunk"). The Strategist's primary role is to translate this high-level scientific directive into a concrete, executable plan, while also considering the problem statement and experimental history.

This team utilizes a two-agent "proposer-critic" configuration. The first Strategist agent (the proposer) synthesizes all inputs—especially the Advisor's specific cure or the user's direct interventions—to generate the initial scientific plan. A second Strategist agent (the Critic) then analyzes this plan for logical consistency and potential flaws. For example, it would flag a plan that suggests normalizing inputs from -1 to 1 while simultaneously using a ReLU activation function. This collaborative process ensures the final plan is robust.

This final hypothesis is produced as the scientific strategy report, the master blueprint for the experiment. This report is the core mechanism for evolution, decoupling the Advisor's high-level scientific curve from the low-level implementation. The scientific strategy report is passed to the Code Retrieval Team. This team is not activated for every cycle; it is specifically engaged when the strategy report calls for new architectures or enhancement modules that are not already part of the active experimental script.

\subsubsection{Code Retrieval Team} The team's first function, handling the main architecture, involves two key agents: the Retriever and the Validator. The Retriever functions as a specialized RAG system. Its crucial role is to reconstruct long, foundational code templates from a database. This methodology is essential for dealing with a large code database and reduces the risk of hallucination in retrieved reference code. This base code is then passed to the Validator, which performs a critical methodological check. For instance, the Validator ensures the code is appropriate for the problem's mathematical character, flagging a mismatch such as selecting a global spectral method for a hyperbolic problem.

The team's second function is to gather the specific enhancement modules. The Collector finds the specific modules and their corresponding submodules (dependencies) from the code library. These are then passed to the Librarian. The Librarian analyzes the full module dataset and, if a required component is missing, is empowered to generate new, complex architectural code from scratch. For example, it could construct a novel Wavelet-KAN architecture. A key aspect of this generation is that the Librarian creates these new components in a modular way, ensuring they are compatible with the main code template.

The Code Retrieval Team's final outputs—the validated main code template, the required enhancement code (both existing and newly generated), and the scientific strategy report—are then passed to the Implementation Team.
\subsubsection{Implementation Team}
This team is a key component of the framework's first-order operator, receives the scientific strategy report, the base code template, and the enhancement code modules. If the cycle is a re-run due to a crash, it also receives a debug report. The process begins with a "study phase". A Planner agent first analyzes the base code and submodules (enhancement code). It then reads the strategy report (for new features) or the debug report (for bug fixes) to build an internal "mental map" of all required modifications.

Once the specific target cells are identified, a Planner Parser agent translates this analysis into a structured format (e.g., a JSON list of cell indices). This explicit structural step acts as a binding constraint, ensuring that the subsequent modification phase remains strictly focused on the identified components and does not drift into unrelated parts of the codebase.

Following this, the Planner begins a "cell-by-cell refactoring" process using the coordinates provided by the parser. This methodology is crucial for reliably modifying long and complex codebases, as the agent focuses on only one code block, or "cell," at a time. This targeted approach significantly tackles the risk of hallucination. For each targeted cell, the Planner provides a conversational model explanation of the changes, and a Patcher agent immediately transforms this into a structured JSON patch file. This patch specifies the exact lines of code that must be added, deleted, or modified from the main template.

Once the refactoring loop is complete and the code is assembled, the workflow engages an Inspector Agent. This agent acts as a strict verifier, auditing the final executable code against the original Strategy Report. Its role is to ensure that the final implementation faithfully realizes the requested plan without drift or hallucination before passing it to execution.

\subsubsection{Debugging Team} This team handles the complete execution and validation workflow. A Filing Agent within this team runs only on the first iteration of a new experiment. Its sole job is to assign an appropriate, programmatic name for the project directory. After this first run, all subsequent versioning is handled deterministically.

The script is saved in its designated directory and executed. This leads to a fork in the workflow. If the script crashes, the error code and log files are sent to the Debugger agent. The Debugger is in charge of reading the code and the errors, after which it generates a debug report. This report is sent back to the Implementation Team, which will tackle the errors accordingly and generate a new code. This "debug loop" is repeated iteratively until the code is bug-free and runs successfully.

\subsubsection{Advisor Team} If the code runs successfully (i.e., it is bug-free), it proceeds to the Advisor (Review) Team, which is the core of the second-order or meta-reasoning operator for scientific validation. The executed script must generate a comprehensive set of plots, which typically include loss and error convergence, residuals (for PINNs), the reference solution, the prediction, and pointwise errors.

The Advisor agent's analysis is multimodal. It takes the plots, the scientific strategy report, the user question, and the trial history log as its inputs. This diagnostic task is where the Conceptual Scaffolding is critical. The Advisor is aware of and uses these expert blueprints. It maps its qualitative analysis of the plots (e.g., "high error is focalized in one region") to a specific, principled cure derived from the blueprints (e.g., ``apply adaptive sampling" per the optimization blueprint). Likewise, by observing the loss curve, it can diagnose issues such as a learning rate that is too high.

This feedback is then processed by the Advisor Parser, a utility agent that translates the Advisor's natural language report into a structured format (like JSON) for easy, machine-readable processing. This structured output, along with the strategy report, then goes to the Experiment Logging Team. A Register Agent takes this information and updates the trial history log, which serves as the persistent memory for the entire framework. This complete report and updated history are then sent to both the Strategy Team (to inform the next decision) and the user.

By default, the loop is fully automatic. However, the user can intervene at this point to provide high-level suggestions or guide the strategy before the next cycle begins.

Although ATHENA is implemented in a model-agnostic way at the interface level, the practical allocation of roles is informed by empirical capability differences across foundation models \cite{openai2025gpt51,google2025gemini3,anthropic2025claude45,xai2025grok41}. In our current configuration, Gemini is primarily used in roles that benefit from long-context reasoning, mathematical reliability, and multimodal analysis \cite{google2025gemini3} (e.g., Strategist/Planner and Advisor), while GPT and Grok are typically assigned to critic-style reviewers \cite{openai2025gpt51,xai2025grok41}, where adversarial checking and assumption-challenging behavior is useful for plan validation. Claude is used in roles that emphasize targeted module generation and repair \cite{anthropic2025claude45} (e.g., Librarian and Debugger). For structured parsing and machine-readable outputs (e.g., JSON extraction), we use Grok and local models such as Ollama GPTOSS120B, which empirically produce more consistent constrained formats \cite{xai2025grok41}. While many roles are interchangeable in principle, these assignments reflect two practical constraints in our experiments: long-context planning is most reliable with Gemini \cite{google2025gemini3}, and structured parsing is most reliable with Grok or GPTOSS120B \cite{xai2025grok41}.

\section{Variability Analysis}\label{secA1}
Complementing the analysis in the main text, this section details the variability observed in the Rayleigh-Taylor Instability and Helmholtz PINN benchmarks. Unlike the Burgers' case, the quantitative similarity analysis for these domains (Figure \ref{Variability_analysis}) indicates a higher degree of adherence to the reference scaffolds ($\approx$ 60\%). This behavior is expected, as these problems rely on established high-performance libraries (Trixi.jl and JAX) where the critical architecture, initialization, and hyperparameters are standardized. Consequently, the agents focused on optimizing the implementation within these constraints rather than restructuring the mathematical foundation.

For the Rayleigh-Taylor instability (Table \ref{tab:rt_diversity}), the diversity emerged in engineering trade-offs between stability and resolution. The agents balanced numerical dissipation against mesh refinement, alternating between sharper HLLC fluxes with moderate grid levels and robust HLL fluxes combined with aggressive adaptive mesh refinement. Similarly, in the Helmholtz PINN benchmark (Table \ref{tab:helmholtz_diversity}), the agents converged on a unified neural architecture but explored diverse optimization strategies. Notably, Run 5 demonstrated an efficient "lazy" strategy, where the agent terminated the second-order training phase early once the error threshold was satisfied, distinguishing it from other runs that utilized the full computational budget to reach a loss plateau. These results suggest that even when constrained by library-standard structures, the system successfully navigates implementation trade-offs to ensure robust performance.

\begin{table}[H]
\small
\centering
\caption{
\textbf{Numerical strategy variations in the Rayleigh-Taylor instability benchmark.}
For this complex fluid dynamics problem, ATHENA converged on a unified solver architecture (DGSEM with Hennemann-Gassner shock capturing) across all five runs. The generative diversity manifests in the trade-off between \textit{Numerical Flux} selection and \textit{Adaptive Mesh Refinement (AMR)} limits.
Runs 1, 2, and 4 utilized the HLLC flux for sharper contact resolution with moderate grid refinement (Level 6).
In contrast, Runs 3 and 5 prioritized solver robustness by selecting the more diffusive HLL flux, compensating for the added dissipation by increasing the maximum AMR resolution to Level 7.
}
\label{tab:rt_diversity}
\begin{tabular}{lllll}
\hline
\textbf{Run} & \textbf{Surface Flux} & \textbf{AMR Strategy} & \textbf{Max Level} & \textbf{Mesh Topology} \\
\hline
1 & HLLC (Contact) & Standard Adaptive & 6 & Anisotropic Trees (1,4) \\
2 & HLLC (Contact) & Dense Initialization & 6 & Initial Refinement 5 \\
3 & HLL (Robust) & Aggressive Refinement & 7 & Adaptive Quadtree \\
4 & HLLC (Contact) & Standard Adaptive & 6 & Anisotropic Trees (1,4) \\
5 & HLL (Robust) & Aggressive Refinement & 7 & Isotropic Trees (1,1) \\
\hline
\end{tabular}
\end{table}

\begin{table}[H]
\small
\centering
\caption{
\textbf{Hyperparameter and optimization diversity in Helmholtz PINN solutions.} ATHENA converged on a unified architecture for the Helmholtz problem: an MLP augmented with  Fourier Features to enforce boundary conditions, trained via a Hybrid (Adam + SSBroyden) optimizer with vRBA adaptive sampling.
The generative diversity is primarily found in the \textit{Computational Budget} allocation---specifically the duration of the second-order SSBroyden phase---and the \textit{Input Normalization} logic.
}
\label{tab:helmholtz_diversity}
\begin{tabular}{lllll}
\hline
\textbf{Run} & \textbf{Input Processing} & \textbf{Training Iterations} & \textbf{Hessian Strategy} & \textbf{Architecture} \\
\hline
1 & Raw Embeddings & 75,000  & Resampling (100 steps) & MLP $3 \times 30$ \\
2 & Split coordinate embeddings & 50,000  & Hessian Recycling & MLP $3 \times 30$ \\
3 & Concatenate embeddings & 50,000  & Cholesky Stabilization & MLP $3 \times 30$ \\
4 & Normalization $[-1,1]$ & 50,000  & Hessian Recycling & MLP $3 \times 30$ \\
5 & Normalization $[0,1]$ & 12,500  & Hessian Recycling & MLP $3 \times 30$ \\
\hline
\end{tabular}
\end{table}

\section{Ablation Study}
\setcounter{figure}{0}
This section reports the ablation analysis discussed in the main text, isolating the effect of the conceptual scaffolding by comparing the full ATHENA system against a direct-prompting baseline and a no-scaffolding multi-agent baseline across analytical, numerical, and physics-informed machine-learning tasks.

\begin{figure}[H]
    \centering
    \includegraphics[width=1\linewidth]{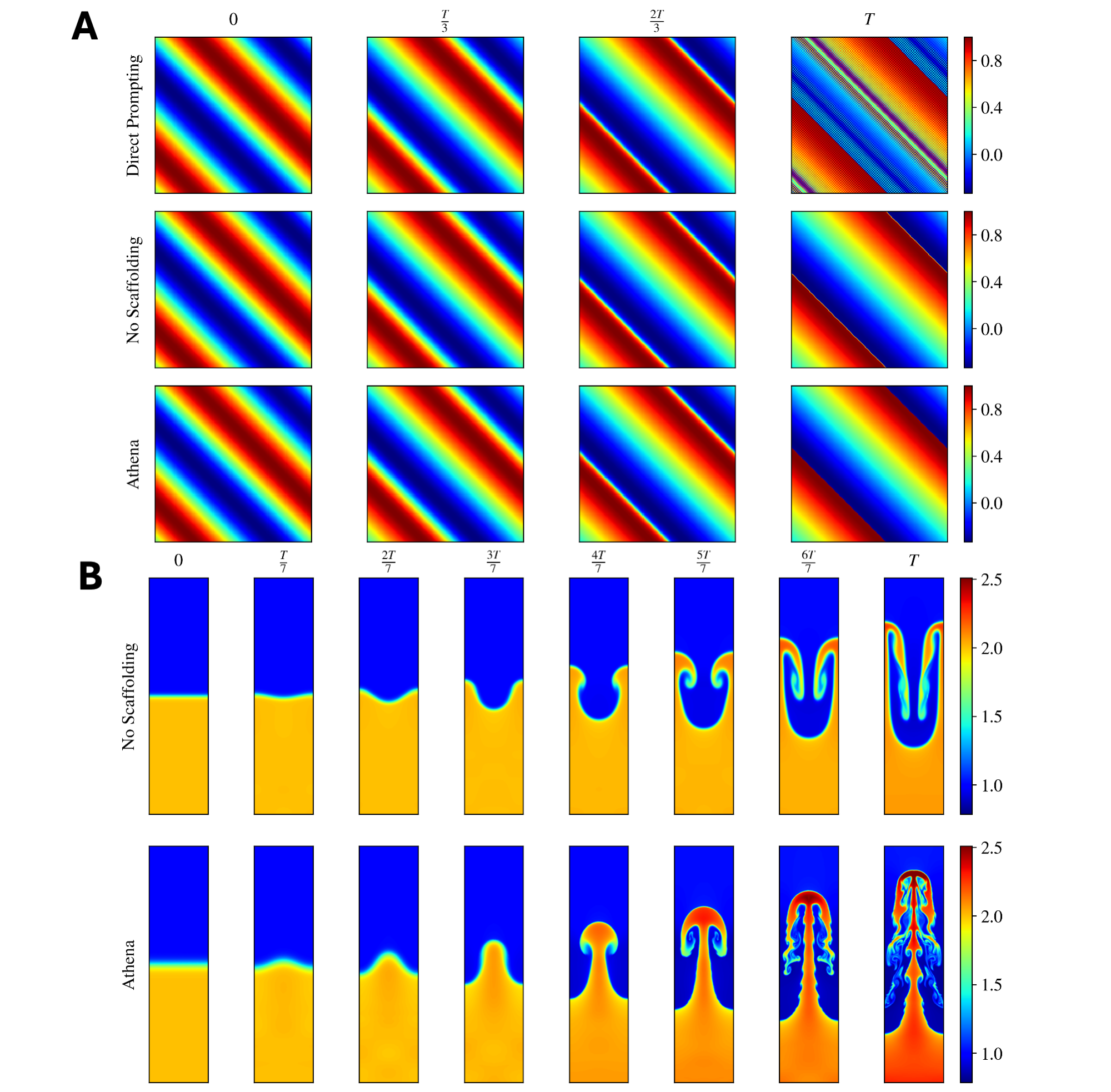}
\caption{\textbf{Ablation analysis (Part I): Analytical solutions and numerical methods} (A) Inviscid Burgers equation. Direct prompting (Gemini 3.0) employs a pseudo-spectral Fourier method with RK4 integration. However, the lack of padding and dealiasing induces significant non-physical oscillations, despite capturing the general transition. The no-scaffolding baseline generates a 2nd-order Finite Volume method (FVM) with a Minmod slope limiter (MUSCL). While functional, the solution is sub-optimal, characterized by a diffusive ``yellow line" artifact at the shock interface. In contrast, ATHENA's strategist agents identify the problem's symmetry, enabling the retrieval of the exact solution via the method of characteristics, which yields a pristine, sharp transition. (B) Rayleigh-Taylor instability. Direct prompting fails to yield a valid solution. The no-scaffolding baseline implements a 2nd-order FVM with MUSCL reconstruction and an HLLC Riemann solver. Due to the complexity of writing the full code without scaffolding, the agents introduce a sign error in the perturbation, resulting in an inverted bubble, while the low-order method causes smearing of fine details. ATHENA successfully deploys a 5th-order Adaptive High-Order Discontinuous Galerkin Spectral Element Method (DGSEM) with AMR on a forest-of-octrees grid and hybrid DG/FV shock capturing, optimally resolving the complex turbulent structures.}
\label{ablation_SiC}
\end{figure}

\begin{figure}[H]
    \centering
    \includegraphics[width=1\linewidth]{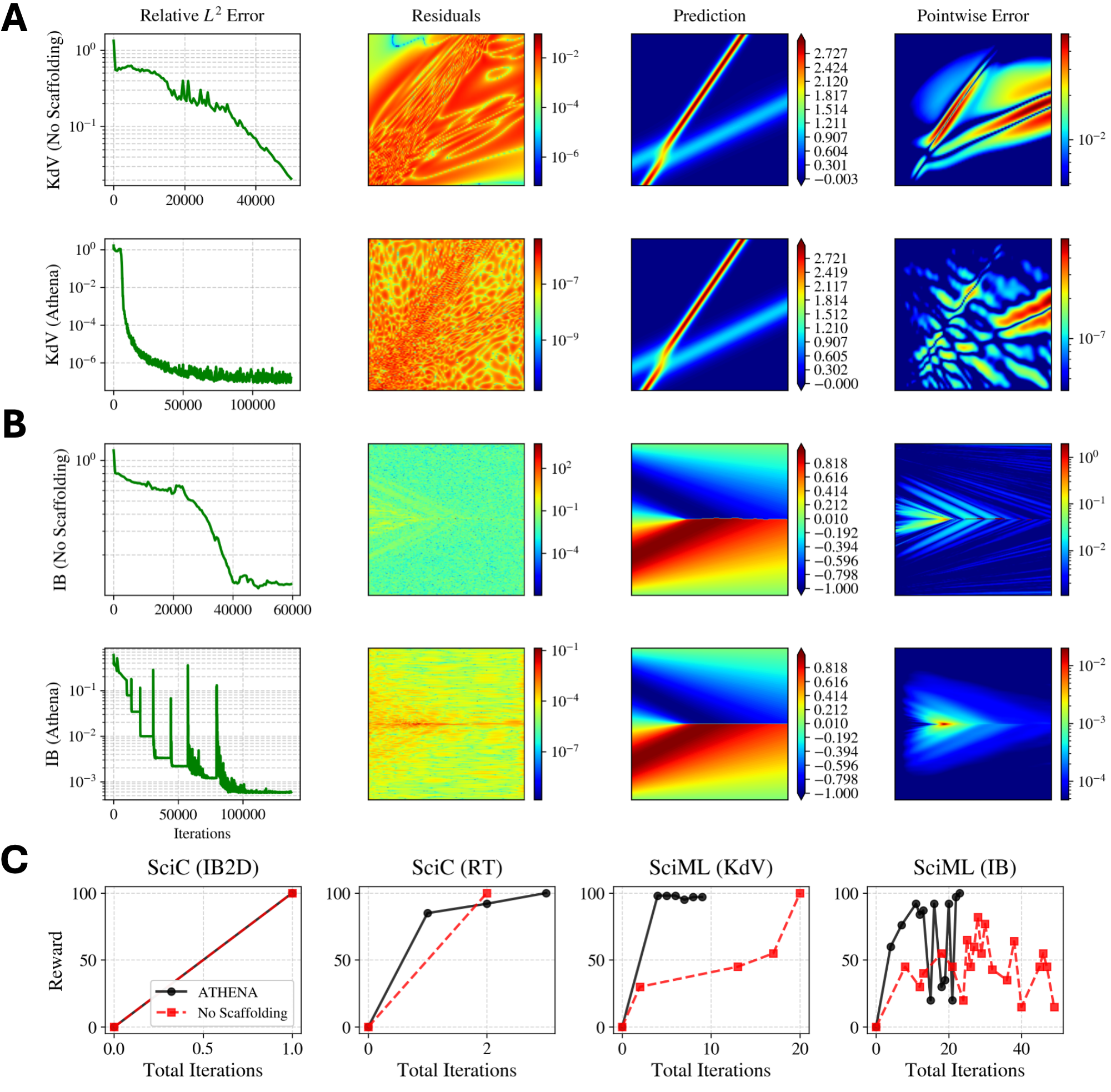}
\caption{\textbf{Ablation study (Part II): Physics-Informed Machine Learning (PIML) tasks and Reward Analysis for all cases.} (A) Korteweg–De Vries (KdV) equation. Direct prompting failed to produce a valid solution. The no-scaffolding baseline employs a suboptimal strategy (Adam, RAR, gradient clipping) with 116,097 parameters. The advisor agent incorrectly identifies the non-converged result (Relative $L^2$ error $2.07 \times 10^{-2}$) as optimal, leading to premature termination. ATHENA employs state-of-the-art optimization (SSBroyden, vRBA), achieving convergence with a relative $L^2$ error of $7.9 \times 10^{-8}$ using only 2,000 parameters. (B) Inviscid Burgers equation (PINNs). The no-scaffolding baseline attempts to implement causality RAR and adaptive activation functions, but these are hallucinated implementations inconsistent with the literature, resulting in a model with 296,586 parameters and a relative error of $1.9 \times 10^{-1}$. ATHENA utilizes viscous continuation, entropy conditions, SSBroyden, and vRBA to achieve a relative $L^2$ error of $5.6 \times 10^{-4}$ with only 3,416 parameters. (C) Reward evolution for SciC tasks (Inviscid Burgers 2D, Rayleigh-Taylor) and SciML tasks (KdV, IB). For the SciC and KdV cases, the no-scaffolding advisor prematurely assigns perfect rewards to suboptimal solutions (e.g., accepting the ``yellow line" artifact or inverted bubble), triggering early termination. In the SciML Inviscid Burgers case, the no-scaffolding approach fails to establish an increasing trend, with the reward wandering aimlessly until the 50-iteration debugging limit is reached without an acceptable run.}
\label{ablation_sciML}
\end{figure}

\section{Auxiliary Field Visualizations}
\setcounter{figure}{0}
This section provides complementary pressure and velocity field visualizations for the compressible-flow benchmarks discussed in the main text (Rayleigh--Taylor and Kelvin--Helmholtz instabilities), supporting the density-field results reported there.

\begin{figure}[H]
    \centering
    \includegraphics[width=1\linewidth]{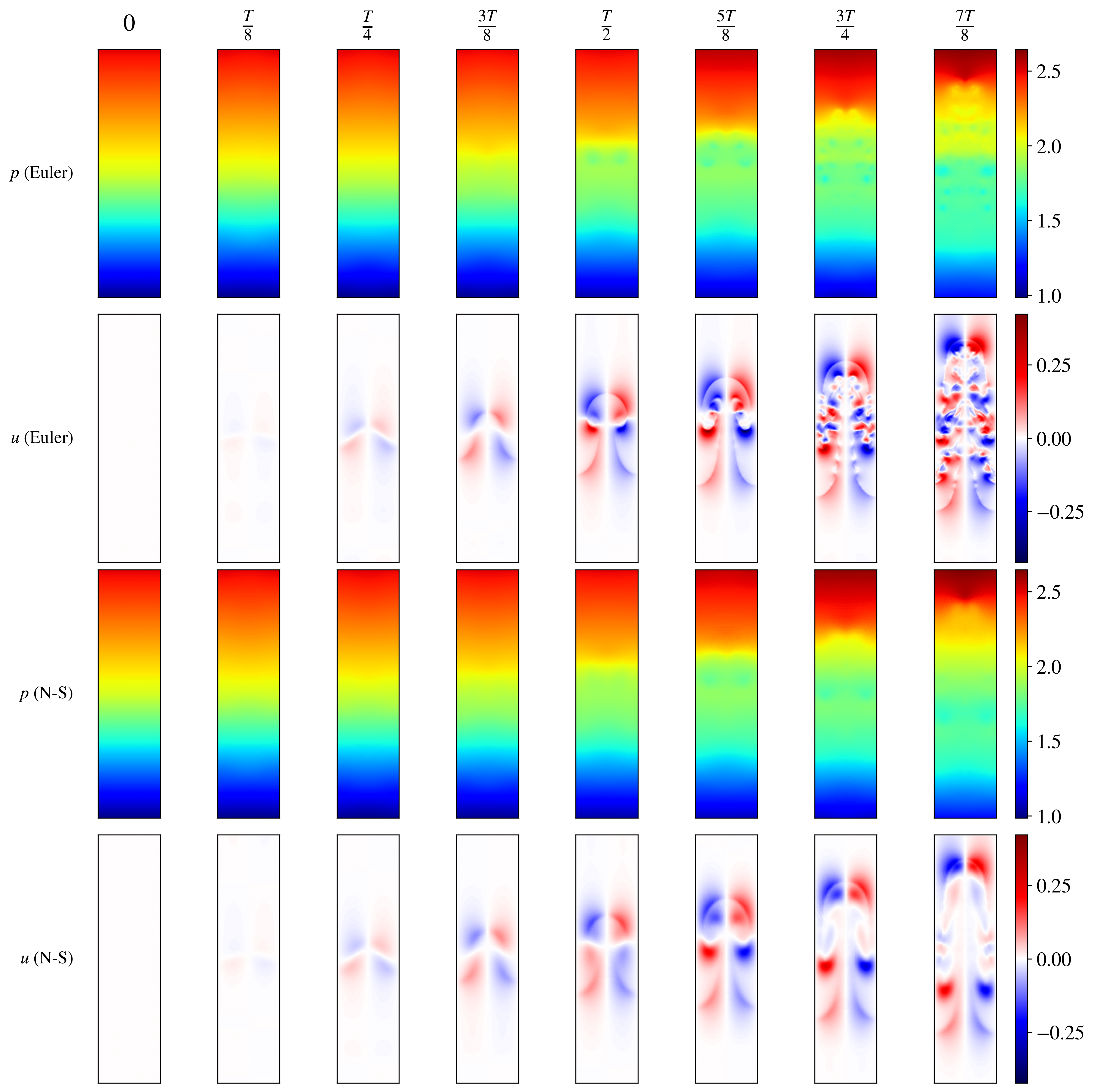}
\caption{\textbf{Rayleigh-Taylor instability auxiliary fields: Inviscid vs. Viscous dynamics.} Temporal evolution of pressure $p$ and velocity $u$ comparing the inviscid Euler regime (top two rows) against the viscous Compressible Navier-Stokes regime (bottom two rows). These results complement the density evolution shown in Figure~\ref{Numerical_Examples}(C, D). In the Euler case, ATHENA enforces hydrostatic balance and utilizes Hennemann--Gassner shock capturing, allowing the solver to sustain sharp gradients and resolve the intricate, fine-scale turbulent mixing visible in the late-time velocity field ($u$, Euler). Conversely, for the Navier-Stokes case, the agent detects the presence of physical viscosity and autonomously adjusts the polynomial degree ($p$-adaptivity). This ensures the accurate resolution of the viscous shear layers, resulting in smoother, coherent structures where physical dissipation naturally suppresses the high-frequency modes observed in the inviscid limit.}
\label{RT_aux}
\end{figure}

\begin{figure}[H]
    \centering
    \includegraphics[width=1\linewidth]{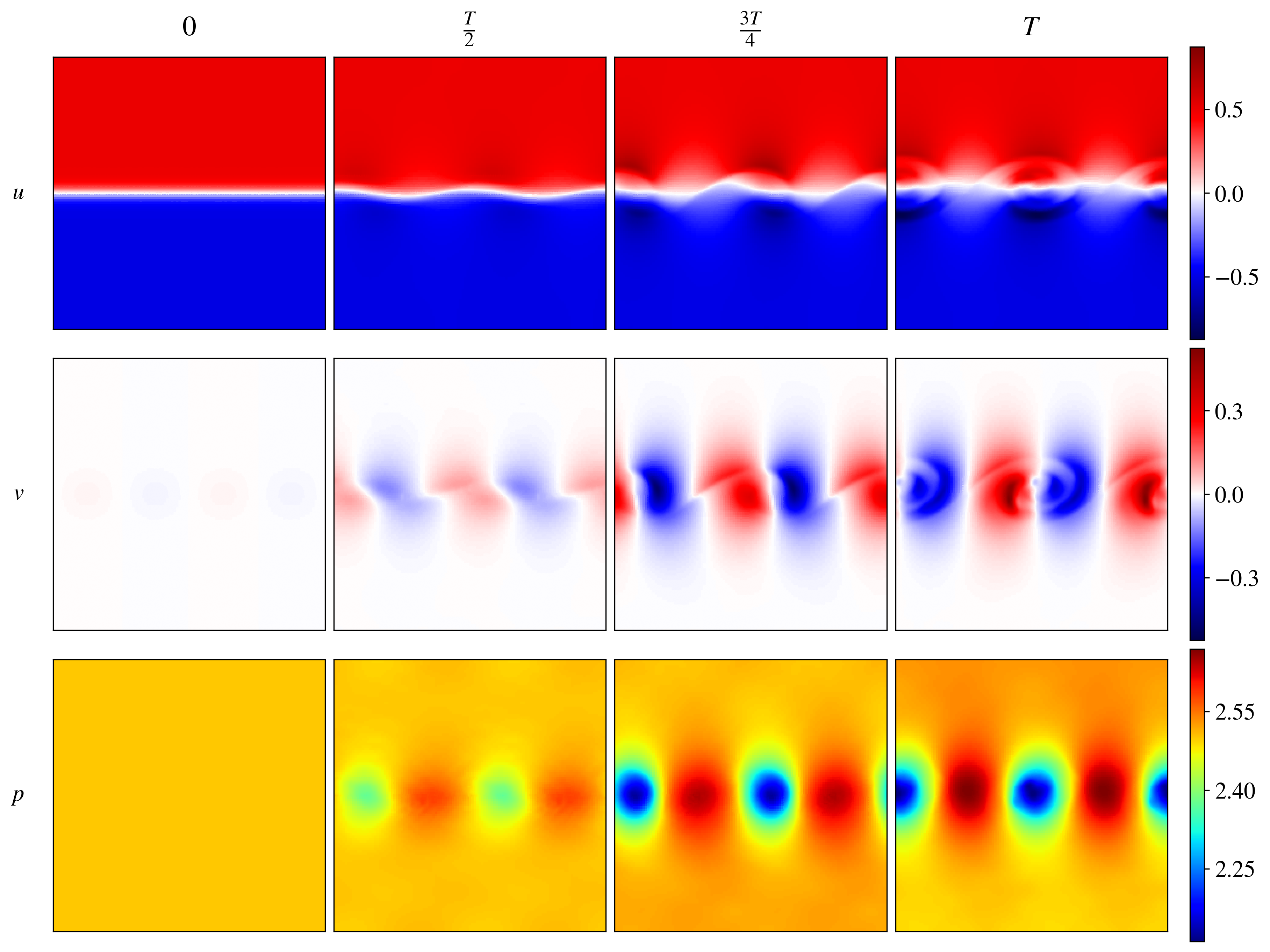}
\caption{\textbf{Kelvin-Helmholtz instability auxiliary fields.} Temporal evolution of the horizontal velocity $u$ (top), vertical velocity $v$ (middle), and pressure $p$ (bottom) at $t=0, T/2, 3T/4,$ and $T$. These results complement the density evolution shown in Figure~\ref{Numerical_Examples}(B). ATHENA's autonomous selection of a high-order Discontinuous Galerkin Spectral Element Method (DGSEM) combined with the HLLC Riemann solver ensures minimal numerical dissipation, maintaining the sharpness of the shear layers in the velocity components. Furthermore, the pressure field demonstrates the accurate resolution of low-pressure vortex cores, free from spurious oscillations, confirming that the density-gradient driven AMR successfully captures the multi-physics dynamics of the mixing layer.}
\label{KH_aux}
\end{figure}

\section{\rev{User Requests (Analytical)}}
\setcounter{figure}{0}

\begin{figure}[H]
    \centering
    \includegraphics[width=1\linewidth]{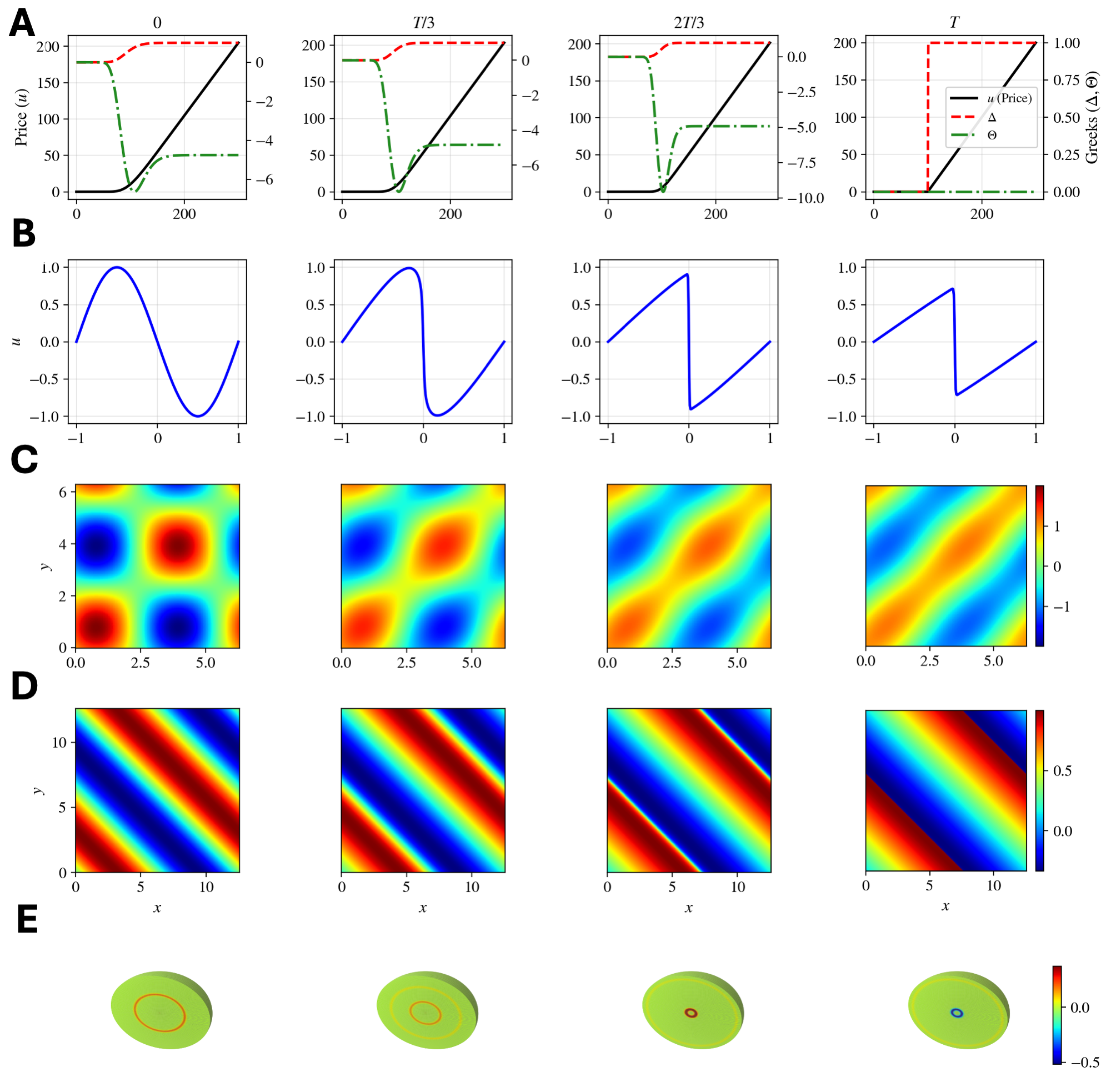}
\caption{
\textbf{ATHENA autonomously uncovers exact analytical solutions via symmetry detection.}
(A) Black-Scholes (Parabolic). ATHENA applied a log-price transformation ($y=\ln S$) to remove drift, reducing the system to the standard Heat Equation.
(B) 1D Viscous Burgers (Non-linear Parabolic). The agent identified the Cole-Hopf transformation ($u = -2\nu \phi_x/\phi$), linearizing the conservation law into a solvable diffusion problem.
(C) 2D Anisotropic Diffusion (Elliptic-Parabolic). The system diagonalized the cross-derivative operator ($\partial_{xy}$) via rotation ($\phi = x+y, \psi = x-y$) to enable exact spectral reconstruction.
(D) 2D Inviscid Burgers (Hyperbolic). Detecting Lagrangian invariance, ATHENA deployed the Method of Characteristics to recover the exact implicit solution, avoiding numerical diffusion.
(E) 3D Wave Equation (Hyperbolic). Exploiting spherical symmetry, the agent applied $v(r,t) = r u(r,t)$ to reduce the 3D Laplacian to a 1D wave operator solved via eigenfunction expansion.
}
\label{Exact_solutions}
\end{figure}

\subsection{\rev{Black-Scholes}}

\subsubsection{User Request}

Let us solve the following variable--coefficient parabolic PDE (backward in time):
\[
\frac{\partial u}{\partial t}
+ \frac{1}{2}\sigma^2 x^2 \frac{\partial^2 u}{\partial x^2}
+ r x \frac{\partial u}{\partial x}
- r u = 0 .
\]

\paragraph{Parameters}
\[
\sigma = 0.2 \quad \text{(volatility)}, \qquad
r = 0.05 \quad \text{(rate)}, \qquad
K = 100 .
\]

\paragraph{Domain}
\[
x \in [0,300] \quad \text{(space)}, \qquad
t \in [0,1] \quad \text{(time)} .
\]

\paragraph{Boundary and Terminal Conditions}
\begin{enumerate}
\item Terminal condition at $t=1$:
\[
u(x,1) = \max(x-K,0).
\]
\item Boundary at $x=0$:
\[
u(0,t) = 0.
\]
\item Boundary at $x=300$:
\[
u(300,t) \approx x - K e^{-r(1-t)}.
\]
\end{enumerate}

\paragraph{Visualization Requirements}
The final plot \texttt{summary\_all.png} must be a $4\times4$ subplot grid showing results at
\[
t = 1.0,\; 0.75,\; 0.5,\; 0.25,\; 0.0 .
\]

\begin{enumerate}
\item First row: Option value $u(x,t)$ versus $x$.
\item Second row: Delta $= \partial u/\partial x$ versus $x$.
\item Third row: Gamma $= \partial^2 u/\partial x^2$ versus $x$.
\item Fourth row: Theta $= \partial u/\partial t$ versus $x$.
\end{enumerate}

Each subplot must include a clear legend indicating time values and appropriate axis labels.

\subsection{Viscous Burgers}

\subsubsection{User Request}

Let's solve the 1D Viscous Burgers' equation; with nu=1/(100pi)

Domain: $(x,) \in [-1,1]$, $t \in [0, 1]$.

BCs: Periodic x

Initial Condition:
\[
u(x,0) = -\sin(\pi x)
\]

The final plot must be saved as \texttt{summary\_all.png} it should include snapshots of several timesteps.

HINT: If you need any numerical integration Use Gauss-Hermite quadrature rule.

\subsection{Anisotropic Diffusion}

\subsubsection{User Request}

Let's solve the 2D Anisotropic Diffusion Equation with nu = 0.25:
\[
\frac{\partial u}{\partial t}
= 0.25\left(
\frac{\partial^2 u}{\partial x^2}
+ 2\frac{\partial^2 u}{\partial x \partial y}
+ \frac{\partial^2 u}{\partial y^2}
\right)
\]

Domain: $(x,y) \in [0, 2\pi]^2$, $t \in [0, 2]$.

BCs: Periodic in both $x$ and $y$.

Initial Condition:
\[
u(x,y,0) = \sin(x+y) + \cos(x-y)
\]

The final plot must be saved as \texttt{summary\_all.png} and include snapshots of the solution field.

\subsection{Inviscid Burgers}

\paragraph{Experiment Report}
\emph{Last Updated: 2025-12-23 12:35:22}

\subsubsection{User Request}

Let's solve the 2D Inviscid Burgers' equation.

Domain: $(x,y) \in [0,4\pi]^2$, $t \in [0,2]$.

BCs: Periodic in both $x$ and $y$.

Initial Condition:
\[
u(x,y,0) = \frac{1}{3} + \frac{2}{3}\sin\!\left(\frac{x+y}{2}\right)
\]

The final plot must be saved as \texttt{summary\_all.png} and include snapshots of several timesteps.

\subsection{Wave Equation (Spherical Symmetry)}

\paragraph{Experiment Report}
\emph{Last Updated: 2026-01-11 12:54:47}

\subsubsection{User Request}

Let's solve the 3D Wave Equation assuming spherical symmetry.

Equation:
\[
\frac{\partial^2 u}{\partial t^2} = c^2 \nabla^2 u
\]
where
\[
\nabla^2 u = \frac{1}{r^2}\frac{\partial}{\partial r}\left(r^2\frac{\partial u}{\partial r}\right).
\]

Parameters: $c = 1.0$

Domain: $r \in [0,10]$, $t \in [0,6]$.

BCs:
\begin{itemize}
\item $u(r,t)$ must be finite at $r=0$.
\item Homogeneous Neumann at $r=10$.
\end{itemize}

Initial Conditions:
\[
u(r,0) = \frac{e^{-10(r-5)^2}}{r}\cdot \sin\left(\frac{\pi r}{10}\right)
\]
\[
\frac{\partial u}{\partial t}(r,0) = 0
\]

The final plot must be saved as \texttt{summary\_all.png} and include snapshots of the solution at $t=0,2,4,6$.

\section{\rev{User Requests (Numerical)}}

\subsection{1D Viscous Burgers' Equation}

\subsubsection{User Request}

Let's solve the 1D Viscous Burgers' equation; with nu=1/(100pi). \\
Domain: $(x,) \in [-1,1]$, $t \in [0, 1]$. \\
BCs: Periodic x

\paragraph{Initial Condition}
\[
u(x,0) = -\sin(\pi x)
\]

The final plot must be saved as \texttt{summary\_all.png} it should include snapshots of several timesteps

\subsection{Cahn-Hilliard}

\subsubsection{User Request}

Let's solve the \textbf{Cahn-Hilliard Equation} in 2D.

\textbf{The Equation:}
\[
u_t = \Delta(u^3 - u) - \epsilon \Delta^2 u
\]

\textbf{Physics Parameters:}
\begin{itemize}
\item $\epsilon = 0.01$.
\item Diffusion coeff = 1.0.
\end{itemize}

\textbf{Domain:}
\begin{itemize}
\item $(x,y) \in [0, 2\pi] \times [0, 2\pi]$
\item Time interval: $t \in [0, 1.0]$
\end{itemize}

\textbf{Resolution}

Lets use a grid of 128x128

\textbf{Boundary Conditions:}
\begin{itemize}
\item \textbf{Periodic} in both X and Y
\end{itemize}

\textbf{Initial Conditions:}
\begin{itemize}
\item Start with a \textbf{small perturbation of the unstable steady state} $u(x) = 0$.
\[
u(x,0) = 0.0 + 0.05 \cdot \text{noise}(x,y)
\]
\end{itemize}

\textbf{Visualization:}

Save a \textbf{heatmap} as \texttt{summary\_all.png}.

\textbf{Validation:}

Please print the Total Mass $\int u \, dA$ at the start and end. It should be conserved to machine precision.

\subsection{Kelvin--Helmholtz Instability (2D Compressible Euler Equations)}

\subsubsection{User Request}

Let's solve the 2D Compressible \textbf{Euler equations} for the \textbf{Kelvin-Helmholtz Instability}.

\paragraph{Physics Parameters:}
\begin{itemize}
    \item $\gamma = 1.4$
    \item Viscosity $\mu = 0.0$ (Inviscid flow)
    \item Gravity $g = 0.0$ (Instability is driven by shear, not buoyancy).
\end{itemize}

\paragraph{Domain:}
$(x,y) \in [0, 1.0] \times [0, 1.0]$ \\
Time interval: $t \in [0, 2.5]$

\paragraph{Boundary Conditions:}
\begin{itemize}
    \item $x$-direction: \textbf{Periodic}
    \item $y$-direction: \textbf{Slip walls} (Reflective, normal velocity $v=0$) at $y=0$ and $y=1$.
\end{itemize}

\paragraph{Initial Conditions:}
The domain is initialized with a smoothed shear layer at $y=0.5$.
\begin{itemize}
    \item \textbf{Density ($\rho$):} Stratified with a smooth transition.
    \[
    \rho(y) = 1.5 - 0.5 \cdot \tanh\left(\frac{y - 0.5}{0.03}\right)
    \]
    (Density varies from 1.0 at the top to 2.0 at the bottom).
    \item \textbf{Pressure ($p$):} Constant uniform pressure.
    \[
    p = 2.5
    \]
    \item \textbf{Velocity:}
    \begin{itemize}
        \item \textbf{Horizontal ($u$):} Opposing flows smoothed by a tanh profile (Shear layer).
        \[
        u(y) = 0.5 \cdot \tanh\left(\frac{y - 0.5}{0.03}\right)
        \]
        \item \textbf{Vertical ($v$):} A single-mode perturbation to trigger the roll-up.
        \[
        v(x,y) = 0.01 \cdot \sin(4\pi x) \cdot \exp\left(-\frac{(y - 0.5)^2}{0.01}\right)
        \]
    \end{itemize}
\end{itemize}

\paragraph{Output:}
Create an appropiate Figure and saved it as \texttt{summary\_all.png}

\subsection{2D Compressible Euler Equations: Rayleigh--Taylor Instability}

\subsubsection{User Request}

Let's solve the 2D Compressible Euler equations for the Rayleigh-Taylor Instability.

\paragraph{Physics Parameters:}
\begin{itemize}
    \item \textbf{Equations:} Compressible Euler ($\gamma = 1.4$).
    \item \textbf{Gravity:} $g = 1.0$ acting in the \textbf{positive $y$-direction} (The source term adds momentum in $+y$).
    \item \textbf{Inviscid:} $\mu = 0$, thermal conductivity $k = 0$.
\end{itemize}

\paragraph{Domain:}
$(x,y) \in [0, 0.25] \times [0, 1.0]$ \\
Time interval: $t \in [0, 3.0]$

\paragraph{Boundary Conditions:}
\begin{itemize}
    \item $x$-direction: \textbf{Periodic}.
    \item $y$-direction: \textbf{Reflective / Free-slip walls} at $y=0$ and $y=1$ (No-penetration condition $v=0$).
\end{itemize}

\paragraph{Initial Conditions:}
The domain is split at the interface $y=0.5$.
\begin{itemize}
    \item \textbf{Density ($\rho$):} A smoothed interface between $\rho_{heavy} = 2.0$ (bottom) and $\rho_{light} = 1.0$ (top).
    \[
    \rho(y) = \rho_{heavy} + (\rho_{light} - \rho_{heavy}) \cdot \frac{1}{2}\left(1 + \tanh\left(50(y - 0.5)\right)\right)
    \]
    \item \textbf{Pressure ($p$):} Hydrostatic balance ($\frac{dp}{dy} = \rho g$).
    \[
    p(y) =
    \begin{cases}
        2y + 1 & \text{if } y < 0.5 \\
        y + 1.5 & \text{if } y \geq 0.5
    \end{cases}
    \]
    \item \textbf{Velocity:} Initialized with zero horizontal velocity ($u=0$) and a perturbation in vertical velocity ($v$):
    \[
    v(x,y) = -0.025 \cdot c \cdot \cos(8\pi x) \cdot \sin(\pi y)^6
    \]
    Where $c = \sqrt{\gamma p / \rho}$ is the local speed of sound.
\end{itemize}

\paragraph{Output:}
The final figure has to be saved as \texttt{summary\_all.png}

\subsection{2D Compressible Navier--Stokes Equations: Rayleigh--Taylor Instability}

\subsubsection{User Request}

Let's solve the 2D Compressible Navier-Stokes equations for the Rayleigh-Taylor Instability.

\paragraph{Physics Parameters:}
\begin{itemize}
    \item $\gamma = 1.4$
    \item Viscosity $\mu = 0.0001$ (Prandtl number $Pr = 0.72$)
    \item Gravity $g = 1.0$ acting in the positive $y$-direction.
\end{itemize}

\paragraph{Domain:}
$(x,y) \in [0, 0.25] \times [0, 1.0]$ \\
Time interval: $t \in [0, 3.0]$

\paragraph{Boundary Conditions:}
\begin{itemize}
    \item $x$-direction: Periodic
    \item $y$-direction: \textbf{No-slip adiabatic walls} at $y=0$ and $y=1$ (Velocity is zero).
\end{itemize}

\paragraph{Initial Conditions:}
The domain is split at $y=0.5$.
\begin{itemize}
    \item \textbf{Density ($\rho$):} A smoothed interface between $\rho_{heavy} = 2.0$ (bottom) and $\rho_{light} = 1.0$ (top).
    \[
    \rho(y) = \rho_{heavy} + (\rho_{light} - \rho_{heavy}) \cdot \frac{1}{2}\left(1 + \tanh\left(50(y - 0.5)\right)\right)
    \]
    \item \textbf{Pressure ($p$):} Hydrostatic balance ($\frac{dp}{dy} = \rho g$).
    \[
    p(y) =
    \begin{cases}
        2y + 1 & \text{if } y < 0.5 \\
        y + 1.5 & \text{if } y \geq 0.5
    \end{cases}
    \]
    \item \textbf{Velocity:} Initialized with zero horizontal velocity ($u=0$) and a perturbation in vertical velocity ($v$):
    \[
    v(x,y) = -0.025 \cdot c \cdot \cos(8\pi x) \cdot \sin(\pi y)^6
    \]
    Where $c = \sqrt{\gamma p / \rho}$ is the local speed of sound.
\end{itemize}

\paragraph{Output:}
The final plot must be saved as \texttt{summary\_all.png} and should include snapshots of several timesteps to visualize the mixing.

\section{\rev{User Requests (SciML-PIML)}}
\subsection{Allen--Cahn Equation}

\subsubsection{User Request}

Now lets solve the Allen--Cahn equation in particular:
\[
u_t - \epsilon u_{xx} + k(u^3 - u) = 0
\]

with initial condition
\[
u(x,0) = x^2 \cos(\pi x).
\]

The domain in $x$ goes from $-1$ to $1$ and in $t$ from $0$ to $1$ with periodic boundary conditions in $x$. Here $\epsilon = 0.0001$ and $k = 5$.

\paragraph{Reference Data}
You can find the data here:
\[
\texttt{LLM/Crunch\_LLM/Documents/codes/Data/allen\_cahn.mat}
\]
The solutions are saved as a dictionary with keys \texttt{`t'}, \texttt{`x'} and \texttt{`usol'}.

\subsection{Viscous Burgers}

\subsubsection{User Request}

Solve the 1D viscous Burgers equation for a single forward transient trajectory with periodic boundary conditions: first generate a high-fidelity numerical reference solution (SciC), then use that solution to train and validate a physics-informed neural network (SciML-PINN).

\paragraph{Equation}
\[
u_t + u\,u_x = \nu\,u_{xx}, \qquad \nu = \frac{1}{100\pi}.
\]

\paragraph{Domain}
\[
x \in [-1,1], \qquad t \in [0,1].
\]

\paragraph{Boundary Conditions}
Periodic in space:
\[
u(-1,t) = u(1,t), \qquad u_x(-1,t) = u_x(1,t), \quad \forall\, t \in [0,1].
\]

\paragraph{Initial Condition}
\[
u(x,0) = -\sin(\pi x), \qquad x \in [-1,1].
\]

\paragraph{Execution Strategy}
\begin{enumerate}
    \item Stage 1 (SciC): solve the equation on the space--time domain with a standard high-fidelity numerical solver, enforcing the periodic boundary and initial conditions, to produce a sufficiently resolved reference solution $u(x,t)$.
    \item Stage 2 (SciML-PINN): set up a PINN mapping $(x,t)\mapsto u(x,t)$, enforcing the PDE, initial condition, and periodic boundary conditions through physics-informed terms, and using the Stage 1 solution as reference data for supervision and validation.
    \item Stage 3 (Evaluation): evaluate the trained PINN at arbitrary $(x,t)$ and compare against the numerical reference to quantify accuracy and identify failure modes near shock-like gradients.
\end{enumerate}

\paragraph{Validation}
The Stage 1 numerical solution serves as ground truth, through pointwise comparison of $u(x,t)$ at multiple time slices, error norms over the space--time grid, and checks of periodicity and smoothness in the PINN output. The Stage 1 solver is resolved finely enough that any discrepancy is attributable to the PINN rather than the reference.

\subsection{Helmholtz Equation}

\subsubsection{User Request}

Lets solve the 2D Helmholtz equation.

The equation is:
\[
u_{xx} + u_{yy} + k^2 u = g(x, y)
\]

where $g(x, y)$ is chosen to satisfy the solution:
\[
u_{an} = \sin(\pi a_1)\sin(\pi a_2)
\]

The spatio-temporal domain is $(x, y) \in [-1, 1] \times [-1, 1]$.

The boundary conditions are \textbf{periodic} in both $x$ and $y$. And let $a_1 = 1$ and $a_2 = 4$ and $k = 1$.

\subsection{Poisson Equation}

\subsubsection{User Request}

Lets solve the 2D Poisson equation.

The equation is:
\[
-\Delta u = f(x, y)
\]

where $f(x, y)$ is chosen to satisfy the solution: $u_{ex} = \sin(4\pi x)\sin(4\pi y)$

The spatio-temporal domain is $(x, y) \in [0, 1] \times [0, 1]$.

The boundary conditions are \textbf{Dirichlet} where $u = u_{ex}$ on the boundary.

\subsection{KdV Equation}

\subsubsection{User Request}

We are solving a forward problem for the Korteweg-De Vries (KdV) equation:
\[
u_t + 6uu_x + u_{xxx} = 0
\]

\paragraph{Domain:}
\begin{itemize}
    \item Spatial: $x \in [0, 20]$
    \item Temporal: $t \in [0, 5]$
\end{itemize}

\paragraph{Boundary Conditions:}
\[
u(0,x) = g_0(x) \quad \text{(Initial Condition)}
\]
\[
u(t,0) = g_1(t) \quad \text{(BC at } x=0\text{)}
\]
\[
u(t,20) = g_2(t) \quad \text{(BC at } x=20\text{)}
\]
\[
u_x(t,20) = g_3(t) \quad \text{(Neumann BC at } x=20\text{)}
\]

\paragraph{Analytical Solution:}
The solution is given by a two-soliton interaction formula:
\[
u(x,t) = \frac{2(c_1 - c_2) \left[ c_1 \cosh^2\left(\frac{\sqrt{c_2}\zeta_2}{2}\right) + c_2 \sinh^2\left(\frac{\sqrt{c_1}\zeta_1}{2}\right) \right]}{\left[ \left(\sqrt{c_1} - \sqrt{c_2}\right) \cosh\left(\frac{\sqrt{c_1}\zeta_1 + \sqrt{c_2}\zeta_2}{2}\right) + \left(\sqrt{c_1} + \sqrt{c_2}\right) \cosh\left(\frac{\sqrt{c_1}\zeta_1 - \sqrt{c_2}\zeta_2}{2}\right) \right]^2}
\]
where:
\begin{itemize}
    \item $\zeta_1 = x - c_1 t - x_1$
    \item $\zeta_2 = x - c_2 t - x_2$
\end{itemize}

\paragraph{Parameters:}
\begin{itemize}
    \item $c_1 = 6.0$ (velocity of first soliton)
    \item $c_2 = 2.0$ (velocity of second soliton)
    \item $x_1 = -2.0$ (initial position of first soliton)
    \item $x_2 = 2.0$ (initial position of second soliton)
\end{itemize}

The boundary conditions $g_0, g_1, g_2, g_3$ are derived by evaluating the analytical solution at the appropriate boundaries.

\subsection{Inviscid Burgers' Equation}

\subsubsection{User Request}

Let's solve the 1D Inviscid Burgers' equation:
\[
u_t + u u_x= 0
\]

\paragraph{Domain:}
$x \in [-1,1]$, $t \in [0, 1]$. \\
BCs: Periodic in $x$.

\paragraph{Initial Condition:}
\[
u(x,y,0) = -\sin(pi x)
\]

Reference Data to compute error metrics: \\
...\_Periodic\_SinPiX/2025\_11\_22\_1730/burgers.npz

The code that generated the data saves burgers.npz in NPZ format containing keys: `t` (200,), `x` (256,), `usol` (200x256).

\section{Additional User Requests}

\subsection{Inverse 2D Steady-State Incompressible Navier-Stokes (Lid-Driven Cavity) (SciML- Step 1)}

\subsubsection{User Request}

Let's solve the inverse 2D steady-state incompressible Navier-Stokes equations (lid-driven cavity):
\[
\nabla \cdot \mathbf{u} = 0
\]
\[
(\mathbf{u} \cdot \nabla)\mathbf{u} = -\nabla p + \frac{1}{Re}\nabla^2 \mathbf{u}
\]

\paragraph{Domain:}
$(x,y) \in [0,1] \times [0,1]$; Here $Re=$unknown.

\paragraph{Boundary Conditions:}
\begin{itemize}
    \item $u(x,1) = 1 - \frac{\cosh(C_0(x-0.5))}{\cosh(0.5 C_0)}$ where the constant is set to $C_0 = 50$.
    \item Bottom wall ($y=0$): $\mathbf{u} = (0, 0)$ (no-slip)
    \item Left wall ($x=0$): $\mathbf{u} = (0, 0)$ (no-slip)
    \item Right wall ($x=1$): $\mathbf{u} = (0, 0)$ (no-slip)
\end{itemize}

\paragraph{Data:}

path: \texttt{LLM/Crunch\_LLM/Documents/codes/Data/ldc\_Re500.npz}

The dataset is unstructured and was saved as an NPZ format containing keys: \texttt{'x', 'y', 'u', 'v', 'p'}
\begin{itemize}
    \item \texttt{'x', 'y'}: Node coordinates (1D arrays)
    \item \texttt{'u', 'v'}: Velocity components at each node (1D arrays, same length as x and y)
    \item \texttt{'p'}: Pressure at each node (1D array, same length as x and y)
\end{itemize}

WE WILL FIND A HIDDEN FIELD FROM OBSERVATIONS. \\
To simulate sparse observations, extract 500 points randomly and add 5 percent noise ONLY TO the x- velocity (u) labels.
\begin{enumerate}
    \item Use those points to infer the true Re.
    \item Morveover we will infer the hidden field y-velocity v.
\end{enumerate}

\paragraph{Error metrics:}
\begin{itemize}
    \item Use the full data (for computing error metrics only )to compute the error of the (u,v) fields.
    \item The true parameter is Re=500.

\end{itemize}

\paragraph{Saving the data:}

Please save the data as npz file including the learned \texttt{'Re'} number, and the sensor points coordinates and values \texttt{'x\_sensor', 'y\_sensor' 'u\_sensor'} and the hidden field at those locations calle it \texttt{'v\_hidden\_at\_sensor'}

\subsection{Inverse 2D Steady-State Incompressible Navier--Stokes (Lid-Driven Cavity) (SciC- Step 2)}

\subsubsection{User Request}

Let’s solve the \textbf{inverse} 2D \textbf{steady-state incompressible Navier--Stokes} equations (lid-driven cavity):
\[
\nabla \cdot \mathbf{u} = 0
\]
\[
(\mathbf{u}\cdot\nabla)\mathbf{u} = -\nabla p + \frac{1}{Re}\nabla^2\mathbf{u}
\]

\textbf{Physics Parameter:}
\begin{itemize}
    \item Reynolds number: \textbf{$Re = 525$}
\end{itemize}

\textbf{Domain:} \\
$(x,y) \in [0,1]\times[0,1]$

\textbf{Boundary Conditions:}
\begin{itemize}
    \item Top lid ($y=1$):
    \[
    u(x,1) = 1 - \frac{\cosh(C_0(x-0.5))}{\cosh(0.5C_0)}, \quad C_0=50
    \]
    \[
    v(x,1)=0
    \]
    \item Bottom wall ($y=0$): $\mathbf{u}=(0,0)$ (no-slip)
    \item Left wall ($x=0$): $\mathbf{u}=(0,0)$ (no-slip)
    \item Right wall ($x=1$): $\mathbf{u}=(0,0)$ (no-slip)
\end{itemize}

\textbf{Cross Validation}

\noindent\rule{\linewidth}{0.4pt}

\paragraph{--- Sensors:}

The data was saved as: \\
path: \texttt{..\_Re\_HiddenField\_v/2025\_11\_27\_0058/}

details: \\
\texttt{Saves 'results.npz' containing keys: 'model\_name' (string), 'all\_errors' (list of \textasciitilde 10000+ floats), 'all\_its' (list), 'all\_loss' (list), 'all\_gamma' (list), 'all\_lamB' (list), 'all\_max\_RBA' (list), 'all\_L\_infty' (list), 'all\_L\_2' (list), 'all\_Re\_est' (list of floats), 'Re' (scalar float), 'x\_sensor' (500,), 'y\_sensor' (500,), 'u\_sensor' (500,), 'v\_hidden\_at\_sensor' (500,).}

This file contains a list of sensors. Once the numerical solution is obtained, please evaluate it on the sensor points and report:
\begin{itemize}
    \item Relative $L_2$ error of $u$ (numerical vs.\ PINN sensors)
    \item Relative $L_2$ error of $v$ (numerical vs.\ PINN sensors)
\end{itemize}

\paragraph{--- True Data:}

path: \texttt{LLM/Crunch\_LLM/Documents/codes/Data/ldc\_Re500.npz}

The dataset is unstructured and was saved as an NPZ format containing keys: \texttt{x}, \texttt{y}, \texttt{u}, \texttt{v}, \texttt{p}
\begin{itemize}
    \item \texttt{x}, \texttt{y}: node coordinates (1D arrays)
    \item \texttt{u}, \texttt{v}: velocity components (1D arrays, same length as \texttt{x} and \texttt{y})
    \item \texttt{p}: pressure at each node (1D array, same length as \texttt{x} and \texttt{y})
\end{itemize}

The true solution is here; please report the $L_2$ error for $u$ and $v$ of the true solution against the new numerical solution.

\paragraph{Output:}
The final plot must be saved as \texttt{summary\_all.png} containing:
\begin{enumerate}
    \item Magnitude of velocity field from the new numerical simulation.
    \item Error maps comparing the numerical solution to the True Data (Re=500).
    \item A scatter plot comparing the PINN predictions at sensors vs.\ the Numerical Solution at those same points.
\end{enumerate}

\noindent\rule{\linewidth}{0.4pt}

\subsection{Viscous Burgers Operator Network (DeepONet) }

\subsubsection{User Request}

Lets solve the viscous burgers equation the domain in x $\in[0,1]$ and $t\in [0,1]$ with periodic boundary conditions. Here $\nu=1/(1000)$.

\subsubsection{Problem}
This will be an operator network that maps the initial condition to the full solution up to t=t\_end (extracted from the data)

\subsubsection{Reference Data:}
You can find the data here: \\
\texttt{LLM/Crunch\_LLM/Data/Operator\_Network\_Examples/Burger\_1000\_5000.mat}

\noindent\rule{\linewidth}{0.4pt}

\section{Reports Examples (Analytical)}

\subsection{Viscous Burgers}

\subsubsection{Latest Strategy Report}
(This was the strategy for the \emph{last successful run})

\textbf{1. Problem Statement}\newline
Compute the solution of 1D viscous Burgers on a periodic domain with \(\nu=\frac{1}{100\pi}\) and \(u(x,0)=-\sin(\pi x)\), for \(t\in[0,1]\), and save snapshot plots to \texttt{summary\_all.png}.

\subsubsection{1. Mathematical Specification (Code-Ready Extraction)}

\begin{itemize}
\item \textbf{Governing Equations:}
\[
\partial_t u(x,t) + u(x,t)\,\partial_x u(x,t) = \nu\,\partial_{xx}u(x,t),
\qquad x\in[-1,1],\ t\in[0,1].
\]

\item \textbf{Domain:}
\[
x\in[-1,1],\quad t\in[0,1].
\]

\item \textbf{Boundary Conditions (Periodic in \(x\)):}
\[
u(-1,t)=u(1,t),\qquad \partial_x u(-1,t)=\partial_x u(1,t)\quad \forall t.
\]

\item \textbf{Initial Conditions:}
\[
u(x,0) = -\sin(\pi x).
\]

\item \textbf{Constants \& Parameters:}
\[
\nu=\frac{1}{100\pi}.
\]

\item \textbf{Expected Physics:}\newline
Nonlinear steepening toward an inviscid-like shock location, but viscosity maintains a smooth, thin internal layer (no discontinuity).
\end{itemize}

\subsubsection{2. Strategic Methodology}

\textbf{Can obtain an exact solution or similar?} \textbf{Yes (analytical/exact).}

We will generate \textbf{ground truth} via the \textbf{Cole--Hopf transform} and an \textbf{exact heat-equation propagator}. This is an analytical, closed-form representation requiring only accurate quadrature for evaluation.

\textbf{Why this satisfies the toolbox ``Method of Characteristics (Analytical/Exact)'' label:}\newline
Viscous Burgers is not solved by classical hyperbolic characteristics directly because diffusion destroys characteristic invariants. However, after Cole--Hopf, the dynamics reduce to the \textbf{linear heat equation}, whose solution is obtained analytically via its fundamental solution (equivalently: Fourier mode propagation). In this lab’s taxonomy, ``Method of Characteristics (Analytical/Exact)'' is the \textbf{umbrella analytical/exact slot} (the only exact slot available) and is used as the template for exact solution implementations; here the exactness comes from analytic reduction + exact linear propagator (no discretization of the PDE).

\paragraph{Exact solution construction (periodic)}

Introduce \(\phi(x,t)\) such that
\[
u(x,t) = -2\nu\,\partial_x \ln \phi(x,t) = -2\nu\,\frac{\phi_x}{\phi}.
\]
Then \(\phi\) satisfies the heat equation
\[
\partial_t \phi = \nu\,\partial_{xx}\phi.
\]

From \(u(x,0)=-\sin(\pi x)\),
\[
u(x,0) = -2\nu \frac{\phi_x(x,0)}{\phi(x,0)}
\quad\Rightarrow\quad
\phi(x,0)=\exp\!\left(-\frac{1}{2\nu}\int^x u(\xi,0)\,d\xi\right).
\]
Choose a periodic-consistent antiderivative:
\[
\int^x u(\xi,0)\,d\xi = \frac{\cos(\pi x)}{\pi}\quad(\text{constant irrelevant}),
\]
hence
\[
\phi(x,0)=\exp\!\left(-\frac{\cos(\pi x)}{2\nu\pi}\right).
\]

On a periodic domain of length \(2\), the heat solution can be written using a periodic heat kernel (image sum):
\[
\phi(x,t)=\int_{-1}^{1} \left[\sum_{m\in\mathbb{Z}}
\frac{1}{\sqrt{4\pi\nu t}}
\exp\!\left(-\frac{(x-y-2m)^2}{4\nu t}\right)\right]\phi(y,0)\,dy.
\]
Then recover
\[
u(x,t)=-2\nu\,\frac{\partial_x \phi(x,t)}{\phi(x,t)},
\]
with \(\partial_x \phi\) obtained by differentiating under the integral (analytic derivative of the kernel).

\textbf{Gauss--Hermite quadrature usage (as hinted):}\newline
Each Gaussian term in the kernel becomes a standard Gaussian-weight integral after the substitution
\[
x-y-2m = \sqrt{4\nu t}\,z,
\]
so integrals reduce to \(\int e^{-z^2} f(z)\,dz\), which is directly compatible with Gauss--Hermite nodes/weights. Truncate the image sum to the minimum \(m\) range needed for machine precision at given \((\nu,t)\).

\subsubsection{3. Implementation Requirements (CRITICAL)}

\begin{itemize}
\item \textbf{Configuration:} use template defaults.
\item \textbf{3.1 Adaptive Mesh Refinement \& Shock Capturing (If DGSEM is selected):} Not applicable (exact analytical evaluation).
\item \textbf{3.1 Advisor-Mandated Adjustments (Overrides):} None.
\end{itemize}

\subsubsection{4. Visualization Strategy}

\begin{itemize}
\item \textbf{Field Selection (Analysis):} plot \textbf{\(u(x,t)\)} only (primary and sufficient).
\item \textbf{Time snapshots:} 4 times, e.g.
\[
t=\{0.0,\ 0.25,\ 0.5,\ 1.0\}.
\]
\item \textbf{Layout:} 1 row (field \(u\)) $\times$ 4 columns (time).
\item \textbf{Saving Data:} save full state \(u(x,t)\) at frequent intervals (e.g. \texttt{dt\_save=0.1}) even if only \(u\) exists.
\item \textbf{Output constraints:}
  \begin{itemize}
  \item Save figure as \textbf{\texttt{summary\_all.png}} (strict).
  \item If using Julia plotting: \textbf{enforce \texttt{colorbar=false}}.
  \item No fixed colormap; use autoscaling.
  \end{itemize}
\end{itemize}

\subsubsection{5. Cross-Validation \& Data Integration (Conditional)}
No external data provided $\to$ none.

\subsubsection{REQUIRED COMPONENTS}

\textbf{Main Method:}\newline
- Method of Characteristics (Analytical/Exact)

\textbf{Enhancements:}\newline
- None

\subsection{Inviscid Burgers}

\subsubsection{Latest Strategy Report}
(This was the strategy for the \emph{last successful run})

\paragraph{1. Problem Statement}

We solve the 2D inviscid Burgers equation on a periodic square with a smooth sinusoidal initial condition, evolved to time $t=2$.

\subsubsection{1. Mathematical Specification (Code-Ready Extraction)}

\paragraph{Governing Equations}

We interpret the 2D inviscid Burgers equation as a scalar conservation law with nonlinear advection velocity equal to the scalar itself in both directions:
\[
\partial_t u(x,y,t)
+ \partial_x\!\left(\frac{u^2}{2}\right)
+ \partial_y\!\left(\frac{u^2}{2}\right)
= 0,
\]
or equivalently,
\[
\partial_t u + u\,\partial_x u + u\,\partial_y u = 0.
\]

\paragraph{Domain}

\[
(x,y) \in [0,4\pi]\times[0,4\pi],\qquad t\in[0,2].
\]

\paragraph{Boundary Conditions}

Periodic in both spatial directions:
\[
u(0,y,t) = u(4\pi,y,t),\qquad
u(x,0,t) = u(x,4\pi,t),
\]
for all admissible $y,x,t$.

\paragraph{Initial Condition}

\[
u(x,y,0) = \frac{1}{3} + \frac{2}{3}\sin\!\left(\frac{x+y}{2}\right).
\]

\paragraph{Constants \& Parameters}

No external parameters beyond domain and time interval. Useful derived quantities:

\begin{itemize}
\item Minimum initial value:
\[
u_{\min}(0) = \frac{1}{3} - \frac{2}{3} = -\frac{1}{3}.
\]
\item Maximum initial value:
\[
u_{\max}(0) = \frac{1}{3} + \frac{2}{3} = 1.
\]
\end{itemize}

\paragraph{Expected Physics}

\begin{itemize}
\item Equation is nonlinear hyperbolic (scalar conservation law).
\item Initial data is smooth and periodic.
\item Characteristics associated with larger $u$ move faster in both $x$ and $y$, so wave steepening and finite-time shock formation are expected.
\item As in 1D Burgers, gradients grow and a shock forms at some $t_c>0$; after $t_c$, the entropy solution contains discontinuities.
\item No viscous smoothing; post-shock solution is piecewise smooth with jump discontinuities.
\end{itemize}

\subsubsection{2. Strategic Methodology}

\paragraph{Exact / Method of Characteristics Feasibility}

The PDE can be written as
\[
\partial_t u + \mathbf{a}(u)\cdot\nabla u = 0,
\qquad
\mathbf{a}(u) = (u,u).
\]

Along characteristics $(X(s),Y(s),T(s))$,
\[
\frac{dX}{ds} = u,\quad
\frac{dY}{ds} = u,\quad
\frac{dT}{ds} = 1,\quad
\frac{d}{ds}u = 0.
\]

Hence $u$ is constant along characteristics. Integrating,
\[
X(t) = \xi + t\,u_0(\xi,\eta),\qquad
Y(t) = \eta + t\,u_0(\xi,\eta),
\]
with
\[
u_0(\xi,\eta) = \frac{1}{3} + \frac{2}{3}\sin\!\left(\frac{\xi+\eta}{2}\right).
\]

The exact solution is implicitly defined by
\[
u(x,y,t) = u_0(\xi,\eta),\quad
x = \xi + t\,u_0(\xi,\eta),\quad
y = \eta + t\,u_0(\xi,\eta),
\]
with periodicity modulo $4\pi$.

Characteristic intersection occurs when
\[
\det\!\left(\frac{\partial(x,y)}{\partial(\xi,\eta)}\right) = 0,
\]
signaling shock formation. After that time, the entropy solution must be selected.

\begin{itemize}
\item Yes, a characteristic-based analytical representation exists.
\item It is implicit and requires entropy selection.
\item The entropy-satisfying branch is unique for this convex flux.
\end{itemize}

\paragraph{Entropy Condition / Root Selection}

\begin{itemize}
\item Before first shock: mapping is one-to-one.
\item After shock:
  \begin{itemize}
  \item Multiple $(\xi,\eta)$ may map to the same $(x,y)$.
  \item The entropy solution selects the physically admissible branch (Lax condition).
  \end{itemize}
\end{itemize}

\subsubsection{3. Implementation Requirements (CRITICAL)}

\paragraph{Configuration}

Use template defaults for spatial resolution and CFL. Domain is $[0,4\pi]^2$ with $t_{\text{final}}=2$.

If implemented numerically, suitable choices include Fourier spectral or 2D DGSEM, but the main template is MoC.

\paragraph{Adaptive Mesh Refinement \& Shock Capturing}

If DGSEM is used, select the scalar field $u(x,y,t)$ itself as the AMR and shock indicator.

\paragraph{Advisor-Mandated Adjustments}

None.

\subsubsection{4. Visualization Strategy}

\paragraph{Field Selection}

Primary and only field: $u(x,y,t)$.

\paragraph{Snapshots / Times}

Select four representative times:
\[
t = 0.0,\quad t \approx 0.5,\quad t \approx 1.0,\quad t = 2.0.
\]

\paragraph{Layout}

\begin{itemize}
\item 1 row, 4 columns.
\item Each subplot: 2D color plot of $u(x,y,t)$.
\item Save as \texttt{summary\_all.png}.
\item Disable colorbars.
\item Automatic scaling.
\end{itemize}

\paragraph{Data Saving}

Save the full solution $u(x,y,t)$ at regular intervals (e.g.\ \texttt{dt\_save = 0.1}).

\subsubsection{5. Cross-Validation \& Data Integration}

No external data provided. No error computation required.

\subsubsection{Required Components}

\paragraph{Main Method}
Method of Characteristics (Analytical/Exact).

\paragraph{Enhancements}
Shock capturing and AMR on $u$ (if numerical solver is used).

\section{Reports Examples (Numerical)} 
\subsection{1D Viscous Burgers' Equation}

\subsubsection{Latest Strategy Report}
(This was the strategy for the \textit{last successful run})

\paragraph{1. Problem Statement}
Solve the \textbf{1D viscous Burgers’ equation} on a periodic domain with smooth sinusoidal initial data and small viscosity $\nu=\frac{1}{100\pi}$. Primary physics: nonlinear steepening toward a shock-like gradient, regularized by viscosity into thin layers.

\paragraph{1. Mathematical Specification (Code-Ready Extraction)}

\paragraph{Governing Equations}
\[
\partial_t u(x,t) + u(x,t)\,\partial_x u(x,t) = \nu\,\partial_{xx}u(x,t),
\qquad x\in[-1,1],\ t\in[0,1].
\]

\paragraph{Domain}
\[
x\in[-1,1],\qquad t\in[0,1].
\]

\paragraph{Boundary Conditions}
Periodic in $x$:
\[
u(-1,t)=u(1,t),\quad \partial_x u(-1,t)=\partial_x u(1,t)\quad \forall t.
\]

\paragraph{Initial Condition}
\[
u(x,0) = -\sin(\pi x).
\]

\paragraph{Constants \& Parameters}
\[
\nu=\frac{1}{100\pi}.
\]

\paragraph{Expected Physics}
Nonlinear wave steepening from the sinusoid; viscosity prevents discontinuities and forms \textbf{thin viscous layers} (sharp gradients) while remaining bounded $u=O(1)$.

\paragraph{2. Strategic Methodology}

\paragraph{Can obtain an exact solution or similar?}
\textbf{Analytically yes (Cole--Hopf transform), but numerically no for this $\nu$} in standard floating-point without specialized stabilization. The advisor’s forensics are decisive: exponential dynamic range makes $\phi$ nearly singular/flat, and reconstruction $u=-2\nu\,\phi_x/\phi$ catastrophically amplifies truncation noise.

\paragraph{Selected ONE best method}
\textbf{Fourier Spectral Method (Numerical)} (periodic domain; high accuracy for smooth/near-smooth solutions; avoids Cole--Hopf conditioning pathologies).

Core numerical strategy:
\begin{itemize}
    \item Spatial derivatives via FFT: $\partial_x$ and $\partial_{xx}$ in Fourier space.
    \item Time integration: explicit RK4 (or similarly robust explicit method) with a CFL-aware small $dt$.
    \item Nonlinear term computed in physical space, transformed back (standard pseudospectral).
\end{itemize}

\paragraph{3. Implementation Requirements (CRITICAL)}

\paragraph{Configuration}
Use template defaults \textbf{except} the advisor-mandated changes below.

\paragraph{3.1 Advisor-Mandated Adjustments (Overrides)}
\begin{itemize}
    \item \textbf{Mandatory:} Abandon Cole--Hopf ``exact'' method.
    \item \textbf{Mandatory:} Use \textbf{direct PDE solve} with \textbf{Fourier Spectral} on periodic domain.
    \item Use \textbf{$N=256$ or $N=512$} (advisor recommendation). Prefer $N=512$ if runtime allows.
    \item Use a \textbf{robust explicit integrator} (e.g.\ RK4) with CFL-safe $dt$ (small fixed $dt$ is acceptable).
\end{itemize}

\paragraph{4. Visualization Strategy}

\paragraph{Field Selection (Analysis)}
Plot \textbf{only $u(x,t)$}. It fully represents steepening and viscous regularization; no other fields exist.

\paragraph{Plot Requirement}
\begin{itemize}
    \item Snapshots at \textbf{4 distinct time steps}:
    \[
    t\in\{0,\ 0.25,\ 0.5,\ 1.0\}.
    \]
    \item Layout: \textbf{1 row (u), 4 columns (time snapshots)}.
    \item Save as: \textbf{\texttt{summary\_all.png}}.
\end{itemize}

\paragraph{Data Saving}
Save \textbf{full solution state} $u(x,t)$ at frequent intervals (e.g.\ \texttt{dt\_save=0.1}) for reproducibility and ground truth.

\paragraph{Julia Plotting Constraint}
\textbf{Disable colorbars}: \texttt{colorbar=false}.

\paragraph{5. Cross-Validation \& Data Integration}
No external reference data paths or sensors provided $\rightarrow$ \textbf{none required}.

\paragraph{REQUIRED COMPONENTS}

\textbf{Main Method}
\begin{itemize}
    \item Fourier Spectral Method (Numerical)
\end{itemize}

\textbf{Enhancements}
\begin{itemize}
    \item Advisor-mandated abandonment of Cole--Hopf transform
    \item Direct PDE DNS with Fourier spectral discretization
    \item CFL-safe explicit time stepping (RK4)
\end{itemize}

\subsection{2D Compressible Euler Equations: Rayleigh--Taylor Instability}

\subsubsection{Latest Strategy Report}
(This was the strategy for the \textit{last successful run})

\paragraph{1. Problem Statement}

\paragraph{1. Mathematical Specification}
\begin{itemize}
    \item \textbf{Governing Equations:} \\
    The 2D Compressible Euler equations with a gravity source term in the positive $y$-direction:
    \[
    \frac{\partial}{\partial t}
    \begin{pmatrix}
    \rho \\ \rho u \\ \rho v \\ \rho E
    \end{pmatrix}
    +
    \frac{\partial}{\partial x}
    \begin{pmatrix}
    \rho u \\ \rho u^2 + p \\ \rho u v \\ (\rho E + p)u
    \end{pmatrix}
    +
    \frac{\partial}{\partial y}
    \begin{pmatrix}
    \rho v \\ \rho u v \\ \rho v^2 + p \\ (\rho E + p)v
    \end{pmatrix}
    =
    \begin{pmatrix}
    0 \\ 0 \\ \rho g \\ \rho v g
    \end{pmatrix}
    \]
    where
    \[
    E = \frac{p}{\rho(\gamma-1)} + \frac{1}{2}(u^2 + v^2).
    \]

    \item \textbf{Domain:}
    \[
    (x,y) \in [0, 0.25] \times [0, 1.0]
    \]

    \item \textbf{Boundary Conditions:}
    \begin{itemize}
        \item $x$-boundaries ($x=0, x=0.25$): \textbf{Periodic}
        \item $y$-boundaries ($y=0, y=1$): \textbf{Slip Wall / Reflective} ($v=0$)
    \end{itemize}

    \item \textbf{Initial Conditions:}
    \begin{itemize}
        \item \textbf{Density:}
        \[
        \rho(x,y) = 2.0 + (1.0 - 2.0) \cdot \frac{1}{2}\left(1 + \tanh\left(50(y - 0.5)\right)\right)
        \]
        \item \textbf{Pressure:}
        \[
        p(x,y) =
        \begin{cases}
            2y + 1 & \text{if } y < 0.5 \\
            y + 1.5 & \text{if } y \geq 0.5
        \end{cases}
        \]
        \item \textbf{Velocity:}
        \[
        u(x,y) = 0
        \]
        \[
        v(x,y) = -0.025 \cdot \sqrt{\frac{\gamma p}{\rho}} \cdot \cos(8\pi x) \cdot \sin(\pi y)^6
        \]
    \end{itemize}

    \item \textbf{Constants \& Parameters:}
    \begin{itemize}
        \item $\gamma = 1.4$
        \item $g = 1.0$ (Gravity acceleration magnitude)
        \item Gravity Vector $\mathbf{g} = [0, 1.0]^T$
        \item $t_{final} = 3.0$
    \end{itemize}

    \item \textbf{Expected Physics:} \\
    Rayleigh-Taylor Instability driven by ``upward'' gravity acting on a heavy fluid at the bottom pushing into a light fluid at the top. Formation of mushrooms caps and turbulent mixing at the interface.
\end{itemize}

\paragraph{2. Strategic Methodology}
\begin{itemize}
    \item \textbf{Selected Main Method:} Discontinuous Galerkin Spectral Element Method in 2D plus time (DGSEM).
    \item \textbf{Rationale:} DGSEM provides the high-order accuracy required to resolve the fine-scale vortices and turbulent mixing inherent in the Rayleigh-Taylor instability, while shock capturing stabilizes the steep density gradients.
\end{itemize}

\paragraph{3. Implementation Requirements}
\begin{itemize}
    \item \textbf{Configuration:} Use template defaults, subject to the overrides below.
\end{itemize}

\paragraph{3.1 Adaptive Mesh Refinement \& Shock Capturing:}
\begin{itemize}
    \item \textbf{Variable Selection Analysis:} The Rayleigh-Taylor instability is fundamentally driven by the density difference across the interface. The density field contains the sharpest gradients (the contact discontinuity) and defines the mixing zone structure. Velocity becomes chaotic and noisy as turbulence develops, making it a poor candidate for refinement.
    \item \textbf{Selected Variable:} \textbf{Density ($\rho$)}.
    \item \textbf{Justification:} Refining on density ensures the mesh focuses on the evolving interface and the ``mushroom'' structures, which are the primary features of interest, without wasting resources on global acoustic waves.
\end{itemize}

\paragraph{3.2 Advisor-Mandated Adjustments (Overrides):}
\begin{itemize}
    \item \textbf{Surface Flux:} \texttt{flux\_hllc} (Replaces \texttt{flux\_hll} to resolve contact discontinuities).
    \item \textbf{Polynomial Degree:} \texttt{polydeg = 4} (Increased from 3 for higher accuracy).
    \item \textbf{Mesh Resolution:} Increase \texttt{initial\_refinement\_level} by 1 or 2 levels compared to the previous trial (e.g., set to 4 or 5 depending on base mesh).
\end{itemize}

\paragraph{4. Visualization Strategy}
\begin{itemize}
    \item \textbf{Field Selection:}
    \begin{itemize}
        \item \textbf{Primary Field:} \textbf{Density ($\rho$)}.
        \item \textbf{Reasoning:} Density is the defining variable for RTI, clearly showing the heavy/light fluid mixing and the interface topology.
    \end{itemize}
    \item \textbf{Task:} Plot Density at 4 distinct time steps (e.g., $t=0.0, 1.0, 2.0, 3.0$).
    \item \textbf{Layout:} Grid (1 Row, 4 Columns).
    \item \textbf{Saving Data:} Save full solution state every \texttt{dt\_save=0.1}.
    \item \textbf{Constraints:}
    \begin{itemize}
        \item Save as: \texttt{summary\_all.png}.
        \item \textbf{CRITICAL:} \texttt{colorbar=false} (Disable colorbars).
        \item Style: Automatic scale adjustment.
    \end{itemize}
\end{itemize}

\paragraph{5. Cross-Validation \& Data Integration}
\begin{itemize}
    \item \textbf{Status:} No external reference data provided.
\end{itemize}

\paragraph{REQUIRED COMPONENTS}
\textbf{Main Method:} \\
Discontinuous Galerkin Spectral Element Method in 2D plus time

\textbf{Enhancements:}
\begin{itemize}
    \item Shock Capturing (Indicator: Density)
    \item Source Terms (Gravity)
\end{itemize}

\section{Reports Examples (SciML-PIML)}
\subsection{KdV Equation}

\subsubsection{Latest Strategy Report}
(This was the strategy for the \textit{last successful run})

\subsubsection{Analysis Steps}

\paragraph{0. Triage - Determine Strategy}
\begin{itemize}
    \item \textbf{User Request:} Forward problem for KdV equation (1D+t).
    \item \textbf{Advisor Report:} \textbf{REJECTED\_FOR\_REFINEMENT}.
    \item \textbf{Recommendation:} The previous run (Run 7) was excellent (Error $\sim 1.7\mathrm{e}{-7}$) but exhibited oscillations in the final SSBroyden phase.
    \item \textbf{Mandatory Action:} \textbf{Increase the SSBroyden resampling interval (\texttt{N\_change}) from 100 to 200} to allow the optimizer to stabilize on the loss landscape before collocation points are moved.
    \item \textbf{Critic Response:} The Critic noted that changing \texttt{N\_change} alters the sampling dynamics. I justify retaining the total iteration count (200,000) because the Advisor's specific goal is to \textit{reduce the frequency} of resampling events to dampen oscillations. Halving the number of resampling events (by doubling the dwell time) is the direct mathematical implementation of this recommendation.
\end{itemize}

\paragraph{1. Data Source}
\begin{itemize}
    \item \textbf{Role:} Validation/Error Metrics ONLY.
    \item \textbf{Instruction:} Use the provided analytical solution for validation.
\end{itemize}

\paragraph{2. Problem Statement}
\begin{itemize}
    \item \textbf{Governing Equation:} Korteweg-De Vries (KdV)
    \[
    u_t + 6uu_x + u_{xxx} = 0
    \]
    \begin{itemize}
        \item Residual: $R = u_t + 6 \cdot u \cdot u_x + u_{xxx}$
    \end{itemize}
    \item \textbf{Boundary/Initial Conditions:}
    \begin{itemize}
        \item IC: $u(0,x) = g_0(x)$ (Analytical solution at $t=0$)
        \item BC Left (Dirichlet): $u(t,0) = g_1(t)$ (Analytical at $x=0$)
        \item BC Right (Dirichlet): $u(t,20) = g_2(t)$ (Analytical at $x=20$)
        \item BC Right (Neumann): $u_x(t,20) = g_3(t)$ (Analytical derivative at $x=20$)
    \end{itemize}
    \item \textbf{Analytical Solution:}
    \[
    u(x,t) = \frac{2(c_1 - c_2) [ c_1 \cosh^2(\frac{\sqrt{c_2}\zeta_2}{2}) + c_2 \sinh^2(\frac{\sqrt{c_1}\zeta_1}{2}) ]}{ [ (\sqrt{c_1} - \sqrt{c_2}) \cosh(\frac{\sqrt{c_1}\zeta_1 + \sqrt{c_2}\zeta_2}{2}) + (\sqrt{c_1} + \sqrt{c_2}) \cosh(\frac{\sqrt{c_1}\zeta_1 - \sqrt{c_2}\zeta_2}{2}) ]^2}
    \]
    where $\zeta_1 = x - 6.0t - (-2.0)$ and $\zeta_2 = x - 2.0t - 2.0$.
    \item \textbf{Problem Type:} Forward.
    \item \textbf{Input:} 2D ($t, x$). Domain: $t \in [0, 5], x \in [0, 20]$.
    \item \textbf{Output:} Scalar $u$.
\end{itemize}

\paragraph{3. Base Model Selection}
\begin{itemize}
    \item \textbf{Type:} Low Dim (1D+t).
    \item \textbf{Optimizer:} Hybrid Adam + SSBroyden.
    \item \textbf{Base MLP Specs:} \texttt{width=30}, \texttt{depth=3}.
\end{itemize}

\paragraph{4. Main Architecture}
\begin{itemize}
    \item \textbf{Structure:} Input $\to$ Normalization $\to$ Embedding $\to$ Main Body $\to$ Output.
    \item \textbf{Input Transformation:}
    \begin{itemize}
        \item \textbf{Normalization:} Normalize $t \in [0, 5] \to [0, 1]$ and $x \in [0, 20] \to [0, 1]$ \textbf{inside} the model's \texttt{forward} method.
        \item \textbf{Feature Embeddings:} \textbf{Option B: Random Fourier Features}.
        \begin{itemize}
            \item \textbf{Specs:} \texttt{degree=5}, \texttt{scale=3.0}.
            \item \textbf{Composition:} Pass \textbf{ONLY} the embeddings to the network (do not concatenate raw inputs).
        \end{itemize}
    \end{itemize}
    \item \textbf{Main Body:}
    \begin{itemize}
        \item MLP with \texttt{width=30}, \texttt{depth=3}. Activation: \texttt{tanh}.
    \end{itemize}
    \item \textbf{Output Transformation:}
    \begin{itemize}
        \item Direct output $u$.
    \end{itemize}
\end{itemize}

\paragraph{5. Constraint Implementation Strategy}
\begin{itemize}
    \item \textbf{Count:} $L_{pde}$ (1) + $L_{ic}$ (1) + $L_{bc\_dirichlet}$ (2) + $L_{bc\_neumann}$ (1) = 5 terms.
    \item \textbf{Strategy:} \textbf{Loss-Based (Soft Constraints)}.
    \begin{itemize}
        \item \textit{Reasoning:} While the count is $>4$, Run 7 proved that soft constraints with high weighting work perfectly for this problem. We will maintain the successful strategy.
    \end{itemize}
\end{itemize}

\paragraph{6. Loss Function Assembly}
\begin{itemize}
    \item \textbf{Symbolic Equation:}
    \[
    L_{total} = L_{pde} + L_{ic} + L_{bc\_dirichlet} + L_{bc\_neumann}
    \]
    \item \textbf{Components:}
    \begin{itemize}
        \item $L_{pde}$: MSE of Residual.
        \item $L_{ic}$: MSE against $g_0(x)$.
        \item $L_{bc\_dirichlet}$: MSE against $g_1(t)$ and $g_2(t)$.
        \item $L_{bc\_neumann}$: MSE against $g_3(t)$.
    \end{itemize}
\end{itemize}

\paragraph{6.1 Continuation Strategy}
\begin{itemize}
    \item \textbf{None.} Direct solve is effective.
\end{itemize}

\paragraph{7. Weighting Strategy}
\begin{itemize}
    \item \textbf{Local Weights (`lambdas`):} Use \textbf{vRBA}.
    \begin{itemize}
        \item \textbf{Adam Phase:} Use for \textbf{Weighting \& Sampling}.
        \item \textbf{SSBroyden Phase:} Use for \textbf{Sampling ONLY}.
    \end{itemize}
    \item \textbf{Global Weights (`lam`):}
    \begin{itemize}
        \item \textbf{Strategy:} Fixed Global Weights.
        \item \textbf{Values:} $L_{pde} = 1.0$, All BCs/ICs = $100.0$.
        \item \textbf{Instruction:} \textbf{FREEZE} global weights. Do not update them dynamically.
    \end{itemize}
\end{itemize}

\paragraph{Optimization Schedule (Specifics):}
\begin{itemize}
    \item \textbf{Phase 1 (Adam):} 5,000 iterations.
    \item \textbf{Phase 2 (SSBroyden):} 200,000 iterations.
    \begin{itemize}
        \item \textbf{CRITICAL ADVISOR INSTRUCTION:} Set the resampling interval \textbf{\texttt{N\_change = 200}} (increase from 100).
        \item \textbf{Justification:} We maintain the total iteration count (200k) to preserve the training budget. Increasing \texttt{N\_change} to 200 reduces the \textit{frequency} of resampling events (halving the total number of resamples), which directly addresses the Advisor's requirement to allow the optimizer more time to stabilize on a fixed landscape between shifts.
        \item \textbf{Reinitialization:} Reinitialize Hessian on every resampling step.
    \end{itemize}
\end{itemize}

\paragraph{8. Visualization Strategy}
\begin{itemize}
    \item Standard plots (Solution, Error, Residuals).
\end{itemize}

\paragraph{9. Saving Data}
\begin{itemize}
    \item Save \texttt{u\_pred}, \texttt{u\_exact}, \texttt{x}, \texttt{t} as \texttt{.npz}.
\end{itemize}

Main Model:
[MLP with Hybrid Adam and SSBrodyden]

Enhancements:
[Random Fourier Features]

\subsection{Inviscid Burgers' Equation}

\subsubsection{Latest Strategy Report}
(This was the strategy for the \textit{last successful run})

\subsubsection{0. Triage - Determine Strategy}
\begin{itemize}
    \item \textbf{Strategy:} \textbf{Refinement (Advisor-Driven)}.
    \item \textbf{Advisor Recommendation:} The previous run (Trial 21) was excellent (Score 97) but showed incomplete convergence in the final stage. The Advisor explicitly requests increasing the dwell time for the final viscosity step ($\nu=0.0001$) from 50,000 to \textbf{80,000} iterations.
    \item \textbf{Action:} Replicate the successful architecture (Chebyshev KAN) and methodology of Trial 21, modifying \textbf{only} the final step of the continuation schedule.
\end{itemize}

\subsubsection{1. Data Source}
\begin{itemize}
    \item \textbf{Role:} Validation/Error Metrics ONLY.
    \item \textbf{Path:} `...\_Periodic\_SinPiX/2025\_11\_22\_1730/burgers.npz`
    \item \textbf{Keys:} `t`, `x`, `usol`.
\end{itemize}

\subsubsection{2. Problem Statement}
\begin{itemize}
    \item \textbf{Governing Equation:} Viscous Burgers' Equation (Conservative Form) used as a vanishing viscosity approximation for the Inviscid problem.
    \[
    u_t + \partial_x(0.5 u^2) - \nu u_{xx} = 0
    \]
    \begin{itemize}
        \item \textbf{Note:} $\nu$ is a trainable parameter controlled by the continuation schedule.
    \end{itemize}
    \item \textbf{Entropy Condition (Crucial for Shocks):}
    \[
    \partial_t(0.5 u^2) + \partial_x(u^3/3) \le 0
    \]
    \item \textbf{Domain:} $x \in [-1, 1]$, $t \in [0, 1]$.
    \item \textbf{BCs:} Periodic in $x$.
    \item \textbf{IC:} $u(x,0) = -\sin(\pi x)$.
    \item \textbf{Problem Type:} Forward (Solving from IC/BCs).
    \item \textbf{Well-Posedness:} Hyperbolic conservation law developing shocks. Requires vanishing viscosity and entropy constraints for the unique physical solution.
\end{itemize}

\subsubsection{3. Base Model Selection}
\begin{itemize}
    \item \textbf{Model:} MLP with Hybrid Adam and SSBroyden.
    \item \textbf{Architecture Mode:} \textbf{Chebyshev KAN} (per Trial 21 success).
\end{itemize}

\subsubsection{4. Main Architecture}
\begin{itemize}
    \item \textbf{Structure:} Input $\rightarrow$ Normalization $\rightarrow$ Feature Embedding $\rightarrow$ KAN Main Body $\rightarrow$ Output Transformation $\rightarrow$ Output.
\end{itemize}

\paragraph{4.1. Input Transformations}
\begin{itemize}
    \item \textbf{Normalization:}
    \begin{itemize}
        \item $t \in [0, 1]$ (No change needed, but ensure bounds).
        \item $x \in [-1, 1]$ (No change needed).
        \item \textbf{CRITICAL:} Apply normalization \textit{inside} the forward pass.
    \end{itemize}
    \item \textbf{Feature Embeddings:}
    \begin{itemize}
        \item \textbf{Periodic Fourier Features:} Applied to $x$ \textbf{ONLY}.
        \item \textbf{Degree:} 10.
        \item \textbf{Input to Network:} Concatenate \verb|[t, Fourier(x)]|.
    \end{itemize}
\end{itemize}

\paragraph{4.2. Main Body Architecture (KAN)}
\begin{itemize}
    \item \textbf{Type:} Chebyshev KAN.
    \item \textbf{Scaling:} Based on Base MLP (width=30).
    \begin{itemize}
        \item \verb|width_KAN| $= 16$.
        \item \verb|depth| $= 3$ hidden layers.
        \item \verb|polynomial_degree| $= 3$.
    \end{itemize}
\end{itemize}

\paragraph{4.3. Output Transformation (Ansatz)}
\begin{itemize}
    \item \textbf{Goal:} Enforce IC and Periodic BCs hard.
    \item \textbf{Ansatz:}
    \[
    u_{pred}(t,x) = -\sin(\pi x) + t \cdot \text{NN}(t, \text{Fourier}(x))
    \]
    \begin{itemize}
        \item \textit{Explanation:} At $t=0$, $u = -\sin(\pi x)$. Periodicity is enforced by the Fourier features on $x$.
    \end{itemize}
\end{itemize}

\subsubsection{5. Constraint Implementation Strategy}
\begin{itemize}
    \item \textbf{IC:} Hard-constrained via Ansatz.
    \item \textbf{BCs:} Periodic Hard-constrained via Fourier Features.
    \item \textbf{Dirichlet:} None required explicitly.
    \item \textbf{Loss-Based Constraints:} PDE Residual and Entropy Condition.
\end{itemize}

\subsubsection{6. Loss Function Assembly}
\begin{itemize}
    \item \textbf{Symbolic Formula:}
    \[
    L_{total} = L_{pde} + \lambda_{ent} L_{ent}
    \]
    \item \textbf{Definitions:}
    \begin{itemize}
        \item $L_{pde}$: MSE of $(u_t + \partial_x(0.5 u^2) - \nu u_{xx})$.
        \item $L_{ent}$: MSE of $\text{ReLU}(\partial_t(0.5 u^2) + \partial_x(u^3/3))^2$.
        \item \textit{Note:} $L_{ent}$ penalizes violations of the entropy condition.
    \end{itemize}
\end{itemize}

\paragraph{6.1. Continuation Strategy (Vanishing Viscosity)}
\begin{itemize}
    \item \textbf{Parameter:} $\nu$ (Viscosity).
    \item \textbf{Phase 1 (Adam):} Static $\nu = 0.01$.
    \item \textbf{Phase 2 (SSBroyden):} Annealing Schedule.
    \begin{itemize}
        \item \textbf{CRITICAL:} Reinitialize Hessian to Identity at each step.
        \item \textbf{Schedule:}
        \begin{enumerate}
            \item $\nu=0.005$, Dwell=5,000
            \item $\nu=0.002$, Dwell=10,000
            \item $\nu=0.001$, Dwell=15,000
            \item $\nu=0.0005$, Dwell=20,000
            \item $\nu=0.00035$, Dwell=20,000
            \item $\nu=0.0002$, Dwell=30,000
            \item $\nu=0.0001$, \textbf{Dwell=80,000} (Increased per Advisor).
        \end{enumerate}
    \end{itemize}
\end{itemize}

\subsubsection{7. Weighting Strategy}
\begin{itemize}
    \item \textbf{Local Weights (`lambdas`):}
    \begin{itemize}
        \item Use \textbf{vRBA} for $L_{pde}$.
        \item \textbf{Adam:} Use for Weighting \& Sampling.
        \item \textbf{SSBroyden:} Use for \textbf{Sampling ONLY}.
    \end{itemize}
    \item \textbf{Global Weights (`lam`):}
    \begin{itemize}
        \item \textbf{Strategy:} Self-Balancing (GradNorm).
        \item \textbf{Adam:} Dynamic updates for $\lambda_{ent}$ (keep $\lambda_{pde}=1$).
        \item \textbf{SSBroyden:} \textbf{FREEZE} $\lambda_{ent}$ at the final Adam value.
    \end{itemize}
\end{itemize}

\subsubsection{8. Visualization Strategy}
\begin{itemize}
    \item Standard plots. Ensure $u$ is saved.
\end{itemize}

\subsubsection{9. Saving Data}
\begin{itemize}
    \item Save $u_{pred}$, $x$, $t$ as \verb|burgers_pred.npz|.
\end{itemize}

\paragraph{Main Model:}
[MLP with Hybrid Adam and SSBrodyden]

\paragraph{Enhancements:}
[Periodic Fourier Features, KAN]

\section{Additional Examples Strategies}

\subsection{Inverse 2D Steady-State Incompressible Navier-Stokes (Lid-Driven Cavity) (SciML- Step 1)}

\subsubsection{Latest Strategy Report}
(This was the strategy for the \textit{last successful run})

\subsubsection{Scientific Strategy Report}

\subsubsection{0. Triage \& Strategy}
\begin{itemize}
    \item \textbf{Role:} Training Data ($L_{data}$ included).
    \item \textbf{Advisor Recommendation:} \textbf{ACCEPTED}.
    \begin{itemize}
        \item \textbf{Action:} Strictly remove Random Fourier Features (RFF). Use a standard MLP to leverage its spectral bias as a low-pass filter against the 5\% noise.
    \end{itemize}
    \item \textbf{Critic Recommendation:} \textbf{ACCEPTED}.
    \begin{itemize}
        \item \textbf{Action:} Switch from primitive variables to a \textbf{Streamfunction-Pressure formulation} ($\psi, p$) to enforce mass conservation ($\nabla \cdot \mathbf{u} = 0$) exactly by construction.
    \end{itemize}
\end{itemize}

\subsubsection{1. Data Source}
\begin{itemize}
    \item \textbf{Path:} \texttt{LLM/Crunch\_LLM/Documents/codes/Data/ldc\_Re500.npz}
    \item \textbf{Preprocessing Instructions:}
    \begin{enumerate}
        \item Load keys: \texttt{x}, \texttt{y}, \texttt{u}, \texttt{v}, \texttt{p}.
        \item \textbf{Sampling:} Randomly select \textbf{500 indices} from the flattened arrays.
        \item \textbf{Noise Injection:} Add \textbf{5\% Gaussian noise} to the selected \texttt{u} values \textbf{ONLY}.
        \begin{itemize}
            \item \texttt{u\_train = u\_selected + 0.05 * std(u\_selected) * random\_normal(...)}
        \end{itemize}
        \item \textbf{Constraint:} Do \textbf{NOT} use \texttt{v} or \texttt{p} for training. Keep them strictly for validation/error metrics.
    \end{enumerate}
\end{itemize}

\subsubsection{2. Problem Statement}
\begin{itemize}
    \item \textbf{Governing Equations:} 2D Steady Incompressible Navier-Stokes (Streamfunction Formulation).
    \begin{itemize}
        \item \textbf{Ansatz:} $u = \frac{\partial \psi}{\partial y}, \quad v = -\frac{\partial \psi}{\partial x}$.
        \item \textbf{Continuity:} Automatically satisfied ($\frac{\partial^2 \psi}{\partial y \partial x} - \frac{\partial^2 \psi}{\partial x \partial y} = 0$).
        \item \textbf{x-Momentum:} $u \partial_x u + v \partial_y u + \partial_x p - \frac{1}{Re} (\partial_{xx} u + \partial_{yy} u) = 0$
        \item \textbf{y-Momentum:} $u \partial_x v + v \partial_y v + \partial_y p - \frac{1}{Re} (\partial_{xx} v + \partial_{yy} v) = 0$
    \end{itemize}
    \item \textbf{Domain:} $(x,y) \in [0,1] \times [0,1]$.
    \item \textbf{Boundary Conditions (Soft Constraints):}
    \begin{itemize}
        \item \textbf{Lid ($y=1$):} $u = 1 - \frac{\cosh(50(x-0.5))}{\cosh(25)}$, $v=0$.
        \item \textbf{Walls ($y=0, x=0, x=1$):} No-slip ($u=0, v=0$).
    \end{itemize}
    \item \textbf{Problem Type:} \textbf{Inverse Problem}.
    \item \textbf{Has Data Loss:} \textbf{Yes} (Sparse noisy $u$ observations).
\end{itemize}

\paragraph{2.1 Inverse Problem Parameterization}
\begin{itemize}
    \item \textbf{Unknown:} Reynolds Number ($Re$).
    \item \textbf{Strategy:} Sigmoid Scaling (Scalar).
    \item \textbf{Bounds:} $[300, 525]$.
    \item \textbf{Formula:}
    \[
    Re_{est} = 300 + (525 - 300)\cdot \text{sigmoid}(\alpha)
    \]
    \item \textbf{Initialization:} Initialize $\alpha$ such that $Re_{est} \approx 300$ (Lower Bound) to start in the most stable (viscous) regime.
\end{itemize}

\subsubsection{3. Base Model Selection}
\begin{itemize}
    \item \textbf{Selection:} MLP with Hybrid Adam and SSBrodyden.
\end{itemize}

\subsubsection{4. Main Architecture}
\begin{itemize}
    \item \textbf{Inputs:} $(x, y)$.
    \item \textbf{Outputs:} $(\psi, p)$. (Note: Output dimension is 2).
\end{itemize}

\paragraph{4.1 Input Transformations}
\begin{itemize}
    \item \textbf{Normalization:} Min-Max normalize $(x, y)$ to $[-1, 1]$ \textit{inside} the model.
    \item \textbf{Feature Embeddings:} \textbf{None}.
    \begin{itemize}
        \item \textit{Rationale:} Standard MLP is required to filter high-frequency noise.
    \end{itemize}
\end{itemize}

\paragraph{4.2 Main Body}
\begin{itemize}
    \item \textbf{Type:} MLP (Multi-Layer Perceptron).
    \item \textbf{Spec:} Standard Template Defaults (e.g., 4 layers, 50 units, tanh activation).
\end{itemize}

\paragraph{4.3 Output Transformation}
\begin{itemize}
    \item \textbf{Strategy:} \textbf{None}.
    \begin{itemize}
        \item \textit{Rationale:} The network outputs raw potentials $(\psi, p)$. Boundary conditions are applied via soft loss on the derived velocities.
    \end{itemize}
\end{itemize}

\subsubsection{5. Constraint Implementation Strategy}
\begin{itemize}
    \item \textbf{Strategy:} \textbf{Loss-Based (Soft Constraints)}.
    \item \textbf{Complexity Check:}
    \begin{itemize}
        \item $L_{pde}$ (2 Momentum equations) + $L_{bc}$ (4 walls) + $L_{data}$ (Sensors).
        \item Count $> 3$, but exact BCs are forbidden by the user.
    \end{itemize}
    \item \textbf{Implementation Note:}
    \begin{itemize}
        \item Derive velocities $u, v$ from $\psi$ using automatic differentiation \textit{inside} the loss function.
        \item Apply BCs and Data loss to these derived quantities.
    \end{itemize}
\end{itemize}

\subsubsection{6. Loss Function Assembly}
\[
L_{total} = L_{pde} + L_{bc} + L_{data}
\]
\begin{itemize}
    \item \textbf{Term Breakdown:}
    \begin{itemize}
        \item \textbf{$L_{pde}$:} MSE of x-Momentum and y-Momentum residuals (using derived $u, v$).
        \item \textbf{$L_{bc}$:} MSE of $(u_{derived}, v_{derived})$ against Dirichlet conditions on all 4 boundaries.
        \item \textbf{$L_{data}$:} MSE of $(u_{derived} - u_{sensor})$ at the 500 training points.
    \end{itemize}
\end{itemize}

\paragraph{6.1 Continuation Strategy}
\begin{itemize}
    \item \textbf{Not Applicable:} The parameter $Re$ is learned dynamically. Initialization at the lower bound serves as the stability anchor.
\end{itemize}

\subsubsection{7. Weighting Strategy}
\begin{itemize}
    \item \textbf{$L_{pde}$:} Use \textbf{vRBA} (Variational Residual-Based Attention) for local adaptive weighting.
    \item \textbf{Global Weighting:}
    \begin{itemize}
        \item \textbf{Phase 1 (Adam):} Use \textbf{Self-Balancing Strategy} to balance PDE, BC, and Data terms.
        \item \textbf{Phase 2 (SSBroyden):} \textbf{Freeze} the weights determined by Adam.
    \end{itemize}
\end{itemize}

\subsubsection{8. Visualization \& Metrics}
\begin{itemize}
    \item \textbf{Standard Plots:} Keep template visualizations for fields and errors.
    \item \textbf{New Plot:} Add a subplot showing the \textbf{Evolution of $Re_{est}$} vs Iterations. Include a horizontal line for the True $Re=500$.
    \item \textbf{Metrics:}
    \begin{itemize}
        \item Compute relative $L_2$ error for $u$ and $v$ using the \textbf{full original dataset} (ground truth).
        \item Compute absolute error for $Re$ against True $Re=500$.
        \item \textbf{Explicitly print} these metrics at the end of training.
    \end{itemize}
\end{itemize}

\subsubsection{9. Saving Data}
\begin{itemize}
    \item \textbf{Format:} \texttt{.npz}
    \item \textbf{Variables:}
    \begin{itemize}
        \item \texttt{Re}: The final estimated Reynolds number.
        \item \texttt{x\_sensor}, \texttt{y\_sensor}: Coordinates of the 500 training points.
        \item \texttt{u\_sensor}: The noisy training data values.
        \item \texttt{v\_hidden\_at\_sensor}: The inferred $v$ values (derived from $\psi$) at the sensor locations.
    \end{itemize}
\end{itemize}

\subsubsection{List of Specialized Architectural Modules}
\[
\text{[None]}
\]

\subsection{Inverse 2D Steady-State Incompressible Navier--Stokes (Lid-Driven Cavity) (SciC- Step 2)}

\subsubsection{Latest Strategy Report}

\subsubsection{1. Problem Statement}

\paragraph{1. Mathematical Specification}

\begin{itemize}
    \item \textbf{Governing Equations:} \\
    The 2D steady-state incompressible Navier--Stokes equations:
    \[
    \nabla\cdot\mathbf{u} = 0, \qquad
    (\mathbf{u}\cdot\nabla)\mathbf{u} = -\nabla p + \frac{1}{Re}\nabla^2\mathbf{u}.
    \]

    \item \textbf{Domain:} \\
    $(x,y)\in[0,1]\times[0,1]$

    \item \textbf{Boundary Conditions:}
    \begin{itemize}
        \item Top lid ($y=1$):
        \[
        u(x,1) = 1 - \frac{\cosh(C_0(x-0.5))}{\cosh(0.5C_0)},\quad C_0=50,\qquad v(x,1)=0.
        \]
        \item No-slip on $y=0$, $x=0$, $x=1$:
        \[
        \mathbf{u}=(0,0).
        \]
    \end{itemize}

    \item \textbf{Constants \& Parameters:}
    \begin{itemize}
        \item $Re=525$
    \end{itemize}

    \item \textbf{Expected Physics:} \\
    Lid-driven cavity flow with a primary recirculation cell; at $Re\approx 525$ the vortex core shifts and boundary layers sharpen.
\end{itemize}

\subsubsection{2. Strategic Methodology}
\begin{itemize}
    \item \textbf{Selected Main Method:} Steady Continuous Galerkin (FEM) in 2D.
    \item \textbf{Rationale:} The request is strictly steady-state; a direct nonlinear solve (Newton/Picard) is efficient and standard for incompressible flow in a cavity.
\end{itemize}

\subsubsection{3. Implementation Requirements}
\begin{itemize}
    \item \textbf{Configuration:} Use template defaults, subject to the requirements below.
\end{itemize}

\paragraph{Continuation Strategy (Robustness Requirement):}
Direct Newton at $Re=525$ from a zero initial guess may fail. Use \textbf{Reynolds-number continuation}, e.g.
\[
Re: 100 \to 250 \to 525,
\]
using each converged solution as the initial guess for the next.

\paragraph{External Data Integration:}
\begin{itemize}
    \item Load \textbf{Sensors} from: \\
    \texttt{.../Inverse\_NavierStokes\_Find\_Re\_HiddenField\_v/2025\_11\_27\_0058/results.npz}
    \item Load \textbf{True Data} from: \\

    \texttt{LLM/Crunch\_LLM/Documents/codes/Data/ldc\_Re500.npz}
\end{itemize}

\textbf{Action:}
\begin{itemize}
    \item Interpolate/evaluate the new numerical solution at:
    \begin{enumerate}
        \item the sensor coordinates $(x_{\text{sensor}},y_{\text{sensor}})$, and
        \item the unstructured nodes $(x_{\text{true}},y_{\text{true}})$.
    \end{enumerate}
\end{itemize}

\textbf{Metrics to Report:}
\begin{itemize}
    \item Relative $L_2$ error for $u$ and $v$ at sensors (numerical vs.\ PINN values).
    \item Relative $L_2$ error for $u$ and $v$ against the true dataset (numerical vs.\ true).
\end{itemize}

\subsubsection{4. Visualization Strategy}
\begin{itemize}
    \item \textbf{Task:} Generate a composite summary figure.
    \item \textbf{Panels (2$\times$2 layout):}
    \begin{enumerate}
        \item Magnitude of velocity field from the new numerical solution: $|\mathbf{u}|$.
        \item Absolute error map for $u$: $|u_{\text{num}}-u_{\text{true}}|$.
        \item Absolute error map for $v$: $|v_{\text{num}}-v_{\text{true}}|$.
        \item Scatter plot comparing PINN sensor values vs.\ numerical solution at sensor points
        (include identity line).
    \end{enumerate}
    \item \textbf{Constraints:}
    \begin{itemize}
        \item Save as: \texttt{summary\_all.png}
        \item Style: automatic scaling.
    \end{itemize}
\end{itemize}

\subsection{Viscous Burgers Operator Network (DeepONet) }

\subsubsection{Problem}
This will be an operator network that maps the initial condition to the full solution up to t=t\_end (extracted from the data)

\subsubsection{Reference Data:}
You can find the data here: \\
\texttt{LLM/Crunch\_LLM/Data/Operator\_Network\_Examples/Burger\_1000\_5000.mat}

\noindent\rule{\linewidth}{0.4pt}

\subsubsection{Latest Strategy Report}
(This was the strategy for the \textit{last successful run})

\subsubsection{Scientific Strategy Report}

\textbf{0. Triage - Determine Strategy} \\
The \texttt{[ADVISOR\_REPORT]} indicates a scientific failure and recommends a specific cure: ``escalating by re-running the best-performing QR-DeepONet strategy but changing the decomposition from QR to SVD.'' The best-performing QR-DeepONet was Run 7, which used an MLP for both the Branch and Trunk. Therefore, the strategy will be to implement an SVD-DeepONet with MLP Branch and Trunk, replicating the architecture of Run 7 but with SVD decomposition. This will be referred to as Strategy 6: SVD-DeepONet (MLP).

\textbf{1. Data Source} \\
The data for this problem is located at

\texttt{LLM/Crunch\_LLM/Data/Operator\_Network\_Examples/Burger\_1000\_5000.mat}.

The Planner must update the data loading cell to use this path and verify the keys and variables within the \texttt{.mat} file.

\textbf{2. Problem Statement} \\
This problem involves solving the viscous Burgers equation, defined as $\partial u/\partial t + u\partial u/\partial x = \nu\partial^{2}u/\partial x^{2}$, with a viscosity parameter $\nu = 1/1000$. The domain is $x \in [0,1]$ and $t \in [0,1]$, with periodic boundary conditions. This is an operator network problem that maps the initial condition $u_0(x)$ to the full solution $u(t,x)$ over time. This is a function-to-trajectory problem where the function space (F\_space) is not equal to the solution space (S\_space).

\textbf{3. Input (F\_space)} \\
The input function F\_space is the initial condition $u_0(x)$.
\begin{itemize}
    \item \textbf{Domain dimension (d):} 1D ($x$).
    \item \textbf{Value dimension (k):} Scalar ($u_0$).
    \item \textbf{Structure:} The input is a grid representing the initial function over $x$.
\end{itemize}

\textbf{4. Output (S\_space)} \\
The output S\_space is the solution $u(t,x)$.
\begin{itemize}
    \item \textbf{Domain dimension (d\_y):} 2D ($t,x$).
    \item \textbf{Value dimension (q):} Scalar ($u$).
    \item \textbf{Characteristics:} The problem specifies periodic boundary conditions. Given the nature of the Burgers equation, the solution may exhibit high-frequency features (e.g., shocks).
\end{itemize}

\textbf{5. Normalization (CRITICAL: In-Graph)} \\
All normalization must be differentiable and implemented \textit{inside} the model's \texttt{apply} function (i.e., within the Flax Module).
\begin{itemize}
    \item \textbf{Trunk (per-dimension):} The Planner must pass the $(t, x)$ coordinates to the model \textit{un-normalized}. The model itself must compute \texttt{t\_min, t\_max, x\_min, x\_max} from the training data and perform the normalization to $[-1, 1]$ \textit{inside} the \texttt{\_\_call\_\_} method before passing to feature layers.
    \item \textbf{Branch (global):} The Planner must pass the input function $u(x, t=0)$ \textit{un-normalized}. The model's Branch network must compute \texttt{u\_min, u\_max} and normalize the input to $[-1, 1]$ \textit{inside} the \texttt{\_\_call\_\_} method before the MLP.
    \item \textbf{Activations:} Both the Trunk and Branch will use \texttt{nn.tanh} activations, which are consistent with $[-1, 1]$ normalization.
\end{itemize}

\textbf{6. Architecture} \\
\textbf{Strategy 6: SVD-DeepONet (MLP)}
\begin{itemize}
    \item \textbf{Methodology:} \texttt{SVD\_DeepONet}.
\end{itemize}

\textbf{7a. Branch:}
\begin{itemize}
    \item \textbf{Type:} Standard MLP.
    \item \textbf{Architecture:} Following Sizing Rule 1, the Branch will have 6 layers with 100 neurons per layer, and a final embedding dimension of 120. The architecture will be \texttt{[100, 100, 100, 100, 100, 120]}.
\end{itemize}

\textbf{7b. Trunk:}
\begin{itemize}
    \item \textbf{Type:} Standard MLP.
    \item \textbf{Architecture:} Following Sizing Rule 1, the Trunk will match the Branch's embedding dimension and layer structure. The architecture will be \texttt{[100, 100, 100, 100, 100, 120]}.
\end{itemize}

\textbf{Enhancements:} No enhancements (e.g., Periodic Fourier Features) will be added. The Advisor's recommendation is to replicate the architecture of Run 7 (QR-DeepONet MLP), which did not include such enhancements, only changing the decomposition to SVD.

\textbf{Number of iterations:} \\
Instruct the Planner to set \texttt{max\_iterations=200000}.

\textbf{7. Loss Function}
\begin{itemize}
    \item \textbf{7a. Lambda (vRBA):} The loss function will incorporate \texttt{Lambda\_spatial (N,P)} and \texttt{Lambda\_latent (N,embed\_dim)}.
    \item \textbf{7b. Error\_fn:} The primary error function will be Mean Squared Error (MSE).
    \item \textbf{7c. Weights (W):} Since the output $q$ is scalar ($q=1$), the weight \texttt{W} will be calculated as \texttt{1.0 / max(abs(S))}.
\end{itemize}

\textbf{8. Optimization (Learning Rate)} \\
The \texttt{[ADVISOR\_REPORT]} did not provide specific learning rate recommendations. Therefore, default values for SVD/QR-DeepONet will be applied.
\begin{itemize}
    \item \textbf{Branch Optimizer:} Instruct the Planner to set the initial learning rate to \texttt{2e-3} for the Branch optimizer.
    \item \textbf{Trunk Optimizer:} Instruct the Planner to set the initial learning rate to \texttt{2e-3} for the Trunk optimizer.
    \item \textbf{Decay Schedule:} Instruct the Planner to use the default decay schedule: set \texttt{decay\_rate = 0.99} and \texttt{decay\_steps = 2500} for all optimizers.
\end{itemize}

\textbf{IMPLEMENTATION CODE REQUIRED:}
\begin{itemize}
    \item Main Model: \\
    \texttt{SVD\_DeepONet}
    \item Enhancements: \\
    None
\end{itemize}

\end{document}